%% file: modelmap2025.tex
\pdfoutput=1
\documentclass[11pt]{article}

\usepackage[whole]{bxcjkjatype}
\usepackage[preprint]{acl}
\usepackage{times}
\usepackage{latexsym}
\usepackage[T1]{fontenc}
\usepackage[utf8]{inputenc}
\usepackage{microtype}
\usepackage{inconsolata}
\usepackage{microtype}
\usepackage{comment}
\usepackage{graphicx}
\usepackage{amsmath}
\usepackage{amssymb}
\usepackage{amsfonts}
\usepackage{array}
\usepackage{multirow}
\usepackage{booktabs}
\usepackage{adjustbox}
\usepackage{subcaption}
\usepackage{stmaryrd}
\usepackage{siunitx}
\usepackage{bm}
\newcommand{\cX}{\mathcal{X}}
\newcommand{\cV}{\mathcal{V}}
\newcommand{\bbR}{\mathbb{R}}
\newcommand{\bmxi}{\bm{\xi}}
\newcommand{\bmzeta}{\bm{\zeta}}
\newcommand{\bmet}{\bm{\eta}}
\newcommand{\bmt}{\bm{\theta}}
\newcommand{\bmell}{\bm{\ell}}
\newcommand{\bmq}{\bm{q}}
\newcommand{\bmQ}{\bm{Q}}
\newcommand{\bmL}{\bm{L}}
\newcommand{\bmb}{\bm{b}}
\newcommand{\bme}{\bm{e}}

\newcommand{\bmzero}{\bm{0}}
\newcommand{\bmone}{\bm{1}}

\DeclareMathOperator*{\Var}{Var}
\DeclareMathOperator*{\Cov}{Cov}
\DeclareMathOperator*{\bbE}{\mathbb{E}}
\usepackage{longtable}

\newcommand\blfootnote[1]{%
  \begingroup
  \renewcommand\thefootnote{}\footnote{#1}%
  \addtocounter{footnote}{-1}%
  \endgroup
}

\title{Mapping 1,000+ Language Models via the Log-Likelihood Vector}

\author{
Momose Oyama${}^{1,2}$\quad Hiroaki Yamagiwa${}^{1}$\quad Yusuke Takase${}^{1}$ \quad Hidetoshi Shimodaira${}^{1,2}$\\
${}^{1}$Kyoto University\quad
${}^{2}$RIKEN\\
\texttt{oyama.momose@sys.i.kyoto-u.ac.jp, h.yamagiwa@i.kyoto-u.ac.jp,}\\
\texttt{y.takase@sys.i.kyoto-u.ac.jp, shimo@i.kyoto-u.ac.jp}
}

\begin{document}
\maketitle

\begin{abstract}
To compare autoregressive language models at scale, we propose using log-likelihood vectors computed on a predefined text set as model features.  
This approach has a solid theoretical basis: when treated as model coordinates, their squared Euclidean distance approximates the Kullback-Leibler divergence of text-generation probabilities.  
Our method is highly scalable, with computational cost growing linearly in both the number of models and text samples, and is easy to implement as the required features are derived from cross-entropy loss.  
Applying this method to over 1,000 language models, we constructed a ``model map,'' providing a new perspective on large-scale model analysis.
\blfootnote{Our code and data are available at \url{https://github.com/shimo-lab/modelmap}.}
\end{abstract}

\section{Introduction} \label{sec:introduction}

Language models have been evolving rapidly, and their community has expanded significantly.  
To understand this landscape and its future directions, it is essential to systematically analyze model similarity and positioning based on language modeling principles.  
On the Hugging Face Hub, models are categorized by name and attributes, while other studies assess similarity based on outputs~\cite{yax2025phylolm} or activations~\cite{zhou-etal-2025-linguistic}. 
Leaderboards~\cite{open-llm-leaderboard, open-llm-leaderboard-v2, chiang2024chatbot} are commonly used to assess model standings.

\begin{figure}[!t]
\centering
\includegraphics[width=\linewidth]{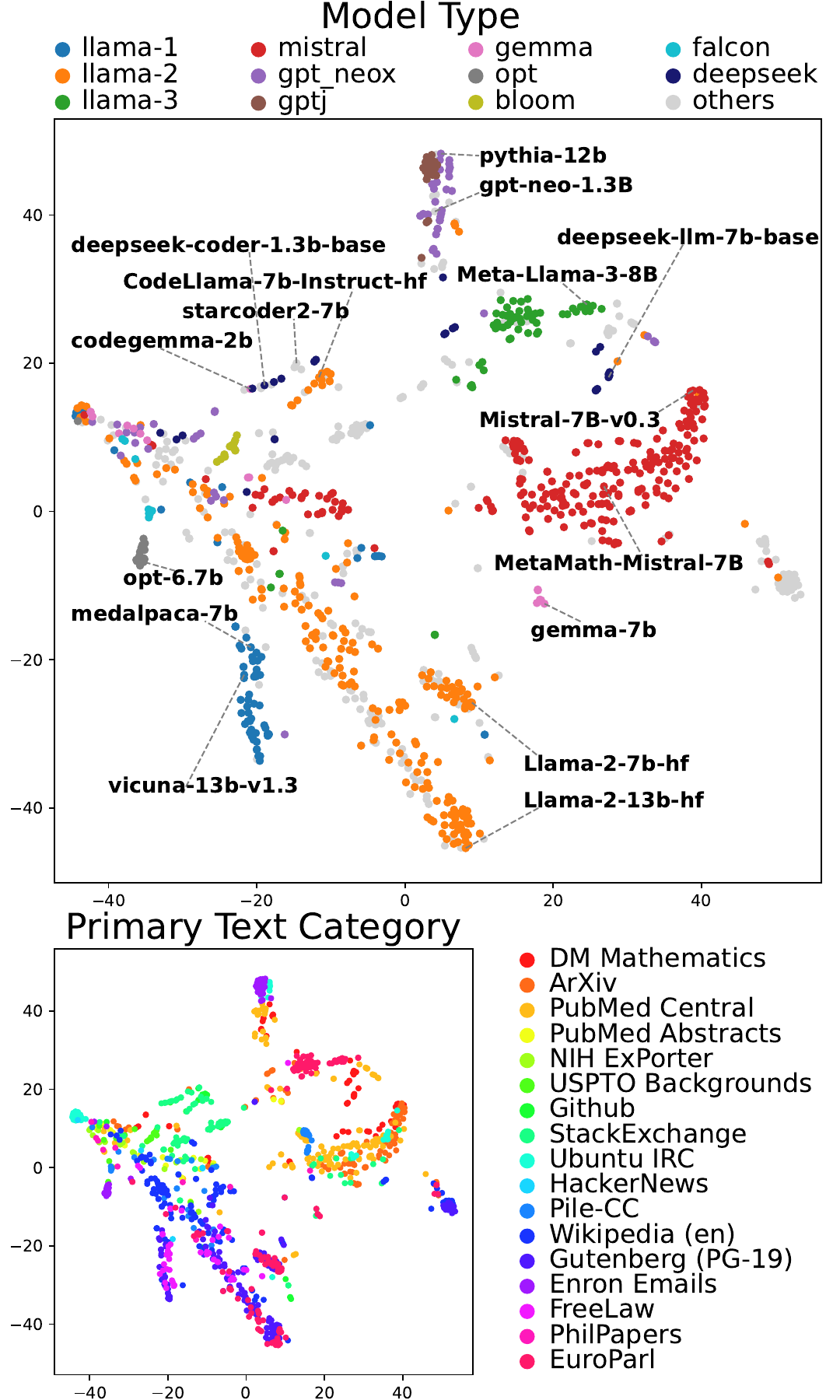}
\caption{
Map of 1,018 language models. Their log-likelihood vectors are visualized using t-SNE.  
(Top) Colors indicate model types.  
(Bottom) Colors indicate the model's ``primary text category,'' the text category where the model achieves the highest standardized log-likelihood score among 17 text categories.  
See Section~\ref{sec:model-map} for details.
}
\label{fig:model-map-tsne-intro}
\end{figure}

Since language models are probability models, we propose representing each model using coordinates that capture the geometric structure of the space of probability distributions.
Concretely, we define a language model's coordinates as its log-likelihood vector across a large collection of texts.  
Figure~\ref{fig:model-map-tsne-intro} shows a model map obtained through dimensionality reduction applied to the coordinates of 1,018 language models.  
This visualization reveals that models of the same type tend to cluster together, while models in close proximity often share the same primary text category, forming a continuous distribution across the map.

\begin{table*}[!t]
\centering

\begin{adjustbox}{width=\linewidth}
\begin{tabular}{@{}lrlrlr@{}}
\toprule
meta-llama/Meta-Llama-3-8B & KL [bpb] & google/codegemma-2b & KL [bpb] & deepseek-ai/deepseek-llm-7b-base & KL [bpb]  \\
\cmidrule(lr){1-2}\cmidrule(lr){3-4}\cmidrule(lr){5-6}
Undi95/Meta-Llama-3-8B-hf & 4.56e-06 & deepseek-ai/deepseek-coder-1.3b-instruct & 2.46 & deepseek-ai/deepseek-moe-16b-base & 0.194 \\
dfurman/Llama-3-8B-Orpo-v0.1 & 0.0104 & bigcode/starcoderbase-1b & 2.69 & deepseek-ai/deepseek-llm-7b-chat & 0.364 \\
migtissera/Tess-2.0-Llama-3-8B & 0.0428 & deepseek-ai/deepseek-coder-1.3b-base & 3.14 & deepseek-ai/deepseek-moe-16b-chat & 0.368 \\
freewheelin/free-llama3-dpo-v0.2 & 0.101 & bigcode/gpt\_bigcode-santacoder & 3.92 & deepseek-ai/DeepSeek-V2-Lite & 0.455 \\
jondurbin/bagel-8b-v1.0 & 0.164 & deepseek-ai/deepseek-coder-6.7b-instruct & 3.97 & deepseek-ai/ESFT-vanilla-lite & 0.456 \\
migtissera/Llama-3-8B-Synthia-v3.5 & 0.190 & Qwen/CodeQwen1.5-7B-Chat & 4.09 & deepseek-ai/DeepSeek-V2-Lite-Chat & 0.868 \\
nvidia/Llama3-ChatQA-1.5-8B & 0.191 & NTQAI/Nxcode-CQ-7B-orpo & 4.11 & mistralai/Mistral-7B-Instruct-v0.1 & 1.356 \\
ruslanmv/Medical-Llama3-8B & 0.206 & Salesforce/codegen-6B-multi & 4.24 & statking/zephyr-7b-sft-full-orpo & 1.616 \\
FairMind/Llama-3-8B-4bit-UltraChat-Ita & 0.296 & bigcode/starcoderbase-7b & 4.44 & Severian/ANIMA-Phi-Neptune-Mistral-7B & 1.692 \\
NousResearch/Hermes-2-Theta-Llama-3-8B & 0.330 & deepseek-ai/deepseek-coder-6.7b-base & 4.87 & sethuiyer/Medichat-Llama3-8B & 1.718 \\
\bottomrule
\end{tabular}
\end{adjustbox}
\caption{
Top 10 nearest neighbors among the 1,018 language models for each model listed in the first row.
The values indicate the KL divergence measured in bits per byte (bpb), as defined in Section~\ref{sec:byte-normalized-KL}.
These are computed using formula~(\ref{eq:main-KL-xi}) in Section~\ref{sec:model-text}, by multiplying the original KL divergence by 0.001484.
Tables for the nearest neighbors of the models labeled in the top panel of Fig.~\ref{fig:model-map-tsne-intro} are provided in Appendix~\ref{sec:top10-tables}.
}
\label{tab:nearest-neighbor-models}
\end{table*}

We find that the distances in our defined coordinate system accurately capture relationships among language models. On this map, each point represents a single model, with those having similar text-generation probability distributions appearing closer together and those with more distinct distributions positioned farther apart. In Section~\ref{sec:model-text}, we show that the squared Euclidean distance in this coordinate system approximates the Kullback-Leibler (KL) divergence among models. Table~\ref{tab:nearest-neighbor-models} lists the nearest neighbors for each language model. For example, many of the closest neighbors of \texttt{meta-llama/Meta-Llama-3-8B}~\cite{llama3modelcard} also contain \texttt{Llama-3} in their names.

Several studies have explored methods for comparing language models (see Appendix~\ref{sec:related-work}).  
In particular, prior work on comparing generated text includes approaches that construct phylogenetic trees based on model-generated text~\citep{yax2025phylolm}  
and approaches that measure differences in text-generation probabilities conditioned on given prompts using KL divergence~\citep{melamed-etal-2024-prompts}.  
However, these methods require generating text with each model, thus incurring the cost of pairwise distance computations, which becomes prohibitively expensive at large scale.  
By contrast, our method does not involve actual text generation; instead, we compute generation probabilities on a predefined text corpus.  
This enables us to derive model coordinates without pairwise comparisons, allowing efficient large-scale comparison of many models.  

To gain insights from visualizing model attributes on the model map, in Section~\ref{sec:model-map}, we analyze various attributes\footnote{Attributes include model type, primary text category, model performance, model size, and creation date.} and their relationships. Additionally, by comparing log-likelihood and benchmark performance, we demonstrate the ability to detect data leakage. Then, in Section~\ref{sec:predict-performance}, we treat the log-likelihood vector as a feature and show how it can predict benchmark performance. Finally, in Section~\ref{sec:confirmation}, we validate the theoretical relationship between model coordinates and KL divergence through experiments.

\section{Mapping Language Models into the Space of Text Probability Distributions} \label{sec:model-text}

In this section, we present our proposed method.  
Sections~\ref{sec:loglikelihood-vector} and \ref{sec:double-centering} introduce model feature vectors derived from text-generation probabilities.  Section~\ref{sec:theorem-KL-text} demonstrates that the squared Euclidean distance in the coordinate system built using these features approximates the KL divergence between models.  
Section~\ref{sec:model-coordinates} offers an interpretation of the resulting model coordinates.  
An extension of this method, which defines model coordinates using the sequence of conditional probabilities for generating a given token sequence, is presented in Appendix~\ref{sec:model-token}.

\subsection{Autoregressive language models} \label{sec:language-models}

Let \(\mathcal{X}\) be the set of all possible texts, and let \(\mathcal{V}\) be the token vocabulary.  
A text \(x \in \mathcal{X}\) is represented as a sequence of tokens:
\[
x = (y_1, \dots, y_n), \quad y_t \in \mathcal{V}.
\]
Denoting the maximum text length by \(n_{\textrm{max}}\), we have
\(\mathcal{X} = \bigcup_{n=0}^{n_{\textrm{max}}} \mathcal{V}^n.\)
We consider a set of \(K\) language models \(\{p_i\}_{i=1}^K\).  
With \(y_0\) denoting the beginning-of-sequence (BOS) token, each language model \(p_i\) predicts the next token \(y_t\) given the preceding token sequence \(y^{t-1} = (y_0, \dots, y_{t-1})\).  
Thus, the conditional probability defined by \(p_i\) is given by
\[
y_t \sim p_i(y_t \mid y^{t-1}), \quad t = 1, \dots, n.
\]
Accordingly, the probability of a text \(x\) under model \(p_i\), denoted \(x \sim p_i\), is
\[
p_i(x) = \prod_{t=1}^{n} p_i(y_t \mid y^{t-1}).
\]
In addition to the \(K\) language models \(p_1, \dots, p_K\), we introduce a language model \(p_0\) that represents an underlying distribution for theoretical purposes.  
We assume we have a dataset (corpus)
\[
D = (x_1, x_2, \dots, x_N) \in \mathcal{X}^N,
\]
consisting of \(N\) texts, where each text is independently drawn from \(p_0\).

\subsection{Log-likelihood vector}\label{sec:loglikelihood-vector}

For a model \(p_i\), the probability of generating a text \(x\) is denoted \(p_i(x)\).  
Following the convention in statistical model selection, we refer to \(p_i(x)\) as the likelihood of model \(p_i\) given the text \(x\).  
The log-likelihood $\ell_i(x) = \log p_i(x)$ is then computed as
\[
\ell_i(x) = \sum_{t=1}^n \log p_i(y_t \mid y^{t-1}).
\]
In language model implementations, \(-\ell_i(x)\) corresponds to the cross-entropy loss for the text \(x\), and \(\exp(-\ell_i(x)/n)\) is known as the perplexity.

Our approach is straightforward. Given that the dataset \(D\) consists of \(N\) texts, we use the log-likelihood vector
\[
\bmell_i = (\ell_i(x_1), \dots, \ell_i(x_N))^\top \in \mathbb{R}^N
\]
as the feature vector for model \(p_i\).  
The first step in our model analysis is to construct the log-likelihood matrix
\[
\bm{L} = (\bmell_1, \dots, \bmell_K)^\top \in \mathbb{R}^{K \times N}
\]
by stacking the vectors \(\bmell_i\) for the \(K\) models.

\subsection{Double centering} \label{sec:double-centering}

As a preprocessing step for model analysis, we apply a technique called double centering~\cite{borg2005modern} to \(\bm{L}\).  
First, we perform row-wise centering.  
The mean of each row, referred to as the mean log-likelihood, is given by
\[
    \bar \ell_i = \sum_{s=1}^N \ell_i(x_s)/N.
\]
Subtracting this value from each component of \(\bmell_i\), we define the centered log-likelihood vector \(\bmxi_i = (\xi_{i1},\ldots,\xi_{iN})^\top \in \mathbb{R}^N\), where
\[
\xi_{is} := \ell_i(x_s) - \bar \ell_i, \quad s=1,\ldots,N.
\]
Next, we apply column-wise centering to the matrix of centered feature vectors \((\bmxi_1,\ldots,\bmxi_K)^\top\).  
The mean vector is
\(
\bar{\bmxi} = \frac{1}{K} \sum_{i=1}^K \bmxi_i,
\)
and by subtracting this vector from each \(\bmxi_i\), we define the double-centered log-likelihood vector
\[
\bmq_i = \bmxi_i - \bar{\bmxi}.
\]
For further details, see Appendices~\ref{sec:theory-double-centering} and \ref{sec:error-analysis}.

\subsection{Kullback-Leibler divergence}\label{sec:theorem-KL-text}

The Kullback-Leibler (KL) divergence is often used to measure how far apart two models \(p_i\) and \(p_j\) are in the space of probability distributions\footnote{\(\mathbb{E}(\cdot)\) denotes expectation and \(\mathrm{Var}(\cdot)\) denotes variance.}. It is defined as
\begin{align}
\mathrm{KL}(p_i,p_j) &= \sum_{x\in\mathcal{X}} p_i(x)\log\frac{p_i(x)}{p_j(x)} \nonumber\\
&= \mathbb{E}_{x\sim p_i}\bigl(\ell_i(x) - \ell_j(x)\bigr).
\label{eq:main-KL-def}
\end{align}
We assume the dataset \(D\) is generated from an unknown underlying model \(p_0\) and that the models \(p_i\) and \(p_j\) provide good approximations of \(p_0\). Under this assumption, the KL divergence can be approximated as follows:
\begin{align}
2 \,\mathrm{KL}(p_i,p_j)\approx
\mathrm{Var}_{x\sim p_0}\bigl(
\ell_i(x)-\ell_j(x)
\bigr).
\label{eq:main-KL-var-model}
\end{align}
While the definition of KL divergence in \eqref{eq:main-KL-def} involves the expectation of \(\ell_i(x) - \ell_j(x)\), the approximation in \eqref{eq:main-KL-var-model} takes the form of a variance. This result is somewhat surprising yet quite insightful.  
Notably, although KL divergence is not symmetric in the two models, the approximation in \eqref{eq:main-KL-var-model} is symmetric.  
We estimate \eqref{eq:main-KL-var-model} from the dataset \(D\) as
\begin{align}
2\,\mathrm{KL}(p_i,p_j) \approx \|\bmq_i - \bmq_j\|^2/N.
\label{eq:main-KL-xi}
\end{align}
Thus, if we regard the model coordinates of \(p_i\) as \(\bmq_i/\sqrt{N}\) by scaling with \(N\), then the squared Euclidean distance between two points approximates \(2\,\mathrm{KL}(p_i,p_j)\).  

The main results, namely \eqref{eq:main-KL-var-model} and \eqref{eq:main-KL-xi}, are proved in Appendix~\ref{sec:theory-text} using the theory of exponential family of distributions~\citep{barndorff2014information,efron1978geometry,efron2022exponential,amari1982-10.1214/aos/1176345779}, similar to the discussion on the relationship between the norm of embeddings and KL divergence~\citep{oyama-etal-2023-norm}.  
Although the concepts of model map and model coordinates have been discussed in statistics~\citep{Shimodaira1993modelmap,shimodaira1998graphical,shimodaira2001multiple}, and there have been a few applications of model maps~\citep{Shimodaira2005,shimodaira2019selective}, they have received little attention or use in practice.

\subsection{Model coordinates}\label{sec:model-coordinates}

We primarily use \(\bmq_i\) as the feature vector of model \(p_i\) and refer to it as the model coordinates\footnote{We refer to the Euclidean space where the vectors \( \bm{\ell}_i \), \( \bm{\xi}_i \), or \( \bm{q}_i \), possibly scaled, reside as the \emph{log-likelihood space}.}. 
As shown in \eqref{eq:main-KL-xi}, the squared Euclidean distance in the \(\bmq\)-coordinate system approximates the KL divergence between language models\footnote{That is, \(\|\bmq_i-\bmq_j\|^2 \approx 2N\,\mathrm{KL}(p_i,p_j)\). For simplicity, we omit the constant scaling factor in expressions of this type.},  
indicating that \(\bmq_i\) represents the position of \(p_i\) in the space of probability distributions.  
Since \(\bmxi_i\) differs from \(\bmq_i\) only by an offset from the origin, \(\bmxi_i\) also serves as a model coordinate, and \(\|\bmq_i - \bmq_j\|^2 = \|\bmxi_i - \bmxi_j\|^2\).  
However, we prefer \(\bmq_i\) for its more interpretable components and thus adopt it throughout this paper.

For visualization purposes, we mainly use \(\bmell_i\) as the coordinates of the model map, as \(\bmell_i\) can be intuitively interpreted as encoding \(\sqrt{N}\,\bar{\ell}_i\) in the ``height'' dimension and \(\bmq_i\) in the ``horizontal'' dimensions.  
As shown in Appendix~\ref{sec:theory-height},
\begin{align}
    \|\bmell_i - \bmell_j\|^2
   =  \|\bmq_i - \bmq_j\|^2 + N(\bar \ell_i - \bar \ell_j)^2,
\label{eq:lik-decomposition}
\end{align}
which means the squared Euclidean distance in the \(\bmell\)-coordinate system can be decomposed into the sum of \(2N\,\mathrm{KL}(p_i,p_j)\) and \(N\,(\bar{\ell}_i - \bar{\ell}_j)^2\).

\section{Experimental Setup} \label{sec:experiment-settings}

\begin{figure}[t]
\centering
\includegraphics[width=\linewidth]{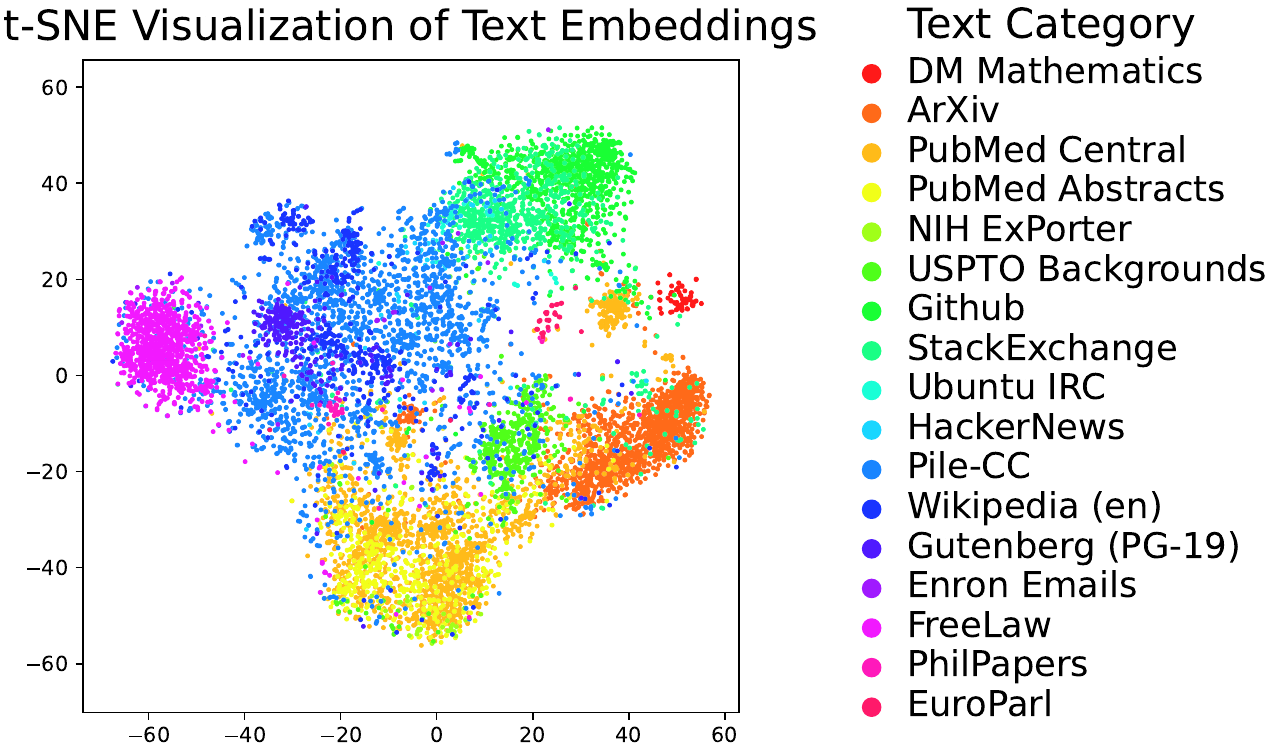}
\caption{
Text embeddings for 10,000 texts in dataset $D$, computed via \texttt{simcse-roberta-large}~\cite{gao2021simcse} and visualized with t-SNE. Colors indicate 17 text categories.}
\label{fig:dataset-visualize}
\end{figure}

\begin{figure*}[!t]
    \centering
    \includegraphics[width=\linewidth]{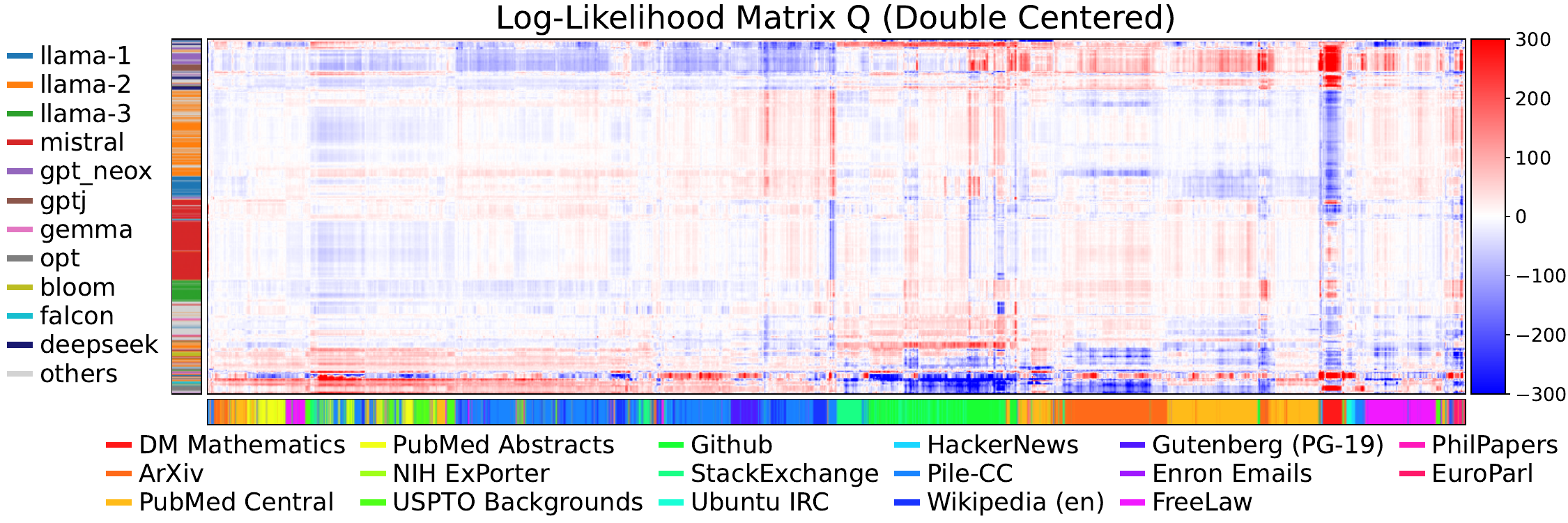}
    \caption{
    The double-centered log-likelihood matrix \(\bm{Q}\), with rows and columns reordered by hierarchical clustering.  
    Each row corresponds to one of the 1,018 models, color-coded by model type.  
    Each column represents one of the 10,000 texts, color-coded by text category.
    }
\label{fig:matrix-Q}
\end{figure*}

\begin{figure}[!t]
    \centering
    \includegraphics[width=\linewidth]{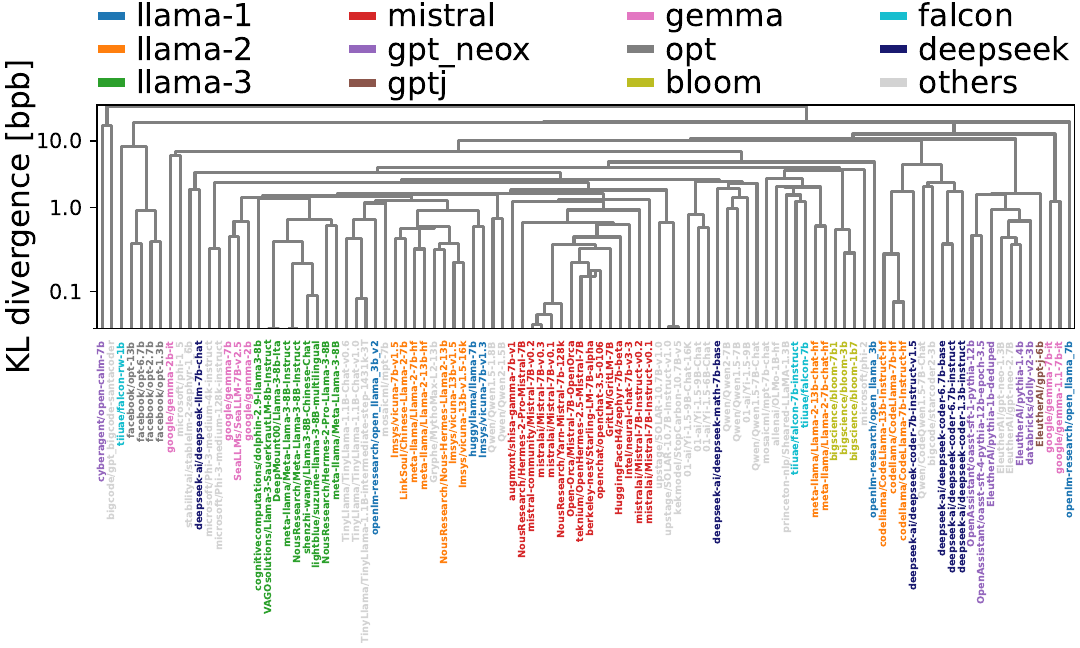}
    \caption{
    Hierarchical clustering of the top 100 most-downloaded models, based on their feature vectors \(\bm{q}_i\).  
    Model names are color-coded by model type. KL divergence is reported in units of bits per byte (bpb).
    }
    \label{fig:cluster-100-models}
\end{figure}

We describe the key components of our experiment. 
In particular, Section~\ref{sec:text-set} explains the procedure for selecting the 10,000 texts used to compute the model coordinates, 
and Section~\ref{sec:models} discusses how we selected the 1,018 language models. 
Further details are given in Appendix~\ref{sec:experiment-details}.

\subsection{Selection of text data} \label{sec:text-set}

The texts used for computing the language models' coordinates were extracted from the Pile~\cite{arxiv:2101.00027},  
with five categories of copyrighted material removed\footnote{\url{https://huggingface.co/datasets/monology/pile-uncopyrighted}}.  
This yielded a dataset \(D\) consisting of 10,000 texts, each tagged with a category label from the Pile.  
Figure~\ref{fig:dataset-visualize} visualizes these texts.

To build the dataset, we began by dividing the first 1M texts from the Pile Uncopyrighted corpus into 1,024-byte chunks (UTF-8 encoded).  
In cases where decoding errors occurred, we truncated by one byte at a time.  
Chunks smaller than 256 bytes were discarded, resulting in about 5.7M valid chunks.  
From these, we randomly sampled 10,000 texts to create the final dataset used for computing model coordinates.
The average length of these 10,000 texts was 972.3188 bytes.

\subsection{Selection of language models} \label{sec:models}

We used \(K = 1{,}018\) language models in total. Of these, 1,000 were selected from models listed on Open LLM Leaderboard v1. Specifically, we considered CausalLM models ranging from 1B to 13B parameters and ranked them by their number of downloads over the 30 days preceding February 1, 2025\footnote{Hugging Face's API provides the total number of downloads in the past 30 days.}. We initially selected the top 1,100 models by download count and attempted log-likelihood calculations. Among these, 1,011 successfully produced valid log-likelihood values, and we chose the 1,000 most frequently downloaded from that set.  
In addition, we included 18 models from the DeepSeek language model series. We obtained information on model parameter sizes and architectures from the Leaderboard. Appendix~\ref{sec:experiment-details} provides basic information on the selected models and details on how model types were defined. A complete list of models used in this study is given in Appendix~\ref{app:model_list}.

\subsection{Computation of the log-likelihood} \label{sec:logp-computation}

The log-likelihood matrix \(\bm{L}\) was computed in float16 precision,
with the bottom 2\% of values clipped.  
This clipping mitigates the large impact of extremely low likelihoods on \eqref{eq:main-KL-xi}.  
After computing \(\bm{L}\), we applied row-wise and column-wise centering to obtain the double-centered log-likelihood matrix \(\bm{Q}\).
Figure~\ref{fig:matrix-Q} visualizes \(\bm{Q}\).  
Each value in the matrix can be interpreted in two ways: as the relative probability of a text for each model or as the relative likelihood of a model for each text.  
Both models and texts exhibit clustering patterns.  
Figure~\ref{fig:cluster-100-models} shows a dendrogram of the top 100 models.  
We examined the effective dimension from the perspective of feature vector dimensionality reduction and found that the cumulative contribution ratio, based on the sum of squared singular values of \(\bm{Q}\), reached 90\% at 42 dimensions and 95\% at 82 dimensions.

\subsection{Obtaining the leaderboard scores}\label{sec:leaderboard_scores}

We obtained benchmark scores for the language models used in our experiments from Open LLM Leaderboard v1\footnote{\url{https://huggingface.co/spaces/open-llm-leaderboard-old/open_llm_leaderboard}}.  
Between April 2023 and June 2024, this leaderboard evaluated language models on six tasks:  
AI2 Reasoning Challenge (ARC)~\cite{arxiv:1803.05457}, HellaSwag~\cite{arxiv:1905.07830}, MMLU~\cite{arxiv:2009.03300}, TruthfulQA~\cite{arxiv:2109.07958}, Winogrande~\cite{arxiv:1907.10641}, and GSM8K~\cite{arxiv:2110.14168}.  
Along with individual task scores, we also use the average score across all six tasks, referred to as 6-TaskMean.

\subsection{Byte-normalized KL divergence for cross-experiment comparison} 
\label{sec:byte-normalized-KL}
The KL divergence of text generation, $\textrm{KL}(p_i, p_j)$, generally increases with the number of tokens in the text. As a result, it cannot be directly compared with the KL divergence from other experiments using different text data. For models that use the same tokenizer, one can normalize the KL divergence by the average number of tokens in the text to obtain the KL divergence per token. However, when comparing KL divergence between models with different tokenizers, as in this study, it is more appropriate to normalize by the average text length in bytes and use the KL divergence per byte. For instance, if $\textrm{KL}(p_i, p_j) = 1{,}000$, dividing by the average text length of 972.3188 bytes yields a KL divergence per byte of $1{,}000 / 972.3 = 1.028$ nats $= 1.484$ bits\footnote{To convert the code length unit from nats to bits, multiply by \( 1/\log 2 = 1.4427 \).}.

KL divergence can be understood from the perspective of coding theory as a measure of how much longer a message becomes when encoded using an incorrect probability distribution. $\textrm{KL}( p_i, p_j )$ represents the extra code length required when encoding text generated from probability distribution \( p_i \) using a different distribution \( p_j \). Dividing this value by the average text length in bytes gives the additional code length per byte. In the example above, this means that, due to differences between the models, an extra 1.484 bits are needed to encode each byte of text.

\begin{figure*}[!t]
    \centering
    \includegraphics[width=\linewidth]{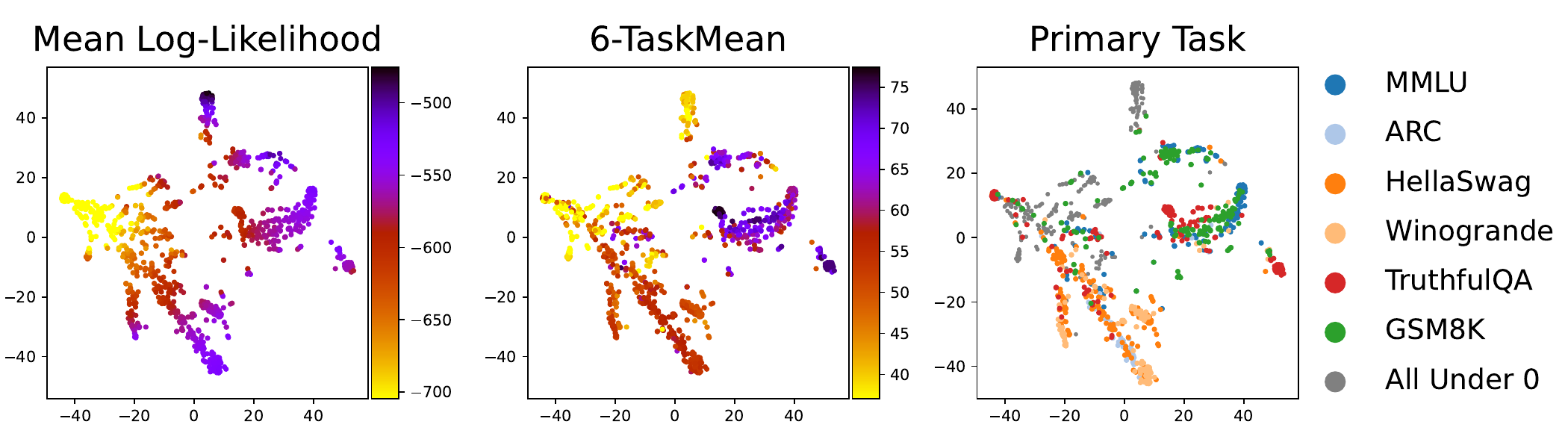}
    \caption{
    Model maps illustrating model performance.  
    From left to right, the panels show each model's mean log-likelihood, 6-TaskMean score, and the ``primary task,'' meaning the task for which each model achieves its highest standardized score among the six tasks (Appendix~\ref{sec:standard-score}).  
    The color bar is clipped at the 10th percentile for mean log-likelihood and 6-TaskMean, with darker colors indicating better performance.  
    In the primary task panel, models with standardized scores below zero on all six tasks are labeled ``All Under 0.''
    }
    \label{fig:modelmap-logp-score-task}
\end{figure*}

\begin{figure}[!t]
    \centering
    \includegraphics[width=\linewidth]{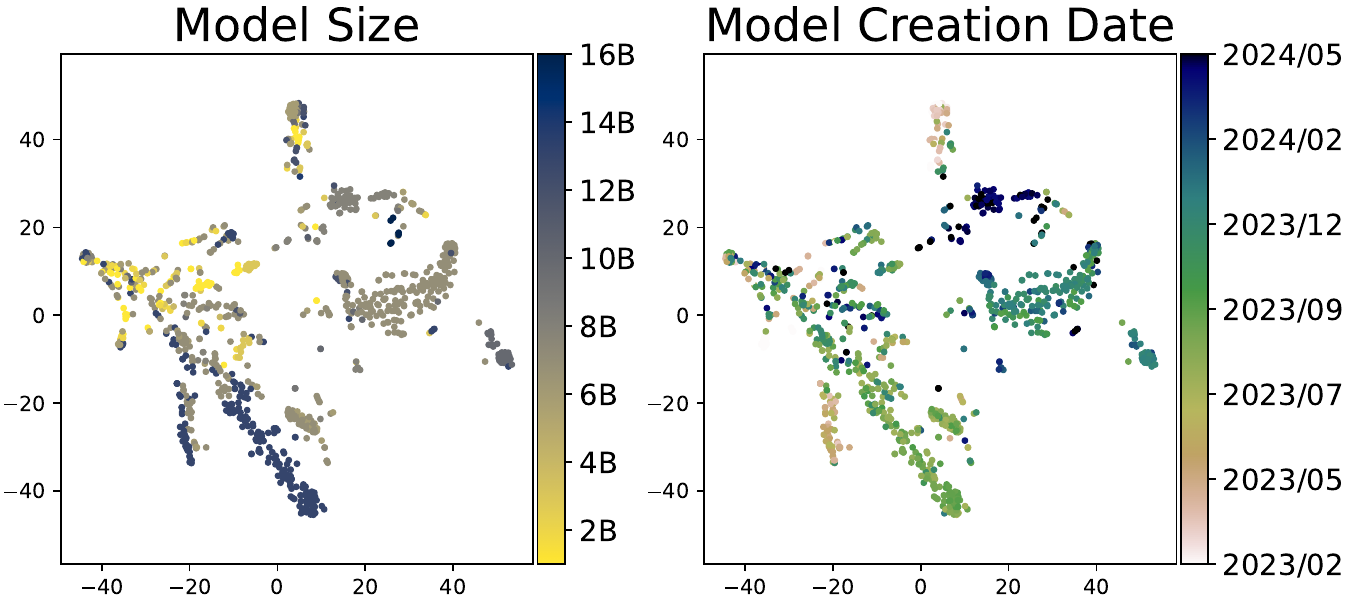}
    \caption{Model maps color-coded by (Left) number of parameters and (Right) model creation date.}
    \label{fig:size-date}
\end{figure}

\section{Map of Language Models} \label{sec:model-map}

We applied t-SNE~\cite{vanderMaaten-2008-tsne} to the log-likelihood matrix $\bm{L}$ for dimensionality reduction\footnote{The perplexity value was set to 30, and we used the scikit-learn implementation~\citep{DBLP:journals/sigmobile/VaroquauxBLGPM15}.}.  
Using this visualization, we analyze the insights gained from the model map in this section.  
While this paper presents model maps using $\bm{L}$, alternative maps using the double-centered log-likelihood matrix $\bm{Q}$, as well as a model map with labels for all language models, are available in Appendix~\ref{sec:add-model-map}.

\subsection{Visualizing attributes on the model map} \label{sec:model-map-discussion}

\paragraph{Model type.}

The top panel of Fig.~\ref{fig:model-map-tsne-intro} visualizes the distribution of model types, with each model color-coded according to its type.  
We observe that models belonging to the same type tend to cluster together, forming distinct regions on the map (e.g., \texttt{llama-2}, \texttt{mistral}, and \texttt{gemma}).  
In particular, models optimized for coding tasks\footnote{For example, \texttt{starcoder2-7b}~\citep{arxiv:2402.19173}, \texttt{deepseek-coder-1.3b-base}~\citep{arxiv:2401.14196}, \texttt{codegemma-2b}~\citep{arxiv:2406.11409} and \texttt{CodeLlama-7b-Instruct-hf}~\citep{arxiv:2308.12950}.} appear in a relatively compact region, suggesting that these models share notable similarities in their probability distributions.

\paragraph{Text category.}

The bottom panel of Fig.~\ref{fig:model-map-tsne-intro} shows the model map with each model color-coded according to the text category in which it achieves the highest
standardized log-likelihood score (Appendix~\ref{sec:standard-score}). 
From this figure, we see that models exhibiting high likelihoods for the same text category are grouped together.  
Notably, the cluster containing coding-specialized models in the top panel aligns with the GitHub/StackExchange region in the bottom panel, suggesting that these models have relatively high likelihoods for text originating from GitHub and StackExchange.

\paragraph{Model performance.}

Figure~\ref{fig:modelmap-logp-score-task} visualizes two evaluation metrics: mean log-likelihood and benchmark task performance.  
From the left and central panels, we see that both metrics exhibit similar trends on the map, where models that lie close together tend to show similar metric values.  
Additionally, in the right panel, the GSM8K/MMLU region corresponds to the ArXiv/PubMed Central region in the bottom panel of Fig.~\ref{fig:model-map-tsne-intro},  
suggesting that models with high likelihoods on academic and scientific texts also tend to perform well on mathematical reasoning and academic knowledge-intensive tasks.

\paragraph{Model size and creation date.}  

Figure~\ref{fig:size-date} shows the distribution of models by size and creation date.  
Compared to Fig.~\ref{fig:modelmap-logp-score-task}, newer models generally perform better, but model size does not always correlate with performance, as some smaller models perform comparably to larger ones.

\subsection{Detection of data leakage} \label{sec:the-pile}

Since the text data we used was extracted from the Pile corpus, models that were pre-trained on the Pile are likely to exhibit higher log-likelihood values than their actual capabilities measured by benchmarks.  
We analyze this effect using the model map in Fig.~\ref{fig:the-pile}.  
The left panel highlights models that used the Pile for pre-training.  
The right panel shows models with high mean log-likelihood relative to their 6-TaskMean score.  
The alignment between these two distributions suggests that models pre-trained on the Pile tend to achieve higher likelihoods on our text data, while their benchmark performance remains comparatively lower.

\section{Predicting Model Performance from Model Coordinates} \label{sec:predict-performance}

As shown in the central panel of Fig.~\ref{fig:modelmap-logp-score-task}, the positioning of models on the map suggests that a model's benchmark performance may be inferred from its coordinates.  
In this section, we conduct a regression analysis using the \(\bmq\)-coordinates to predict benchmark scores and evaluate predictive performance.

\subsection{Benchmark scores and models}  

We use six benchmark scores from Open LLM Leaderboard v1, as described in Section~\ref{sec:leaderboard_scores}.  
Our experiments are conducted on 996 models for which these benchmark scores are available\footnote{  
From the 1,018 models, we excluded four that lacked TruthfulQA scores and 18 from \texttt{deepseek-ai}, as their benchmark scores were incomplete due to the models being relatively new, leaving 996 for analysis. For simplicity, we denote the number of models as $K$.  
}.

\begin{table*}[t!]
\scriptsize
    \centering
    \begin{tabular}{lrrrrrrrr}
    \toprule
        & ARC & HellaSwag & MMLU & TruthfulQA & Winogrande & GSM8K & 6-TaskMean & mean log-likelihood\\
        \midrule
        Pearson's $r$ & $0.946$ & $0.909$ & $0.932$ & $0.901$ & $0.941$ & $0.884$ & $0.953$ & $0.989$ \\
        Spearman's $\rho$ & $0.948$ & $0.956$ & $0.934$ & $0.884$ & $0.948$ & $0.857$ & $0.960$ & $0.974$\\
    \bottomrule
    \end{tabular}
\caption{Results of ridge regression for predicting benchmark scores from model coordinates.  
Predictions for 6-TaskMean and mean log-likelihood are also included.
High correlation coefficients are observed across all settings.
See Table~\ref{tab:group-result-regression_mean_and_std} in Appendix~\ref{app:model_pred} for the mean and standard deviation of correlation coefficients across the five data splits.
}
\label{tab:group-result-regression}
\end{table*}

\begin{table*}[t!]
\scriptsize
    \centering
    \begin{tabular}{lrrrrrrr}
    \toprule
        & ARC & HellaSwag & MMLU & TruthfulQA & Winogrande & GSM8K & 6-TaskMean\\
        \midrule
        Pearson's $r$ &  $0.453$ & $0.598$ & $0.346$  & $0.072$ & $0.508$ & $0.239$ & $0.395$\\
        Spearman's $\rho$ & $0.432$ & $0.467$ & $0.422$ & $0.048$ & $0.512$ & $0.364$ & $0.400$\\
    \bottomrule
    \end{tabular}
\caption{
Correlation between mean log-likelihoods and benchmark scores.
The results show that perplexity has only moderate correlations on most tasks, especially compared to model coordinates in Table~\ref{tab:group-result-regression}.
}
\label{tab:meanlogp-vs-benchmark}
\end{table*}

\subsection{Setting for regression analysis}\label{sec:setting_for_ridge}
For each benchmark task, the dataset is given as $\{(\bmq_1, v_1), \dots, (\bmq_K, v_K)\}$,  
where $\bmq_i \in\mathbb{R}^N$ is the double-centered log-likelihood vector of the language model $p_i$,  
and $v_i \in [0, 100]$ is its corresponding benchmark score.
We use ridge regression to predict each benchmark score.  
Let $\bmQ \in\mathbb{R}^{K\times N}$ be the matrix of explanatory variables, and let $\bm{v} = (v_1,\ldots,v_K)^{\top} \in \mathbb{R}^{K}$ be the vector for the target variable.  
The objective function with parameter $\bm{w} \in \mathbb{R}^N$ is given by:  
\begin{align}
    \mathcal{L}(\bm{w}) = \|\bm{v} - \bmQ\bm{w}\|^2 + \alpha \|\bm{w}\|^2,\label{eq:ridge}
\end{align}
where $\alpha \in \mathbb{R}_{>0}$ is a hyperparameter that controls the strength of regularization.
Since the number of variables $N$ is much larger than the sample size $K$ ($N \gg K$), making this a high-dimensional regression setting, we carefully set $\alpha$ using cross-validation to avoid overfitting.

We partition the models into five folds based on model types and perform parameter training and benchmark score prediction\footnote{We used the \texttt{GroupKFold} and \texttt{RidgeCV} implementations provided by scikit-learn~\cite{DBLP:journals/sigmobile/VaroquauxBLGPM15}.}.
To mitigate the effect of randomness, we repeat the data splitting with five different seeds and take the average of the predictions as the final predicted score.  
As evaluation metrics, we compute Pearson's $r$ and Spearman's $\rho$ to measure the correlation between the predicted and benchmark scores.  
Additionally, we conduct experiments by replacing the target variable with 6-TaskMean and mean log-likelihood, leading to a total of eight experimental settings.  
See Appendix~\ref{app:model_pred_training_details} for details.

\begin{figure}[t]
    \centering
    \includegraphics[width=\linewidth]{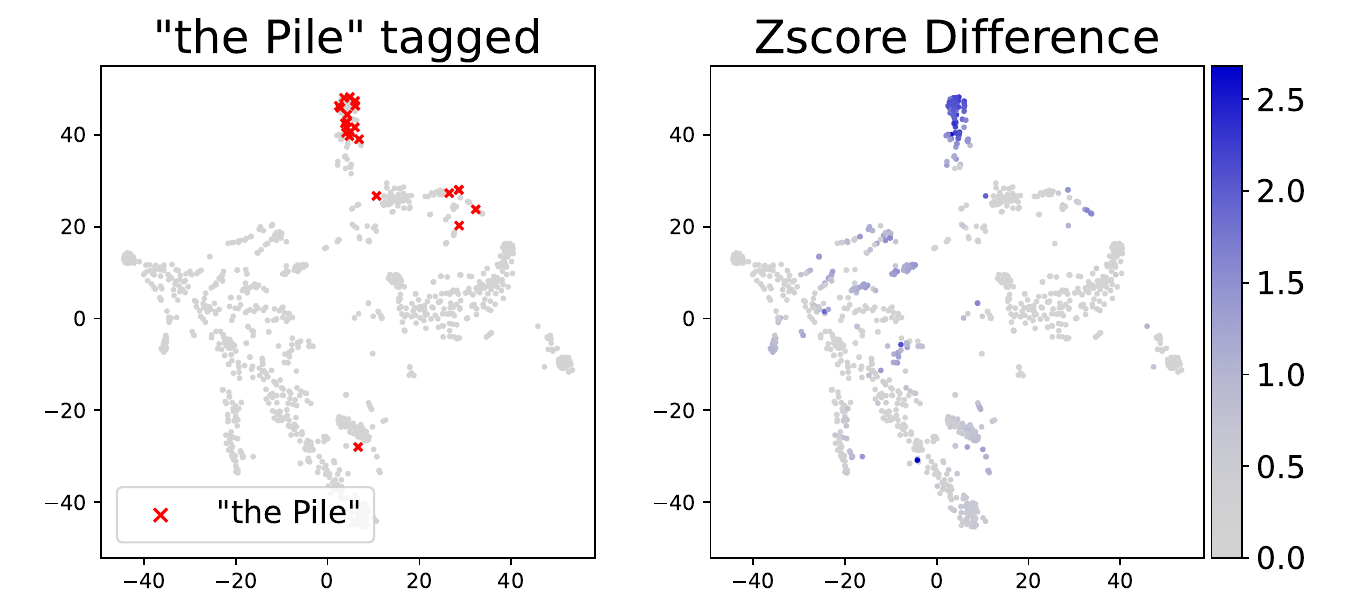}
    \caption{(Left) Models tagged with ``the Pile.'' (Right) Difference between the standardized mean log-likelihood and the standardized 6-TaskMean score.}
    \label{fig:the-pile}
\end{figure}

\subsection{Results and discussion}

\begin{figure}[t]
    \centering
    \includegraphics[width=0.7\linewidth]{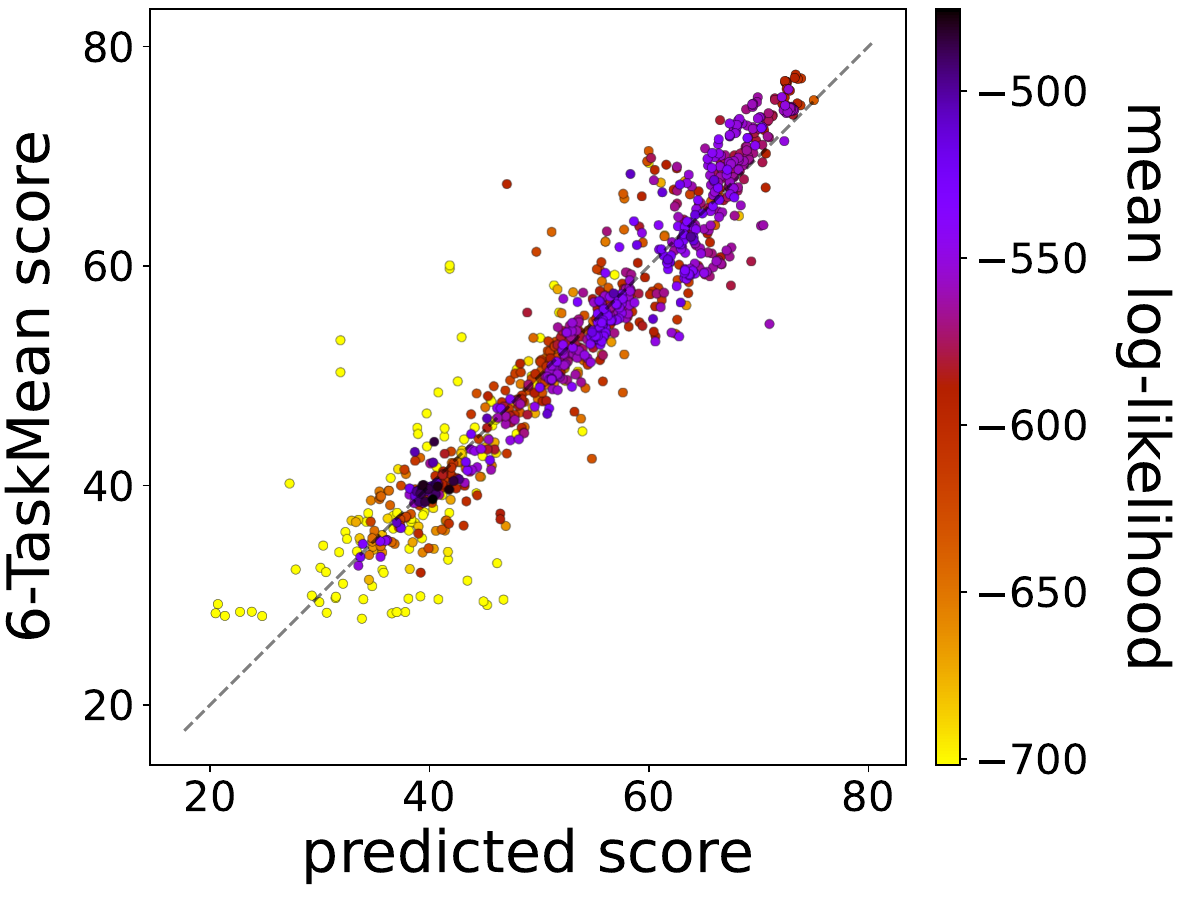}
\caption{
Scatter plot comparing predicted and actual 6-TaskMean scores for test sets (the dashed line indicates the identity line).
Pearson's $r$ is 0.953 and Spearman's $\rho$ is 0.960, indicating strong predictive performance.
Points are color-coded by the mean log-likelihood, with the color scale clipped at the 10th percentile.
While higher mean log-likelihood values tend to correspond to higher benchmark scores, a few models with unusually high log-likelihoods due to data leakage (Section~\ref{sec:the-pile}) deviate from this trend.
Nevertheless, the predictions remain accurate overall.
Scatter plots for individual benchmark tasks are shown in Fig.~\ref{fig:group_5fold_split_all_tasks} in Appendix~\ref{app:model_pred_result_details}.
}
\label{fig:group_5fold_split_average_task}
\end{figure}

\paragraph{Regression analysis.}
Table~\ref{tab:group-result-regression} summarizes the results of the regression analysis using $q$-coordinates.
Across all benchmark tasks, both Pearson's $r$ and Spearman's $\rho$ are consistently high, indicating that the ridge regression model based on model coordinates achieves strong predictive performance.
This trend holds not only for individual tasks but also for the 6-TaskMean, where the correlation coefficients remain high, as illustrated in the scatter plot in Fig.~\ref{fig:group_5fold_split_average_task}.
Even when the target variable is the mean log-likelihood, defined as the simple average over the 10,000 texts, the regression model still achieves high correlation.
This suggests that the model coordinates, despite being double-centered and excluding direct information about the mean, retain a sufficiently rich structure to accurately predict the average log-likelihood.

\paragraph{Performance prediction based on perplexity.}
The mean log-likelihood equals $-\log(\text{perplexity})$ and is often used as a performance indicator.
We evaluated its correlation with benchmark scores, and Table~\ref{tab:meanlogp-vs-benchmark} shows moderate correlations across all tasks, consistent with prior findings~\cite{liu2023same,fang2025wrong,thrush2025improving}.
Unlike mean log-likelihood, which uniformly averages over texts, regression using the log-likelihood matrix $\bmL$ can assign task-dependent weights.
As shown in Table~\ref{tab:group-result-regression_mean_and_std_using_L} in Appendix~\ref{app:model_pred}, its prediction accuracy is nearly the same as that of $\bmQ$.
This indicates that combining $\bmQ$ and the mean in $\bmL$ offers little improvement.
Accurate prediction with $\bmQ$ alone suggests that model positions already encode a structure aligned with benchmark performance.  

\section{Empirical Validation of Theory} \label{sec:confirmation}
In Section~\ref{sec:model-text}, we discussed model coordinates for text probability models.  
Appendix~\ref{sec:model-token} extends this framework to token-sequence probability models.  
These theoretical results state that the squared Euclidean distance in the log-likelihood space approximates the KL divergence between models.  
To validate this, we first evaluate token-level conditional probability models (Section~\ref{sec:experiment-KL-token}), since token-level experiments are generally easier to conduct than text-level experiments.  
We then extend our analysis to text probability models (Section~\ref{sec:experiment-KL-text}).  
Additionally, in Section~\ref{sec:weight-parameter}, we explore the relationship between model weight parameters and model coordinates.

\subsection{Validation of (\ref{eq:main-token-KL-xi}) for token-level models} \label{sec:experiment-KL-token}

\paragraph{Settings.}  
Two models with a shared tokenizer \texttt{Llama-2-7b-hf} and \texttt{Llama-2-7b-chat-hf}~\citep{arxiv:2307.09288}, denoted as $p_1$ and $p_2$, were used.
Following the method described in Appendix~\ref{sec:token-logp}, we computed the model coordinates $\bmzeta_{1}(x)$ and $\bmzeta_{2}(x)$ for each text $x = (y_1, \cdots, y_{n}) \in D$, where these coordinates are centered vectors with elements $\log p_i(y_t|y^{t-1})$ as defined in (\ref{eq:zeta-coordinate}).
We then calculated the squared Euclidean distance between these coordinates, $\|\bmzeta_1(x) - \bmzeta_2(x)\|^2$.  
To obtain the exact KL divergence between models, we used the outputs of the softmax function in the language models.  
We computed the sum of per-token KL divergences:
\(
\sum_{t=1}^{n}\mathrm{KL}(p_1(y_t|y^{t-1}),p_2(y_t|y^{t-1})),
\)
which is also used in \citet{lv-etal-2023-parameter}.

\paragraph{Results and discussion.}  
The left panel of Fig.~\ref{fig:token-text_KL} shows a scatter plot of the squared Euclidean distance and KL divergence for $x\in D$, with a Pearson's correlation coefficient of $r=0.893$.  
This result indicates that (\ref{eq:main-token-KL-xi}) provides a good approximation in actual language models.

\subsection{Validation of (\ref{eq:main-KL-xi}) for text-level models} \label{sec:experiment-KL-text}

\paragraph{Settings.}  
Among the 292 language models sharing the tokenizer with \texttt{Llama-2-7b-hf}~\citep{arxiv:2307.09288}, we excluded the five models with the largest values of \(\sum_{x\in D}\|\bmzeta_i(x)\|^2\), leaving 287 models for our experiment.  
Using the approach described in Section~\ref{sec:model-text}, we computed the model coordinates for each model and calculated the squared Euclidean distance \(\|\bmq_i - \bmq_j\|^2\) between every pair of models.  
Because it is extremely difficult to directly compute \(\mathrm{KL}(p_i, p_j)\), we instead used
\(
\frac{1}{N} \sum_{x \in D} \bigl\| \bmzeta_i(x) - \bmzeta_j(x) \bigr\|^2
\)
as a proxy.  
For a theoretical justification that this quantity approximates \(\mathrm{KL}(p_i, p_j)\), see \eqref{eq:main-token-KL-text-KL-xi} in Appendix~\ref{sec:model-token}.

\begin{figure}[t]
    \centering
    \includegraphics[width=\linewidth]{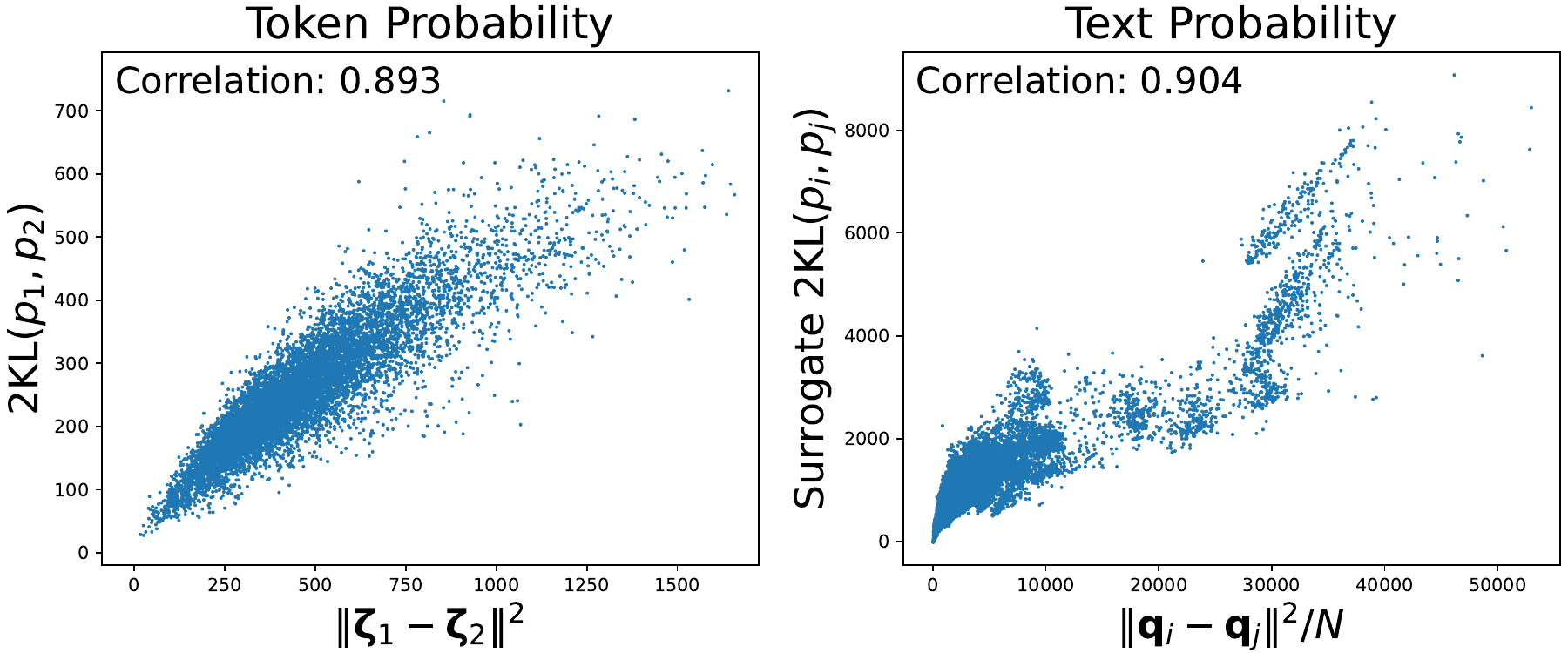}
    \caption{Relationship between the squared Euclidean distance of model coordinates and KL divergence.  
(Left) In the token-level experiment (Section~\ref{sec:experiment-KL-token}), each point represents a text.  
(Right) In the text-level experiment (Section~\ref{sec:experiment-KL-text}), each point represents a pair of models.}

    \label{fig:token-text_KL}
\end{figure}

\paragraph{Results and discussion.}  
As shown in the right panel of Fig.~\ref{fig:token-text_KL}, the scatter plot of squared Euclidean distance versus KL divergence exhibits a Pearson's correlation coefficient of $r=0.904$.  
This finding confirms that the relationship in \eqref{eq:main-KL-xi} holds approximately in practical language models.

\subsection{Relationship between model weights and model coordinates} \label{sec:weight-parameter}

\begin{figure}
    \centering\includegraphics[width=\linewidth]{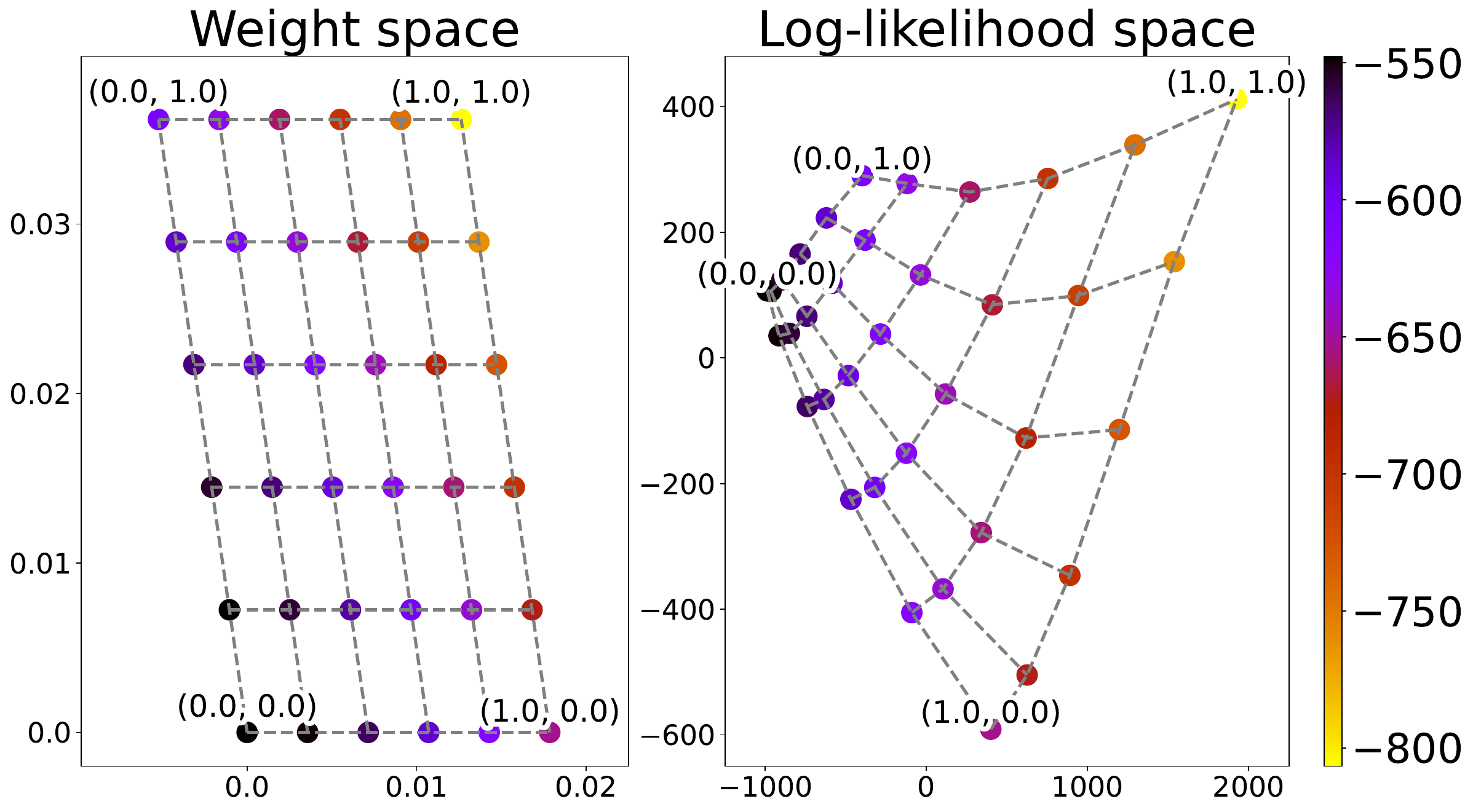}
    \caption{
    Visualization of 36 language models obtained by linearly interpolating pretrained model weights based on \texttt{Llama-2-7b-hf}.  
    Each point is color-coded according to its mean log-likelihood.  
    (Left) Models in the weight parameter space.  
    (Right) Models in the log-likelihood space, represented by the \(\bmq\)-coordinate system.
}    
\label{fig:weight-interpolation-grids-llama}
\end{figure}

For language models with the same architecture, comparison can also be conducted via weight parameters.  
We investigated how the structure of the log-likelihood space aligns with the structure of the weight-parameter space.  
We generated 36 new language models by linearly interpolating the weights of \texttt{Llama-2-7b-hf}, \texttt{Llama-2-7b-chat-hf}~\citep{arxiv:2307.09288}, and \texttt{vicuna-7b-v1.5}~\citep{arxiv:2306.05685}, using a $6\times6$ grid of coefficients.
For these 36 models, we computed text-generation log-likelihoods and visualized the resulting model coordinates in two dimensions using Principal Component Analysis (PCA).  
Figure~\ref{fig:weight-interpolation-grids-llama} shows these 36 models in both weight space and log-likelihood space, where the two-dimensional grid structure is preserved.
Interestingly, the mean log-likelihoods of the interpolated models closely follow the linear interpolation of the three base models' mean log-likelihoods.  
This suggests that such interpolation can be a practical tool for exploring high-performing models.  
See Appendix~\ref{app:weight-parameter} for details.

\section{Conclusion}
We proposed a method to compare autoregressive language models using log-likelihoods on a predefined text set. By interpreting these as model coordinates, we showed that the squared Euclidean distance approximates KL divergence. Experiments on over 1,000 models confirmed the method's effectiveness for analyzing model relationships, predicting benchmark performance, and validating the theory.

\section*{Limitations}
\begin{itemize}
    \item Changing the text data will alter the analysis results of the model map. This is both a limitation and an advantage of the proposed method, because it allows us to choose text data according to the analysis objective. For example, if we want to investigate code-focused language models in more detail, we can increase the proportion of code data from GitHub, thereby increasing the resolution of code-focused models on the model map.
    
    \item The proof (Section~\ref{sec:model-text} and Appendix~\ref{sec:theory-text}) that the squared Euclidean distance in the model coordinate system approximates the KL divergence assumes that the language model's text-generation probabilities closely match the distribution of the text data. When this assumption does not hold, the approximation accuracy decreases. However, even under such circumstances, the model coordinates should still function sufficiently as model features.

    \item If the text data used for the model coordinates is contained in a language model's pre-training corpus, that model's mean log-likelihood may be overestimated. This data leakage, or data contamination, is generally non-negligible, as shown in Fig.~\ref{fig:the-pile} in Section~\ref{sec:the-pile}, which illustrates the effect of using the Pile corpus. However, comparing it against benchmark scores makes it possible to detect such data leakage, and one can remove models that are affected.
Furthermore, the model map based on the \(\bmq\)-coordinate system is robust to contamination in the data,  
as the estimation of KL divergence using the squared Euclidean distance in the \(\bmq\)-coordinate system remains valid even if the true generative model \(p_0\) that produces the dataset \(D\) varies,  
as long as all compared models remain sufficiently close to \(p_0\).  
For instance, even if \(D\) is generated from one of the \(K\) models, such as \(p_i\), the estimation formula (\ref{eq:main-KL-xi}) remains correct (Appendix~\ref{sec:theory-note-modifyp0}).
    
    \item Beyond the data leakage mentioned above, other systematic errors introduced into the model coordinates can also affect the model map. As noted in Appendix~\ref{sec:error-analysis}, $\bmell_i$ and $\bar{\ell}_i$ are susceptible to systematic biases. However, thanks to double centering, $\bmq_i$ is less influenced by bias terms.

    \item Although the calculation of model coordinates is linear time $O(KN)$ in the number of models $K$ and the number of texts $N$, it still requires a non-negligible amount of computation. In our experiments, it took about 10 minutes on a single GPU (RTX 6000 Ada) to compute coordinates ($N=10^4$, float16) for a single 7B model.

    \item Computing the model map visualization from the model coordinates is generally not linear in $K$. For example, t-SNE requires a distance matrix, incurring $O(K^2 N)$ computational cost. However, in modern computing environments, as long as $K$ is not extremely large (e.g., in the millions), the cost of visualization is negligible compared to the cost of calculating the model coordinates.

    \item A sufficiently large number of text samples, $N$, is desirable for the model coordinates. We used $N=10^4$. Since the error in the KL divergence estimate due to randomness decreases proportionally to $N^{-1/2}$, $N$ must be increased according to the desired resolution of the model map.

    \item When using model coordinates as feature vectors, $N=10^4$ can be unwieldy. According to the experiment in Section~\ref{sec:logp-computation}, applying PCA to reduce the dimensionality of $\bmq$-coordinates to 82 dimensions still retains 95\% of the information. However, the predictive performance of such dimension-reduced features has not yet been tested.

    \item Since the language models used in our experiments were obtained from the Open LLM Leaderboard v1 (which ran from April 2023 to June 2024), our discussion of models released after June 2024 is limited.

    \item The results of the task-performance prediction in Section~\ref{sec:predict-performance} should be interpreted conservatively.
We employed a proper cross-validation, splitting the models based on their model types so that the training, validation, and test sets were disjoint, thereby limiting data leakage.
However, if nearly identical models appear in both the training and test partitions (for example, models labeled with different types but built on the same base model and modified only slightly by fine-tuning), prediction becomes artificially easier.
As shown in Tables~\ref{tab:group-result-regression_mean_and_std} and~\ref{tab:result-regression_mean_and_std} in Appendix~\ref{app:model_pred_comp}, random splits that permit such leakage yield higher predictive performance than splits that strictly follow model-type groupings.  
Finally, note that models not listed on the leaderboard were excluded from our evaluation.

    \item The theoretical validation experiment in Section~\ref{sec:confirmation} is limited. Currently, it is difficult to directly compute exact KL divergence values among text-generation probability models, so conducting more precise validation experiments remains a future challenge.

    \item In the method of computing model coordinates from token-sequence conditional probabilities (Appendix~\ref{sec:model-token} and Appendix~\ref{sec:theory-token}), the proof that the squared Euclidean distance in the model coordinate system approximates the KL divergence requires additional assumptions (Appendix~\ref{sec:token-theory-assumptions}). In practice, Assumption~2 does not hold, and due to the variations in (\ref{eq:token-assumtion2}), $\|\bmzeta_i - \bmzeta_j\|^2$ in (\ref{eq:main-token-KL-xi}) tends to overestimate the KL divergence. Nonetheless, even in such situations, token-level model coordinates will still likely function sufficiently as features.
\end{itemize}

\section*{Acknowledgments}
This study was partially supported by JSPS KAKENHI 22H05106, 23H03355, JST CREST JPMJCR21N3, JST BOOST JPMJBS2407.

\bibliography{custom, model_list}

\appendix

\section{Related Work}\label{sec:related-work}
In recent years, research on comparing large language models (LLMs) has gained attention.  
This section provides an overview of existing studies from three perspectives: model parameters, activations\footnote{In general, ``activations'' refer to the intermediate outputs of Transformer models (e.g., the residual stream or neurons)~\cite{bereska2024mechanistic}.}, and 
model outputs.

\paragraph{Comparison of model parameters.}
One approach to comparing LLMs is to analyze their parameters.
\citet{zhu2025independence} proposed a statistical framework for evaluating parameter similarity between different models and introduced a method for determining whether these models were trained independently. 
\citet{horwitz2025charting} compare weights to complement undocumented model relationships on Hugging Face.
Additionally, \citet{NEURIPS2023_1644c9af} focused on parameter changes due to task adaptation, specifically analyzing task vectors\footnote{The difference in parameters before and after fine-tuning is referred to as a task vector, and arithmetic operations on these vectors are effective~\cite{ilharco2023editing}.}. They proposed a method to mitigate interference when integrating task vectors from different models. Specifically, by reducing redundant numerical components and adjusting for conflicting signs, their approach enables effective model merging.

\paragraph{Comparison of activations.}
Comparisons of LLMs based on activations have also been studied.  
\citet{zhou-etal-2025-linguistic} quantified the similarity between LLMs by measuring the cosine similarity of activation differences for linguistic minimal pairs.
In particular, they used datasets such as BLiMP~\cite{10.1162/tacl_a_00321} and showed that model similarity is significantly influenced by the pre-training dataset.

\paragraph{Comparison of model outputs.}
Several approaches compare LLMs based on their outputs.
\citet{lv-etal-2023-parameter} proposed a method for computing coefficients in parameter ensembling by providing the same input text to two models and comparing the softmax probability distributions at each token. 
Specifically, they used KL divergence and summed the results to derive appropriate coefficients.
Furthermore, \citet{yax2025phylolm} proposed a similarity metric based on the conditional probabilities of LLMs and introduced a method for calculating the phylogenetic distance between different models.
\citet{zhuang2025embedllm} propose a framework for vector representations of language models based on their performance on downstream tasks.
Additionally, a method has been proposed for measuring differences in conditional probabilities based on prompts using KL divergence~\cite{melamed-etal-2024-prompts}.

\section{Double Centering} \label{sec:theory-double-centering}
We confirm the notation and computational operations.  
The matrices $\bm{L}$, $\bm{\Xi}$, and $\bmQ$ are all of size $K \times N$, and their elements are denoted by $\ell_{is}, \xi_{is}, q_{is}$, respectively. In particular, we have $\ell_{is} = \ell_i(x_s)$.  
First, row-wise centering of $\bm{L}$ is performed by subtracting the mean log-likelihood $\bar\ell_i$ of each model from each row $(\ell_{i1},\ldots,\ell_{iN})$, resulting in $\bm{\Xi}=(\bmxi_1,\ldots,\bmxi_K)^\top$.  
Next, column-wise centering of $\bm{\Xi}$ is performed by subtracting the coordinate component $\bar \xi_s$ of the mean vector $\bar\bmxi$ from each column $(\xi_{1s},\ldots,\xi_{Ks})^\top$, yielding $\bmQ=(\bmq_1,\ldots,\bmq_K)^\top$.  
Thus, this process involves \emph{double centering}, where column-wise centering follows row-wise centering.  
Notably, even after column-wise centering, the row-wise mean of $\bmQ$ remains zero:
\begin{align}
\frac{1}{N}\sum_{s=1}^N q_{is} &= \frac{1}{N}\sum_{s=1}^{N} (\xi_{is} - \bar \xi_{s}) \nonumber \\
&=\frac{1}{N}\sum_{s=1}^{N} \xi_{is} - \frac{1}{NK}\sum_{i=1}^{K}\sum_{s=1}^{N} \xi_{is}  \nonumber \\
&=0-0=0. \label{eq:q-zero-mean}
\end{align}

The column-wise centering can be interpreted as follows.  
In the $\bmxi$-coordinate system, the mean vector $\bar\bmxi$ of the $K$ model coordinates $\bmxi_1,\ldots,\bmxi_K$ can be regarded as representing an ``average model.''  
By redefining this average model as the new origin, we obtain the $\bmq$-coordinate system.  
From the definition $\bmq_i = \bmxi_i - \bar\bmxi$, its mean satisfies  
\[
\sum_{i=1}^K \bmq_i /K= \bmzero.
\]

The row-wise centering can be interpreted as follows.  
Let $\bmone_N = (1,\ldots,1)^\top \in \bbR^N$. Since $\bar \ell_i = \bmone_N^\top \bmell_i/N$, we have $\bmxi_i = \bmell_i - \bar\ell_i \bmone_N$.  
Thus,  
\[
\bmone_N^\top \bmxi_i = \bmone_N^\top \bmell_i -\bar\ell_i N = 0,
\]
and furthermore, $\bmone_N^\top \bar\bmxi = 0$.  
From this, equation (\ref{eq:q-zero-mean}) is actually trivial, as  
\[
\bmone_N^\top \bmq_i = \bmone^\top (\bmxi_i - \bar\bmxi) = 0-0=0.
\]
The row-wise centering implies that $\bmxi_1,\ldots,\bmxi_K$ and $\bmq_1,\ldots,\bmq_K$ lie in the subspace orthogonal to $\bmone_N$.

\section{Effect of Errors in Model Coordinates} \label{sec:error-analysis}

We analyze the impact of additive errors $\epsilon_{is}$ in the log-likelihood vector components $\ell_{is}$ for $i=1,\ldots,K$ and $s=1,\ldots,N$.  
Denoting the true values with an asterisk as $\ell_{is}^*$, the observed values can be expressed as  
\[
\ell_{is} = \ell_{is}^* + \epsilon_{is}.
\]
We decompose the error as follows:
\[
\epsilon_{is} = a + b_i + c_s + d_{is},
\]
where we assume, without loss of generality, the constraints  
\[
\sum_{i=1}^K b_i =\sum_{s=1}^N c_s = \sum_{i=1}^K d_{is} = \sum_{s=1}^N d_{is} = 0.
\]
Here, $a$, $b_i$, and $c_s$ are bias terms, while $d_{is}$ represents interaction terms.  
Using simple calculations, we obtain:
\[
\bar \ell_i = \frac{1}{N}\sum_{s=1}^N \ell_{is} = \bar \ell_i^* + a + b_i,
\]
\[
\xi_{is} = \ell_{is} - \bar \ell_i = \xi_{is}^* + c_s + d_{is},
\]
\[
\bar \xi_s = \frac{1}{K} \sum_{i=1}^K \xi_{is} = \bar\xi_s^* + c_s,
\]
\[
q_{is} = \xi_{is} - \bar\xi_s = q_{is}^* + d_{is}.
\]
Additionally, for the differences between two models, which are crucial for the model map, we obtain:
\[
\ell_{is} - \ell_{js} = \ell_{is}^* - \ell_{js}^* + b_i - b_j + d_{is} - d_{js},
\]
\[
\xi_{is} - \xi_{js} = \xi_{is}^* - \xi_{js}^* + d_{is} - d_{js},
\]
\[
q_{is} - q_{js} = q_{is}^* - q_{js}^* + d_{is} - d_{js}.
\]
Thus, the terms affected by the error are $\bar\ell_i$, which is influenced by $a + b_i$, and $\ell_{is} - \ell_{js}$, which is affected by $b_i + d_{is}$, meaning it is influenced by the bias terms. However, in the centered values $\xi_{is} - \xi_{js}$ and $q_{is} - q_{js}$, only the interaction term $d_{is}$ contributes to the error.

\section{Theory of Model Coordinates for Text Probability Distributions}\label{sec:theory-text}

In this section, we prove the main results of Section~\ref{sec:model-text}, namely (\ref{eq:main-KL-var-model}) and (\ref{eq:main-KL-xi}).  
Our discussion applies not only to text probability distributions but also more generally to any setting where i.i.d.\ observations \( x_1, \dots, x_N \sim p_i(x) \) are available.  
Compared to the previous study that proposed model maps~\citep{Shimodaira1993modelmap,shimodaira1998graphical}, we conduct a more precise analysis in this paper.  
Specifically, while the previous study provides only a brief evaluation of the approximation, we present a more transparent discussion based on the properties of the exponential family of distributions.
In the next section, as the starting point of our discussion, we construct a \emph{super model} that includes the \( K \) models \( p_i \), \( i=1,\dots,K \), as submodels.  
This model is introduced as a mathematical tool to rigorously prove the main theorem of this paper, and we do not compute it numerically in practice.

\subsection{Exponential family of distributions} \label{sec:theory-exponential-family-setup}

We first consider a model in the exponential family of distributions parameterized by a $K$-dimensional parameter $\bmt \in \bbR^K$:
\begin{align}\label{eq:exponential-family}
    p(x;\bmt) = p_0(x)\exp(\bmt^\top \bmb(x) - \psi(\bmt)).
\end{align}
Here, the function $\bmb(x) = (b_1(x), \dots, b_K(x))^\top$ will be defined later using the $K$ models.  
The normalization constant is given by  
\begin{align*}
    Z(\bmt) &= \sum_{x\in \cX} p_0(x)\exp(\bmt^\top \bmb(x)), \\
    \psi(\bmt) &= \log Z(\bmt),
\end{align*}
which ensures that $\sum_{x\in\cX} p(x;\bmt) = 1$.  
For $\bmt = \bmzero$, where $\bmzero = (0, \dots, 0)^\top$, we obtain  
\[
p(x;\bmzero) = p_0(x).
\]

To associate the $K$ models with (\ref{eq:exponential-family}), we define $\bmb(x)$. For a constant $\lambda > 0$, we set  
\begin{align} \label{eq:bix}
    \lambda b_i(x) := \ell_i(x) - \ell_0(x).
\end{align}
The constant $\lambda$ is an order parameter introduced for theoretical convenience, and in our theoretical framework, we assume that $b_i(x)$ is of constant order and that $\lambda$ is sufficiently small\footnote{If numerical computation were to be performed, $\lambda$ could be set to any arbitrary value (e.g., $\lambda=1$).}.  
Thus, we essentially assume that $|\ell_i(x) - \ell_0(x)| = O_p(\lambda)$ is sufficiently small, implying that each model $p_i$ provides a good approximation of the true generative model $p_0$.  
In the proof of the main theorem, we consider the asymptotic theory as $\lambda \to 0$, retaining terms up to $O(\lambda^2)$ while ignoring those of $O(\lambda^3)$.

A one-hot vector is defined as $\bme_i = (0,\ldots,0,1,0,\ldots,0)^\top \in \bbR^K$ for $i=1,\ldots,K$, where only the $i$-th element is 1.  
Then, setting $\bmt = \lambda \bme_i$ gives  
\begin{align} \label{eq:pix}
  p(x;\lambda \bme_i) = p_i(x).
\end{align}
Indeed, substituting (\ref{eq:bix}) into (\ref{eq:exponential-family}) yields  
\begin{align*}
&    p(x;\lambda \bme_i) \\
=&p_0(x)\exp(\lambda \bme_i^\top \bmb(x)-\psi(\lambda \bme_i))\\
=&p_0(x)\exp(\ell_i(x)-\ell_0(x) - \psi(\lambda \bme_i))\\
=&p_0(x) (p_i(x)/p_0(x)) \exp(- \psi(\lambda \bme_i))\\
=&p_i(x)\exp(- \psi(\lambda \bme_i))\\
=&p_i(x),
\end{align*}
where $\psi(\lambda \bme_i) = 0$.

\subsection{Properties of the exponential family of distributions}

This section outlines some well-known basic properties of the exponential family of distributions, which have been established in the literature~\citep{barndorff2014information, efron1978geometry, efron2022exponential, amari1982-10.1214/aos/1176345779}.  
We define the expectation and covariance matrix of $\bmb(x)$ as follows:
\begin{align*}
    \bmet(\bmt):=& \bbE_{x\sim p(\bmt)}\bigl(\bmb(x)\bigr)
    =\sum_{x\in\cX} \bmb(x) p(x;\bmt),\\
    G(\bmt):=& \bbE_{x\sim p(\bmt)}
    \bigl\{ (\bmb(x)- \bmet(\bmt)) (\bmb(x)- \bmet(\bmt))^\top\bigr\}\\
    =&\Var_{x\sim p(\bmt)}\bigl(\bmb(x)\bigr).
\end{align*}
Here, the elements of $\bmet(\bmt)$ are expectations given by $\bbE_{x\sim p(\bmt)}(b_i(x))$, and the elements of $G(\bmt)$ are covariances given by $G_{ij}(\bmt) = \Cov_{x\sim p(\bmt)}(b_i(x), b_j(x))$.
These quantities can be expressed in terms of $\psi(\bmt)$ as follows:
\begin{align}
 \bmet(\bmt) =     \frac{\partial\psi(\bmt)}{\partial \bmt}, \label{eq:psi-eta}\\
 G(\bmt) =    \frac{\partial^2\psi(\bmt)}{\partial \bmt \partial \bmt^\top}.  \label{eq:psi-G}
\end{align}
We now derive these two equations. First, from
\[
\frac{\partial Z(\bmt)}{\partial \bmt}
= \sum_{x\in\cX} \bmb(x)p_0(x)e^{\bmt^\top \bmb(x)},
\]
we obtain
\begin{align*}
\frac{\partial\psi(\bmt)}{\partial \bmt}
=&\frac{\partial \log Z}{\partial \bmt}
=\frac{1}{Z(\bmt)}\frac{\partial Z}{\partial \bmt}\\
=& \frac{1}{Z(\bmt)} \sum_{x\in\cX}
\bmb(x) p_0(x) e^{\bmt^\top \bmb(x)}\\
=&\sum_{x\in\cX} \bmb(x)p(x;\bmt) = \bmet(\bmt).
\end{align*}
Thus, we have established (\ref{eq:psi-eta}).  
Next, using
\begin{align*}
\frac{\partial p(x;\bmt)}{\partial \bmt}
=&\Bigl(\bmb(x)-\frac{\partial \psi}{\partial \bmt}
\Bigr) p(x;\bmt)\\
=&(\bmb(x)-\bmet(\bmt)) p(x;\bmt),
\end{align*}
we obtain
\begin{align*}
&\frac{\partial^2 \psi(\bmt)}
{\partial \bmt \partial \bmt^\top}
=\frac{\partial \bmet(\bmt)^\top}{\partial \bmt}\\
=&\frac{\partial}{\partial\bmt}
\sum_{x\in\cX} \bmb(x)^\top p(x;\bmt)\\
=&\sum_{x\in\cX}(\bmb(x)-\bmet(\bmt)) \bmb(x)^\top p(x;\bmt)\\
=&\sum_{x\in\cX}(\bmb(x)-\bmet(\bmt)) (\bmb(x)-\bmet(\bmt))^\top p(x;\bmt)\\
=&G(\bmt).
\end{align*}
Thus, we have established (\ref{eq:psi-G}).

\subsection{Approximation of the KL divergence} \label{sec:theory-approximate-KL}

The Kullback-Leibler (KL) divergence between the models $p(\bmt)$ and $p(\bmt')$ at parameter values $\bmt, \bmt' \in \bbR^K$ is given by  
\begin{align}
 &  \mathrm{KL}(p(\bmt), p(\bmt')) \nonumber\\
=& \sum_{x\in\cX}p(x;\bmt) \log\frac{p(x;\bmt)}{p(x;\bmt')} \nonumber\\
=& \sum_{x\in\cX} p(x;\bmt) \bigl\{
(\bmt-\bmt')^\top \bmb(x) - \psi(\bmt) + \psi(\bmt')
\bigr\}\nonumber\\
=&(\bmt-\bmt')^\top \bmet(\bmt) - \psi(\bmt) + \psi(\bmt').
\label{eq:KL-exponential-family}
\end{align}
Here, we assume that the parameter values $\bmt$ and $\bmt'$ are sufficiently close to $\bmzero$.  
In particular, we assume $\|\bmt\| = O(\lambda)$ and $\|\bmt'\| = O(\lambda)$.  
Substituting (\ref{eq:psi-eta}) and (\ref{eq:psi-G}) into the Taylor expansion of $\psi(\bmt)$ gives  
\begin{align}
 &\psi(\bmt') \nonumber \\
=&\psi(\bmt)
+\frac{\partial \psi}{\partial \bmt^\top}
(\bmt'-\bmt)\nonumber \\
&+\frac{1}{2}(\bmt'-\bmt)^\top
\frac{\partial^2 \psi(\bmt)}
{\partial \bmt \partial \bmt^\top}
(\bmt'-\bmt)\nonumber \\
& +O(\|\bmt'-\bmt\|^3)\nonumber\\
=&\psi(\bmt) + \bmet(\bmt)^\top (\bmt'-\bmt)\nonumber\\
&+\frac{1}{2}(\bmt'-\bmt)^\top G(\bmt)(\bmt'-\bmt)+O(\lambda^3).
\label{eq:psi-taylor}
\end{align}
Substituting (\ref{eq:psi-taylor}) into (\ref{eq:KL-exponential-family}) gives  
\begin{align}
&\mathrm{KL}(p(\bmt),p(\bmt'))\nonumber \\
&=\frac{1}{2}(\bmt'-\bmt)^\top G(\bmt)(\bmt'-\bmt)+O(\lambda^3).
\label{eq:KL-approx-exponential-family-theta}
\end{align}
This corresponds to eq.~(9) of \citet{oyama-etal-2023-norm}.  
Here, the equation holds approximately by ignoring higher-order terms of $O(\lambda^3)$.  
For more details, refer to \citet[p.~369]{amari1982-10.1214/aos/1176345779} and \citet[p.~35]{efron2022exponential}.  
More generally, $G(\bmt)$ represents the Fisher information metric, and \eqref{eq:KL-approx-exponential-family-theta} holds for a wide class of probability models~\cite{amari1998natural} with $\|\bmt' - \bmt\|=O(\lambda)$.  
Furthermore, since $G(\bmt) = G(\bmzero) + O(\|\bmt\|) = G(\bmzero) + O(\lambda)$, we obtain  
\begin{align}
&\mathrm{KL}(p(\bmt),p(\bmt'))\nonumber\\
&=\frac{1}{2}(\bmt'-\bmt)^\top G(\bmzero)(\bmt'-\bmt)+O(\lambda^3).
\label{eq:KL-approx-exponential-family}
\end{align}

\subsection{The variance representation of the KL divergence} \label{sec:theory-variance-KL}

Substituting $p_i = p(\lambda\bme_i)$ into (\ref{eq:KL-approx-exponential-family}) gives  
\begin{align}
&2\mathrm{KL}(p_i,p_j)
=2\mathrm{KL}(p(\lambda\bme_i), p(\lambda\bme_j))
\nonumber\\
&=\lambda^2(\bme_i-\bme_j)^\top G(\bmzero)(\bme_i-\bme_j)+O(\lambda^3).
\label{eq:KL-approx-model}
\end{align}
Here, we have  
\begin{align}
&\bme_i^\top G(\bmzero) \bme_j = G_{ij}(\bmzero)\nonumber\\
&=\bbE_{x\sim p_0}\Bigl\{
(b_i(x)-\eta_i(\bmzero))(b_j(x) - \eta_j(\bmzero))
\Bigr\}.
\label{eq:eiGej}
\end{align}
Next, we substitute (\ref{eq:eiGej}) and (\ref{eq:bix}) into the right-hand side of (\ref{eq:KL-approx-model}) to derive an alternative expression for the KL divergence:
\begin{align*}
&\lambda^2 (\bme_i-\bme_j)^\top G(\bmzero)(\bme_i-\bme_j)\\
=&\lambda^2\bbE_{x\sim p_0}
\Bigl[
\Bigl\{(b_i(x)-\eta_i(\bmzero))\\
&\qquad -(b_j(x) - \eta_j(\bmzero))\Bigr\}^2
\Bigr]\\
=&\bbE_{x\sim p_0}
\Bigl[\Bigl\{
(\ell_i(x)-\ell_0(x))-(\ell_j(x)-\ell_0(x)) \\
&\qquad - \bbE_{x'\sim p_0}\Bigl(
(\ell_i(x')-\ell_0(x'))\\
&\qquad\qquad -(\ell_j(x')-\ell_0(x'))\Bigr)
\Bigr\}^2\Bigr]\\
=&\bbE_{x\sim p_0}
\Bigl[\Bigl\{
\ell_i(x)-\ell_j(x) \\
&\qquad - \bbE_{x'\sim p_0}\Bigl(
\ell_i(x')-\ell_j(x')\Bigr)
\Bigr\}^2\Bigr]\\
=&\Var_{x\sim p_0} \Bigl(
\ell_i(x)-\ell_j(x)
\Bigr).
\end{align*}
Finally, substituting this result into (\ref{eq:KL-approx-model}) yields  
\begin{align}
2\mathrm{KL}(p_i,p_j)=
\Var_{x\sim p_0} \Bigl(
\ell_i(x)-\ell_j(x)
\Bigr)+O(\lambda^3).
\label{eq:KL-var-model}
\end{align}
This establishes (\ref{eq:main-KL-var-model}).  
Furthermore, since $|\ell_i(x)-\ell_j(x)|=O_p(\lambda)$, the magnitude of (\ref{eq:KL-var-model}) is $O(\lambda^2)$.

\subsection{Estimation of the KL divergence} \label{sec:theory-estimate-KL}

If the expected value $\bbE_{x\sim p_0}(f(x))$ of a function $f(x)$ exists and is bounded, then by the law of large numbers, the sample mean\footnote{$\bbE_{x\sim D}(f(x)) = \frac{1}{N} \sum_{x\in D} f(x) = \frac{1}{N}\sum_{s=1}^N f(x_s)$ represents the sample mean, and $\Var_{x\sim D}(\cdot)$ represents the sample variance.} converges to the expected value as $N \to \infty$, and we have  
\[
\bbE_{x\sim D}(f(x)) = \bbE_{x\sim p_0}(f(x)) + O_p(N^{-1/2}).
\]
Applying this to (\ref{eq:KL-var-model}) and ignoring the terms of order $O_p(\lambda^3 + \lambda^2 N^{-1/2})$, we obtain the following approximation:
\begin{align}
2\mathrm{KL}(p_i,p_j)\approx
\Var_{x\sim D} \Bigl(
\ell_i(x)-\ell_j(x)
\Bigr).
\label{eq:KL-data-var-model}
\end{align}
We define the coordinates $\bmxi_i\in \bbR^N$ of model $p_i$ as  
\[
\bmxi_i=(\xi_{i1},\ldots,\xi_{iN})^\top
\]
with  
\begin{align*}
\xi_{is}:= \ell_i(x_s) - \bbE_{x\sim D} ( \ell_i(x))
\end{align*}
for $s=1,\ldots,N$.  
From (\ref{eq:KL-data-var-model}), we obtain  
\begin{align}
2\mathrm{KL}(p_i,p_j) \approx&
\frac{1}{N}\sum_{s=1}^N (\xi_{is}-\xi_{js})^2 \nonumber\\
=&
\frac{1}{N}\|\bmxi_i - \bmxi_j\|^2.
\label{eq:KL-xi}
\end{align}
Since  
\[
\|\bmxi_i - \bmxi_j\|^2=\|(\bmq_i + \bar\bmxi) - (\bmq_j + \bar\bmxi)\|^2 = \|\bmq_i - \bmq_j\|^2,
\]
this establishes (\ref{eq:main-KL-xi}).

\subsection{Relationships among the three types of model coordinates}\label{sec:theory-height}

Let $\bmone_N = (1,\ldots,1)^\top \in \bbR^N$.  
From the definitions of the $\bmxi$-coordinate system and the $\bmq$-coordinate system, we have  
\begin{align}
    \bmxi_i &= \bmq_i + \bar\bmxi, \nonumber\\
    \bmell_i &= \bmxi_i + \bar\ell_i \bmone_N \nonumber\\
    &= \bmq_i + \bar\ell_i \bmone_N + \bar\bmxi.
    \label{eq:ell-q-transform}
\end{align}
Additionally, equation (\ref{eq:q-zero-mean}) in Appendix~\ref{sec:theory-double-centering} can be rewritten as  
\begin{align}
    \bmone_N^\top \bmq_i = 0.
    \label{eq:q-centered}
\end{align}
Thus, we obtain  
\begin{align*}
 &    \|\bmell_i - \bmell_j\|^2\\
=& \| (\bmq_i - \bmq_j ) + (\bar\ell_i - \bar\ell_j)\bmone_N \|^2\\
=& \|\bmq_i - \bmq_j\|^2 + N (\bar\ell_i - \bar\ell_j)^2\\
& \quad +2(\bar\ell_i - \bar\ell_j)\bmone_N^\top (\bmq_i - \bmq_j)\\
=& \|\bmq_i - \bmq_j\|^2 + N (\bar\ell_i - \bar\ell_j)^2,
\end{align*}
where (\ref{eq:ell-q-transform}) and (\ref{eq:q-centered}) are used in the first and last equations, respectively.  
This establishes (\ref{eq:lik-decomposition}).  

Moreover, since $\bar\ell_i = \bmone_N^\top \bmell_i/N$, it is straightforward that the component of the $\bmell$-coordinate system in the $\bmone_N$ direction is given by  
\[
(\bmone_N/\sqrt{N})^\top \bmell_i = \sqrt{N} \bar \ell_i.
\]

\subsection{Additional notes on modifying the underlying generative model}\label{sec:theory-note-modifyp0}

We examine the effect of changing the true generative model \(p_0 = p(\bmzero)\) that produces the data \(D\).  
For clarity, we continue to use the same \(p_0\) as before in constructing \(p(x;\bmt)\) as described in Section~\ref{sec:theory-exponential-family-setup}.  
We then introduce a new parameter value \(\bmt^*\).  
We assume that each element \(x_s\) of the dataset \(D\) is generated independently from  
\begin{align}
    x_1,\ldots,x_N \sim p(x;\bmt^*),
    \label{eq:data-generation-star}
\end{align}
and that \(\|\bmt^*\| = O(\lambda)\).  
In other words, the true generative model remains sufficiently close to the \(K\) models, satisfying \(\|\bmt_i - \bmt^*\| = O(\lambda)\) for \(i = 1,\ldots,K\).  

First, consider replacing \(G(\bmt)\) in \eqref{eq:KL-approx-exponential-family-theta} of Section~\ref{sec:theory-approximate-KL} with \(G(\bmt^*)\).  
Because \(G(\bmt^*) = G(\bmt) + O(\lambda)\), we obtain  
\begin{align}
&\mathrm{KL}(p(\bmt),p(\bmt'))\nonumber\\
&=\frac{1}{2}(\bmt' - \bmt)^\top G(\bmt^*)(\bmt' - \bmt) + O(\lambda^3).
\label{eq:KL-approx-exponential-family-star}
\end{align}
We are generalizing the discussion in the previous sections, and indeed, if we set \(\bmt^* = \bmzero\) in \eqref{eq:KL-approx-exponential-family-star}, then \eqref{eq:KL-approx-exponential-family} is recovered.  

Now substitute \(p_i = p(\lambda \bme_i)\) into \eqref{eq:KL-approx-exponential-family-star}, yielding  
\begin{align}
&2\,\mathrm{KL}(p_i,p_j)
=2\,\mathrm{KL}\bigl(p(\lambda\bme_i), p(\lambda\bme_j)\bigr)
\nonumber\\
&=\lambda^2(\bme_i - \bme_j)^\top G(\bmt^*)(\bme_i - \bme_j) + O(\lambda^3),
\label{eq:KL-approx-model-star}
\end{align}
which is a generalization of \eqref{eq:KL-approx-model} in Section~\ref{sec:theory-variance-KL}.  
Using the definition of \(\bm{G}(\bmt)\), we have  
\begin{align}
\bme_i^\top G(\bmt^*) \bme_j 
= G_{ij}(\bmt^*) 
= \Cov_{x\sim p(\bmt^*)}\bigl(b_i(x), b_j(x)\bigr).
\label{eq:eiGej-star}
\end{align}
Substituting \eqref{eq:eiGej-star} and \eqref{eq:bix} into the right-hand side of \eqref{eq:KL-approx-model-star} yields  
\begin{align*}
&\lambda^2(\bme_i - \bme_j)^\top G(\bmt^*)(\bme_i - \bme_j)\\
=&\,\lambda^2\Bigl\{
\Var_{x\sim p(\bmt^*)}\bigl(b_i(x)\bigr)
+\Var_{x\sim p(\bmt^*)}\bigl(b_j(x)\bigr)
\\
&\qquad -2\,\Cov_{x\sim p(\bmt^*)}\bigl(b_i(x), b_j(x)\bigr)
\Bigr\}
\\
=&\,\lambda^2 \Var_{x\sim p(\bmt^*)}\bigl(b_i(x) - b_j(x)\bigr)
\\
=&\,\Var_{x\sim p(\bmt^*)}\bigl(\ell_i(x) - \ell_j(x)\bigr).
\end{align*}
Finally, substituting this back into \eqref{eq:KL-approx-model-star} gives  
\begin{align}
2\,\mathrm{KL}(p_i,p_j)=
\Var_{x\sim p(\bmt^*)}\bigl(\ell_i(x) - \ell_j(x)\bigr)+O(\lambda^3),
\label{eq:KL-var-model-star}
\end{align}
which is a generalization of \eqref{eq:main-KL-var-model}.  

Hence, the estimators for KL divergence in Section~\ref{sec:theory-estimate-KL}, specifically \eqref{eq:KL-data-var-model} and \eqref{eq:KL-xi}, and also \eqref{eq:main-KL-xi} in Section~\ref{sec:model-text}, remain valid even when the texts are generated by \eqref{eq:data-generation-star}.  
Since \eqref{eq:KL-var-model-star} holds for any \(\bmt^*\) with \(\|\bmt^*\| = O(\lambda)\), this result also applies when the true data-generating model is \(p(\bmt^*) = p_0\), or, for instance, one of the models \(p(\bmt^*) = p_i\), or a mixture of the \(K\) models, \(\bmt^* = \sum_{i=1}^K \alpha_i \lambda \bme_i\) with \(\alpha_i = O(1)\).  
Therefore, this method is robust to contamination in the dataset \(D\) (e.g., when the text corpus used for pre-training a model is included in \(D\)), as the estimation of KL divergence via the squared Euclidean distance in the \(\bmxi\)-coordinate system or the \(\bmq\)-coordinate system remains relatively unaffected.

\section{Mapping Language Models into the Space of Token Probability Distributions} \label{sec:model-token}

In Section~\ref{sec:model-text}, we discussed model maps based on the probability distributions $p_i(x)$ of texts generated by language models.  
This approach requires computing probabilities for a large number of texts in the dataset $D = (x_1,\ldots,x_N)$, leading to high computational costs.  
To mitigate this issue, we focus on the fact that a text $x = (y_1,\ldots,y_n)$ is a sequence of tokens.  
Instead of using text probabilities, we discuss model maps based on the conditional probability distributions of token generation, $p_i(y_t|y^{t-1})$.  
In this approach, model coordinates are computed using only a single text $x$.  
A limitation of this approach is that it can only be used for comparing models that share the same tokenizer.  
Furthermore, the current estimation method ignores the variance of the expected log-likelihood ratio of conditional probabilities, resulting in a rough approximation.  
Thus, the estimated values should be regarded only as reference values rather than precise measurements.

\subsection{Model coordinates} \label{sec:token-logp}

For a text $x = (y_1,\ldots,y_n)$, the coordinates of model $p_i$  
\[
\bmzeta_i = (\zeta_{i1},\ldots,\zeta_{in})^\top \in \bbR^n
\]
are defined as  
\begin{align}
\zeta_{it} := \log p_i(y_t|y^{t-1}) - \ell_i(x)/n
\label{eq:zeta-coordinate}
\end{align}
for $t=1,\ldots,n$.  
This is centered for each $i$ and for each text, satisfying $\sum_{t=1}^n \zeta_{it} = 0$.

\subsection{Kullback-Leibler divergence}

The KL divergence for next-token generation in language models, where $y_t \sim p_i(y_t|y^{t-1})$, is given by  
\begin{align}
&\mathrm{KL}(p_i(y_t|y^{t-1}) ,p_j(y_t|y^{t-1}) ) =\\
&\qquad \sum_{y_t\in\cV} p_i(y_t|y^{t-1})\log\frac{p_i(y_t|y^{t-1})}{p_j(y_t|y^{t-1})}.
\label{eq:token-KL-def}
\end{align}
We apply the results for text probability distributions from Section~\ref{sec:model-text} and Appendix~\ref{sec:theory-text} to the conditional probability distributions of token generation.  
The equation corresponding to (\ref{eq:main-KL-var-model}) is  
\begin{align}
&2\mathrm{KL}(p_i(y_t|y^{t-1}) ,p_j(y_t|y^{t-1}) )\approx \nonumber \\
&\Var_{y_t\sim p_0(y_t|y^{t-1})} \biggl\{
\log\frac{p_i(y_t|y^{t-1})}{p_j(y_t|y^{t-1})}  
\biggr\}.
\label{eq:main-token-KL-var-model}
\end{align}
The squared Euclidean distance in the $\bmzeta$-coordinate system provides an estimate of the sum of (\ref{eq:main-token-KL-var-model}) over all tokens in the text $x$:  
\begin{align}
&\|\bmzeta_i - \bmzeta_j\|^2 \nonumber\\
\approx& 2\sum_{t=1}^n\mathrm{KL}(p_i(y_t|y^{t-1}),p_j(y_t|y^{t-1})) 
\label{eq:main-token-KL-xi}\\
\approx&2\mathrm{KL}(p_i,p_j).
\label{eq:main-token-KL-text-KL-xi}
\end{align}
The proof is provided in Appendix~\ref{sec:theory-token}.  
To justify the estimation in (\ref{eq:main-token-KL-xi}), we assume the following:  
\begin{align}
\bbE_{y_t\sim p_0(y_t|y^{t-1})}\biggl\{ \log \frac{p_i(y_t|y^{t-1})}{p_j(y_t|y^{t-1})}\biggr\}
\label{eq:main-token-assumtion2}
\end{align}
takes a constant value independent of $t$.
In reality, this assumption is not entirely correct, and the degree of variation affects the accuracy of the approximation in (\ref{eq:main-token-KL-xi})\footnote{Since this assumption does not affect (\ref{eq:main-KL-xi}), there is no concern regarding the use of model maps based on text probabilities.}.  
On the other hand, the approximation in (\ref{eq:main-token-KL-text-KL-xi}) holds more generally and is demonstrated in Appendix~\ref{sec:token-text-KL-relation}.

\section{Theory of Model Coordinates for Token Probability Distributions} \label{sec:theory-token}

In this section, we provide a more detailed explanation of the content discussed in Appendix~\ref{sec:model-token}.  
We extend the discussion of text probability distributions in Appendix~\ref{sec:theory-text} to the case of conditional probability distributions for token generation.

 \subsection{Exponential family of distributions}

We apply the same setting as for $p_i(x)$ in Section~\ref{sec:theory-text} to the conditional probability distributions of tokens:  
\[
y_t \sim p_i(y_t|y^{t-1}),\quad t=1,\ldots,n.
\]
The exponential family of distributions incorporating $K$ models, corresponding to (\ref{eq:exponential-family}), is given here as  
\begin{align}
\label{eq:token-exponential-family}
&p(y_t|y^{t-1};\bmt) := p_0(y_t|y^{t-1})\nonumber\\
&\qquad\exp(\bmt^\top \bmb(y_t|y^{t-1}) - \psi(\bmt |y^{t-1})).
\end{align}

The setting (\ref{eq:bix}), which associates the $K$ models with (\ref{eq:token-exponential-family}), is given here as  
\begin{align} \label{eq:token-bix}
    &\lambda b_i(y_t|y^{t-1}) :=\nonumber \\
    &\quad \log p_i(y_t|y^{t-1}) - \log 
 p_0(y_t|y^{t-1}).
\end{align}
Thus, we have
\[
p_i(y_t|y^{t-1}) = p(y_t|y^{t-1}; \lambda \bme_i)
\]
for $i=1,\ldots,K$.

\subsection{The variance representation of the KL divergence}

The KL divergence is given by (\ref{eq:token-KL-def}).  
Applying the result for the model $p(x;\bmt)$ in (\ref{eq:KL-var-model}) to the token-level conditional distribution model $p(y_t|y^{t-1};\bmt)$, we obtain  
\begin{align}
&2\mathrm{KL}(p_i(y_t|y^{t-1}) ,p_j(y_t|y^{t-1}) )=\nonumber \\
&\Var_{y_t\sim p_0(y_t|y^{t-1})} \biggl\{
\log\frac{p_i(y_t|y^{t-1})}{p_j(y_t|y^{t-1})}  
\biggr\}+O(\lambda^3).
\label{eq:token-KL-var-model}
\end{align}

\subsection{Two additional assumptions}\label{sec:token-theory-assumptions}

To estimate the KL divergence from a single text $x$, two additional assumptions are required, as described below.  
Such assumptions were not necessary when estimating the KL divergence from the dataset $D$ in Appendix~\ref{sec:theory-text}.  
In reality, these two assumptions are not strictly satisfied, and the discrepancy between these assumptions and reality affects the accuracy of the KL divergence approximation.

\paragraph{Assumption 1:}
We assume that the probability distribution of $y_t$ depends only on the past $k$ tokens, denoted as $y^{t-1}_{t-k} = (y_{t-k},y_{t-k+1},\ldots,y_{t-1})$.  
That is,  
\[
p_i(y_t|y^{t-1}) = p_i(y_t|y^{t-1}_{t-k}),
\]
which allows us to regard $y^t_{t-k}$ as the state of a Markov chain. More generally, we use the notation $y^k$ to represent a state variable.
We consider a function $f$ of the state variable $y^k$.  
Furthermore, we assume that this Markov chain is positive Harris recurrent, has a stationary distribution $\pi$, and that $f$ is absolutely integrable, i.e.,  
\[
\bbE_{y^k\sim \pi}(|f(y^k)|)<\infty.
\]
Then, by the strong law of large numbers for Markov chains~\cite[Theorem~17.0.1~(i)]{Meyn_Tweedie_Glynn_2009}, in the limit $n\to\infty$,  
\[
\frac{1}{n}\sum_{t=1}^n f(y^t_{t-k})\to \bbE_{y^k\sim \pi}(f(y^k))\quad\text{a.s.}
\]
For simplicity in notation and discussion, we assume that $y_{-k+1},\ldots,y_0$ are appropriately defined.
Since the Markov chain converges to $\pi$, we also have  
\begin{align}\label{eq:lln-markov}
\frac{1}{n}\sum_{t=1}^n f(y^t_{t-k}) \to
\frac{1}{n}\sum_{t=1}^n \bbE_{y^t \sim p_0}(f(y^t_{t-k}))
\end{align}
almost surely as $n\to\infty$.

\paragraph{Assumption 2:}
\begin{align}
\bbE_{y_t\sim p_0(y_t|y^{t-1})} \log \frac{p_i(y_t|y^{t-1})}{p_j(y_t|y^{t-1})} = c
\label{eq:token-assumtion2}
\end{align}
for some $c\in\bbR$ that can depend on the indices $i$ and $j$ but not on $t$.  
In other words, (\ref{eq:token-assumtion2}) takes a constant value independent of $t$.

\subsection{Estimation of the KL divergence}

Define  
\[
h(y^t) := \log\frac{p_i(y_t|y^{t-1})}{p_j(y_t|y^{t-1})}.
\]
From Assumption 1, $h(y^t)$ can be written in the form $h(y^t) = f_1(y^t_{t-k})$ for some $f_1$, so applying (\ref{eq:lln-markov}), for sufficiently large $n$, we obtain  
\[
\frac{1}{n}\sum_{t=1}^n h(y^t) \approx \frac{1}{n}\sum_{t=1}^n \bbE_{y^t\sim p_0}(h(y^t)).
\]
Applying (\ref{eq:token-assumtion2}) to the right-hand side gives  
\[
\frac{1}{n}\sum_{t=1}^n h(y^t) \approx c.
\]
Next, since $(h(y^t) - c )^2$ can be written in the form of $f_2(y^t_{t-k})$ for some function $f_2$, applying (\ref{eq:lln-markov}) again yields  
\begin{align*}
&  \frac{1}{n} \sum_{t=1}^n (h(y^t) - c)^2\\
\approx &  \frac{1}{n} \sum_{t=1}^n
\bbE_{y^{t-1}\sim p_0} \Bigl\{
\Var_{y_t \sim p_0(y_t|y^{t-1})} ( h(y^t))  \Bigr\}\\
\approx &\frac{1}{n} \sum_{t=1}^n
\Var_{y_t \sim p_0(y_t|y^{t-1})} ( h(y^t)).
\end{align*}
In the final equation, we applied (\ref{eq:lln-markov}) using the fact that $\Var_{y_t \sim p_0(y_t|y^{t-1})} ( h(y^t)) = f_3(y^t_{t-k})$ for some $f_3$.  
Using (\ref{eq:token-KL-var-model}), we obtain  
\begin{align}
&\sum_{t=1}^n (h(y^t) - c)^2 \approx\nonumber\\
&\qquad
2\sum_{t=1}^n\mathrm{KL}(p_i(y_t|y^{t-1}),p_j(y_t|y^{t-1})).
\label{eq:token-var-KL}
\end{align}
Meanwhile, the components of the model coordinate $\bmzeta_i$ are given by  
\[
\zeta_{it}= \log p_i(y_t|y^{t-1}) - c_i
\]
where  
\[
c_i = \frac{1}{n} \sum_{t=1}^n \log p_i(y_t|y^{t-1}).
\]
Since  
\[
\zeta_{it}-\zeta_{jt} = h(y^t) -(c_i-c_j)
\]
with $c_i-c_j \approx c$, equation (\ref{eq:token-var-KL}) can be rewritten as  
\begin{align*}
&    \|\bmzeta_i - \bmzeta_j\|^2 \approx\\
&\qquad 2\sum_{t=1}^n\mathrm{KL}(p_i(y_t|y^{t-1}),p_j(y_t|y^{t-1})).
\end{align*}
Thus, (\ref{eq:main-token-KL-xi}) is established.

\subsection{Connecting the KL divergence of token and text probability distributions}
\label{sec:token-text-KL-relation}

Here, we fix the sequence length of the text $x = (y_1,\ldots,y_n)$ as $n$, i.e., we set $\cX = \cV^n$.  
For notational simplicity, we define  
\[
g_i(y^t) = \log p_i(y_t|y^{t-1}).
\]
Noting that  
\[
p_i(x) = \prod_{t=1}^n p_i(y_t|y^{t-1})=\prod_{t=1}^n e^{g_i(y^t)},
\]
we obtain  
\begin{align*}
& \textrm{KL}(p_i, p_j)\\
=&\sum_{x\in \cX}\prod_{t'=1}^n e^{g_i(y^{t'})} \sum_{t=1}^n ( g_i(y^t) - g_j(y^t))\\
=& \sum_{t=1}^n \sum_{y^t\in \cV^t} \prod_{t'=1}^t e^{g_i(y^{t'})}  ( g_i(y^t) - g_j(y^t))\\
=& \sum_{t=1}^n \sum_{y^{t-1}\in \cV^{t-1}} \prod_{t'=1}^{t-1} e^{g_i(y^{t'})}\\
&\qquad \sum_{y_t\in \cV} e^{g_i(y^t)}  ( g_i(y^t) - g_j(y^t))\\
=& \sum_{t=1}^n \sum_{y^{t-1}\in \cV^{t-1}} p_i(y^{t-1}) \\
&\qquad \textrm{KL}(p_i(y_t|y^{t-1}),p_j(y_t|y^{t-1}))\\
=& \sum_{t=1}^n \bbE_{y^{t-1} \sim p_i}
\Bigl\{ \textrm{KL}(p_i(y_t|y^{t-1}),p_j(y_t|y^{t-1})) \Bigr\}\\
=&  \bbE_{x \sim p_i}\Bigl\{ \sum_{t=1}^n\textrm{KL}(p_i(y_t|y^{t-1}),p_j(y_t|y^{t-1})) \Bigr\}.
\end{align*}
Thus, for sufficiently large $n$, by the strong law of large numbers for Markov chains, we obtain  
\begin{align}
\textrm{KL}(p_i, p_j)\approx
 \sum_{t=1}^n\textrm{KL}(p_i(y_t|y^{t-1}),p_j(y_t|y^{t-1})).
 \label{eq:KL-text-KL-token}
\end{align}
This corresponds to (\ref{eq:main-token-KL-text-KL-xi}).  
Assumption 1 from Appendix~\ref{sec:token-theory-assumptions} is used in (\ref{eq:KL-text-KL-token}), but Assumption 2 is not needed in the discussion of this subsection.

\section{Details of Experiments} \label{sec:experiment-details}

\subsection{Information obtained via the Hugging Face Hub API}
We used the Hugging Face Hub API to retrieve information about each language model's tags, the date the model was created, the number of downloads over the past 30 days, and the model's configuration details. All of this information is current as of February 1, 2025.

Among the model tags, we specifically used \texttt{llama2}, \texttt{llama-2}, \texttt{license:llama2}, \texttt{llama3}, \texttt{llama-3}, and \texttt{license:llama3} to determine the model type (llama-1, llama-2, or llama-3). Furthermore, to identify language models that were pre-trained on the Pile in Section~\ref{sec:model-map}, we employed tags such as \texttt{dataset:eleutherai/pile}, \texttt{dataset:eleutherai/the\_pile}, \texttt{dataset:eleutherai/the\_pile\_deduplicated}, and \texttt{arxiv:2101.00027}.

\subsection{How the model type was determined}

\begin{table}[t!]
    \centering
    \begin{tabular}{lr}
    \toprule
Model type & Models \\
\midrule
\texttt{llama-1} & 69\\
\texttt{llama-2} & 223\\
\texttt{llama-3} & 62\\
\texttt{llama} & 217\\
\texttt{mistral} & 232\\
\texttt{gpt\_neox} & 54 \\
\texttt{deepseek} & 26 \\
\texttt{gptj} & 19 \\
\texttt{gemma} & 18 \\
\texttt{opt} & 15 \\
\texttt{bloom} & 12 \\
\texttt{falcon} & 11 \\
\texttt{qwen2} & 10 \\
\texttt{mixtral} & 9 \\
\texttt{mpt} & 6 \\
\texttt{stablelm} & 6 \\
\texttt{gpt\_neo} & 3 \\
\texttt{phi} & 3 \\
\texttt{gpt\_bigcode} & 3 \\
\texttt{phi3} & 3 \\
\texttt{xglm} & 3 \\
\texttt{rwkv} & 3 \\
\texttt{starcoder2} & 2 \\
\texttt{olmo} & 2 \\
\texttt{camelidae} & 2 \\
\texttt{codegen} & 2 \\
\texttt{deci} & 1 \\
\texttt{recurrent\_gemma} & 1 \\
\texttt{stablelm\_alpha} & 1 \\
\midrule
Total & 1,018\\
    \bottomrule
    \end{tabular}
    \caption{Number of models by model type.}
    \label{tab:model-type}
\end{table}

In principle, we used the value of \texttt{model\_type} in the config retrieved from the Hugging Face Hub API as the model type. However, out of the 1,018 language models we examined, there were 587 whose config \texttt{model\_type} was listed as \texttt{llama}. For these, we used the following procedure to determine whether they were the original Llama (llama-1), Llama-2, or Llama-3; if we were able to identify which version they were, we reclassified them as \texttt{llama-1}, \texttt{llama-2}, or \texttt{llama-3} accordingly.

\begin{enumerate}
    \item We checked the tags assigned to each model. Of these, 136 models that included any of \texttt{llama2}, \texttt{llama-2}, or \texttt{license:llama2} were classified as \texttt{llama-2}. Similarly, 39 models that included any of \texttt{llama3}, \texttt{llama-3}, or \texttt{license:llama3} were classified as \texttt{llama-3}.
    \item For the remaining 412 models whose classification was not determined by tags alone, we used the creation date and the model name (converted to lowercase) to make a decision. First, 69 models that were created prior to July 18, 2023 (the Llama-2 release date) were classified as \texttt{llama-1}. Next, 88 models whose lowercase model name contained either \texttt{llama2} or \texttt{llama-2} were classified as \texttt{llama-2}. Among those whose lowercase model name contained \texttt{llama3} or \texttt{llama-3}, 22 models whose creation date was after April 18, 2024 (the Llama-3 release date) were classified as \texttt{llama-3}.
    \item After following the steps above, the 217 models that could not be classified were left as \texttt{llama}.
\end{enumerate}

Furthermore, any model whose name prior to the slash (\texttt{/}) was \texttt{deepseek-ai} was defined as \texttt{deepseek}. In addition, even though \texttt{abacusai/Llama-3-Smaug-8B} was tagged with \texttt{license:llama2}, we manually reclassified it as \texttt{llama-3}.

Table~\ref{tab:model-type} shows the number of models classified into each model type.

\subsection{Basic information on the dataset}

\begin{table}
    \centering
    \begin{tabular}{lr}
    \toprule
    Text category & Texts \\
    \midrule
Pile-CC & 2,353 \\
PubMed Central & 1,763 \\
ArXiv & 1,172 \\
Github & 925 \\
FreeLaw & 837 \\
StackExchange & 712 \\
Wikipedia (en) & 567 \\
USPTO Backgrounds & 487 \\
PubMed Abstracts & 464 \\
Gutenberg (PG-19) & 251 \\
DM Mathematics & 151 \\
EuroParl & 83 \\
HackerNews & 67 \\
Ubuntu IRC & 54 \\
PhilPapers & 51 \\
NIH ExPorter & 41 \\
Enron Emails & 22 \\
\midrule
Total & 10,000\\
    \bottomrule
    \end{tabular}
    \caption{Number of texts in each text category.}
    \label{tab:text-category}
\end{table}

The dataset used in our experiments consists of a total of 10,000 texts, which are divided into 17 text categories. Table~\ref{tab:text-category} shows the number of texts in each category.

To assign colors to the text categories, we first compute the average text embedding for each category in the Pile using \texttt{simcse-roberta-large}~\cite{gao2021simcse}. Next, we calculate a tour over the 17 average embedding vectors by solving the traveling salesman problem\footnote{Inspired by~\citet{DBLP:conf/naacl/Sato22,DBLP:conf/emnlp/YamagiwaTS24}.}. The TSP is solved using the nearest neighbor method to generate an initial tour, which is then refined using a 2-opt improvement procedure, and Euclidean distance is used as the metric. Based on the adjacency relationships along this tour, we segment the hue circle at equal intervals and color each category accordingly.

\subsection{Standard scores} \label{sec:standard-score}
\begin{figure}[t!]
    \centering
    \includegraphics[width=\linewidth]{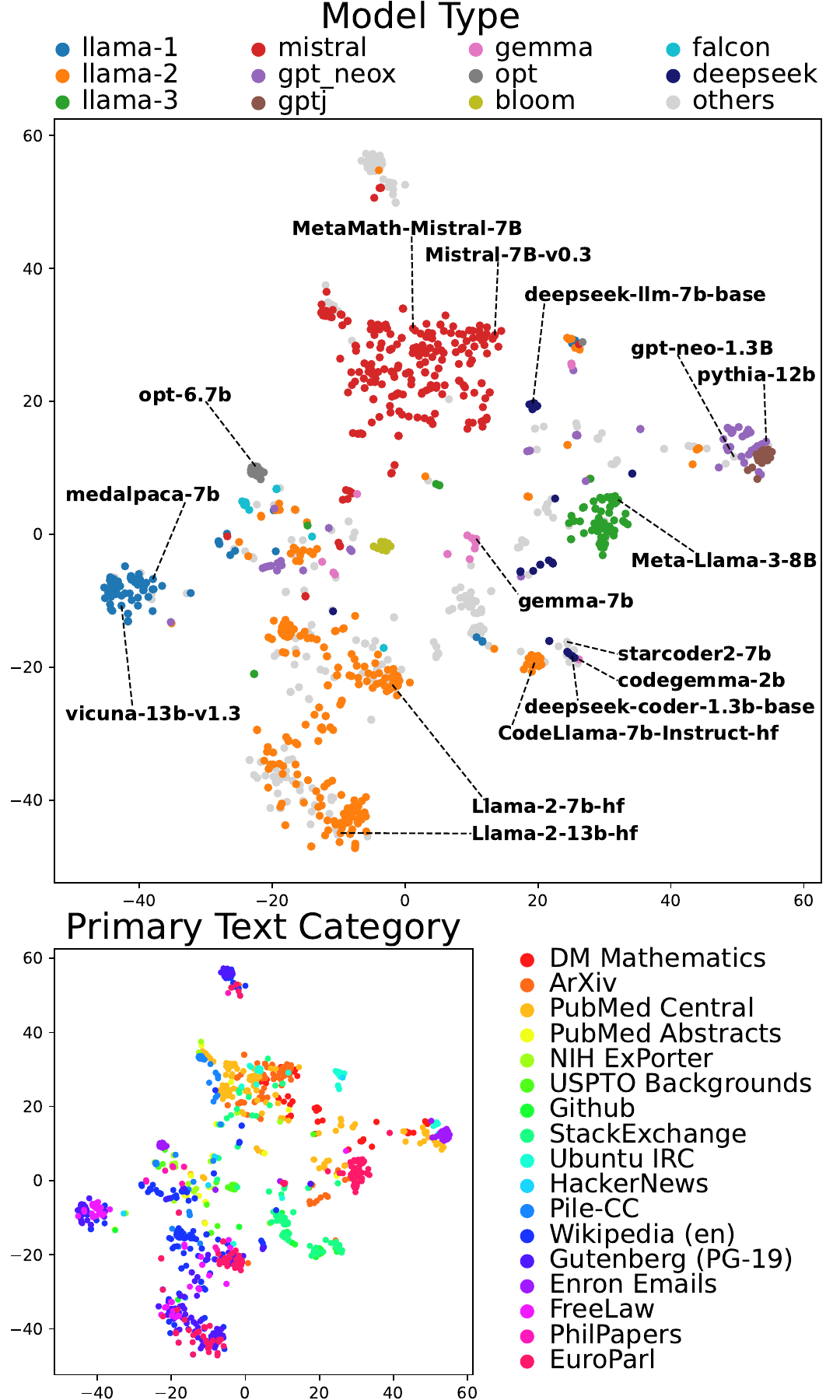}
    \caption{Model maps obtained by double centered log-likelihood matrix $\bmQ$. These maps correspond to Figure~\ref{fig:model-map-tsne-intro}. (Top) Colors indicate model types.  
(Bottom) Colors indicate the text category in which each model attains the highest 
standardized log-likelihood score among 17 categories.}
    \label{fig:Q-fig1}
\end{figure}

In the three experiments described below, we use standardized values\footnote{By subtracting the mean from each value and dividing by the standard deviation, the data is transformed to have a mean of 0 and a variance of 1.}, or $Z$-score normalization, of both the log-likelihood and the benchmark scores calculated for the $K$ language models.
\begin{itemize}
    \item In Figures~\ref{fig:model-map-tsne-intro} and \ref{fig:Q-fig1}, where we define each language model's primary text category, we use the average log-likelihood for each category, standardized across all models.
    \item In Section~\ref{sec:model-map}, to determine each language model's primary task, we standardize each task's score across the $K$ models.
    \item Furthermore, in the data leakage detection described in Section~\ref{sec:model-map}, we use the difference between the standardized mean log-likelihood and the standardized 6-TaskMean score as the indicator.
\end{itemize}

\begin{figure*}
    \centering
    \includegraphics[width=\linewidth]{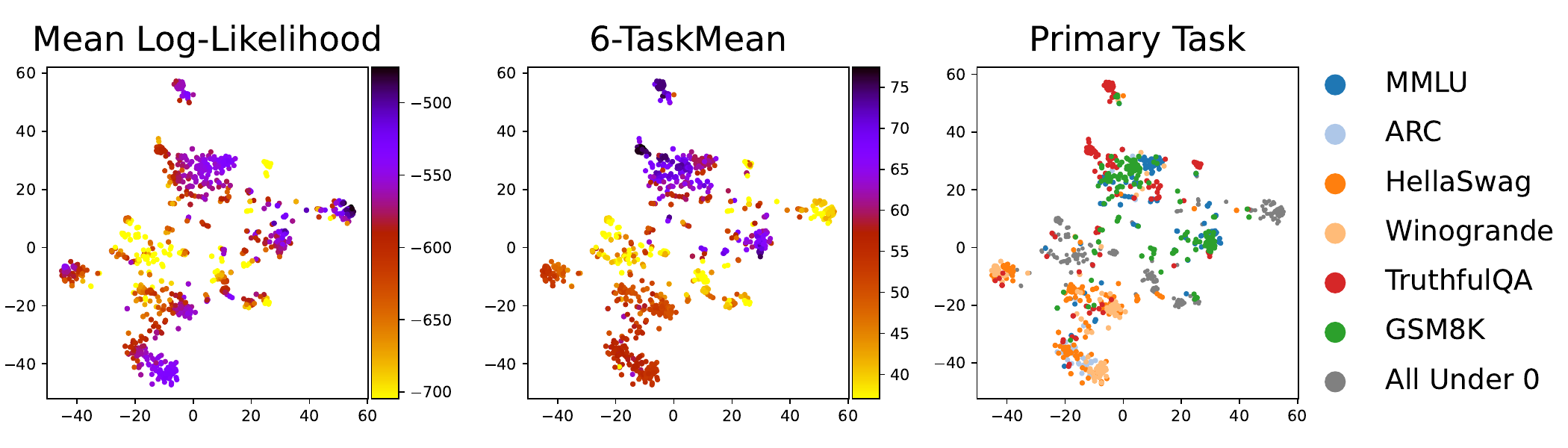}
    \caption{Model maps obtained by double centered log-likelihood matrix $\bmQ$. These maps correspond to Figure~\ref{fig:modelmap-logp-score-task}. These maps are illustrating model performance.  
From left to right, the panels show each model's mean log-likelihood, 6-TaskMean score, and the ``primary task,'' which refers to the task where each model achieves the highest standardized score among the six tasks, color-coded accordingly.  
The color bar is clipped at the 10th percentile for mean log-likelihood and 6-TaskMean, with darker colors indicating better performance.  
In the primary task panel, models with standardized scores below zero across all six tasks are labeled as ``All Under 0.''}
    \label{fig:Q-fig5}
\end{figure*}

\begin{figure}[t!]
    \centering
    \includegraphics[width=\linewidth]{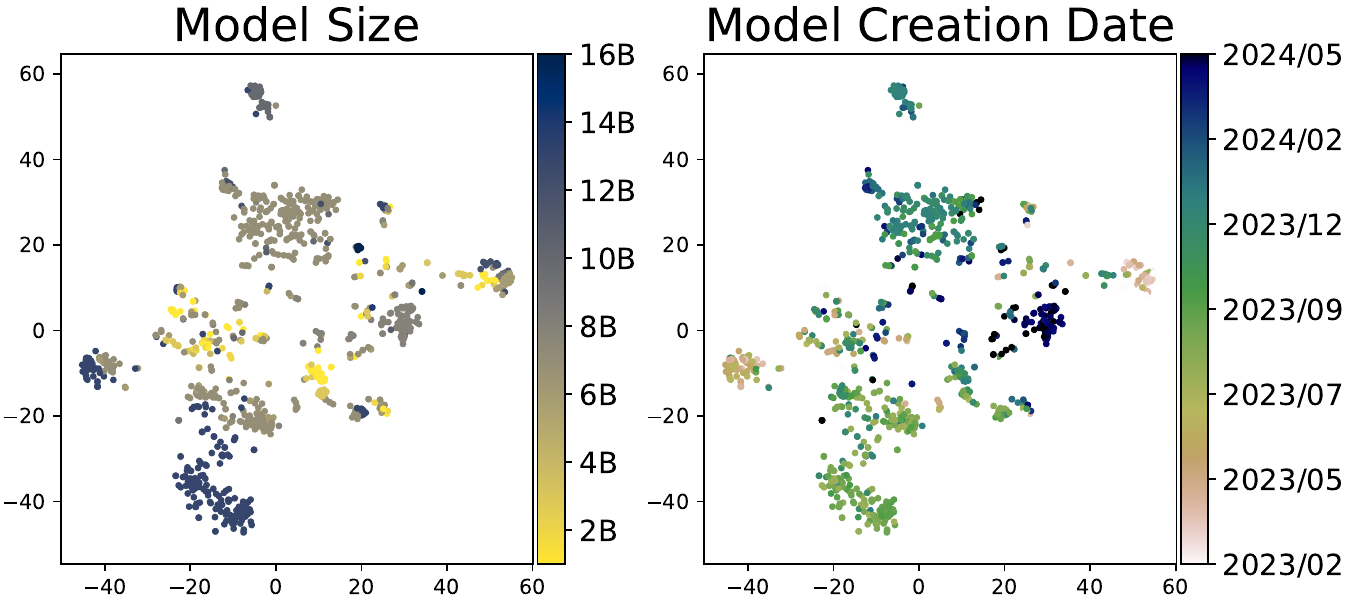}
    \caption{Model maps obtained by double centered log-likelihood matrix $\bmQ$ color-coded by (Left) model size and (Right) model creation date.}
    \label{fig:Q-size-date}
\end{figure}

\subsection{Hierarchical clustering settings}
Figure~\ref{fig:matrix-Q} displays the double-centered log-likelihood matrix $\bmQ$, with hierarchical clustering applied to both its rows and columns.
We implemented the clustering using SciPy~\citep{2020SciPy-NMeth}.
Distance matrices were computed using \texttt{scipy.spatial.distance.pdist}, and clustering was performed using \texttt{scipy.cluster.hierarchy.linkage}.
For clustering models, we used \texttt{sqeuclidean} as the metric and \texttt{median} as the linkage method.
For clustering texts, we used \texttt{correlation} as the metric and \texttt{average} as the linkage method.
For the hierarchical clustering shown in Fig.~\ref{fig:cluster-100-models}, which presents a dendrogram of 100 language models, we used \texttt{sqeuclidean} as the metric and \texttt{median} as the linkage method.
The vertical axis of the dendrogram uses the symmetric logarithmic scale (with a linear threshold of 250) implemented in \texttt{matplotlib}. To ensure that the values on the vertical axis correspond to the Kullback-Leibler divergence, we used the $\bmq$-coordinates divided by $\sqrt{2N}$.

\section{Additional Model Maps} \label{sec:add-model-map}

In this section, we present additional model maps, including a figure that lists all the model names and a map obtained through dimensionality reduction of the double-centered log-likelihood matrix $\bmQ$.

\subsection{Model map via the double centered log-likelihood matrix} \label{sec:model-map-Q}

In the main text, we use a model map generated by dimensionality reduction of the log-likelihood matrix $\bm{L}$.
Here, in Figs.~\ref{fig:Q-fig1}, \ref{fig:Q-fig5} and \ref{fig:Q-size-date}, we present model maps obtained by dimensionality reduction of the double-centered log-likelihood matrix $\bmQ$.

\subsection{Model map with model names} \label{sec:model-map-with-names}

\begin{figure*}
    \centering
    \includegraphics[width=\linewidth]{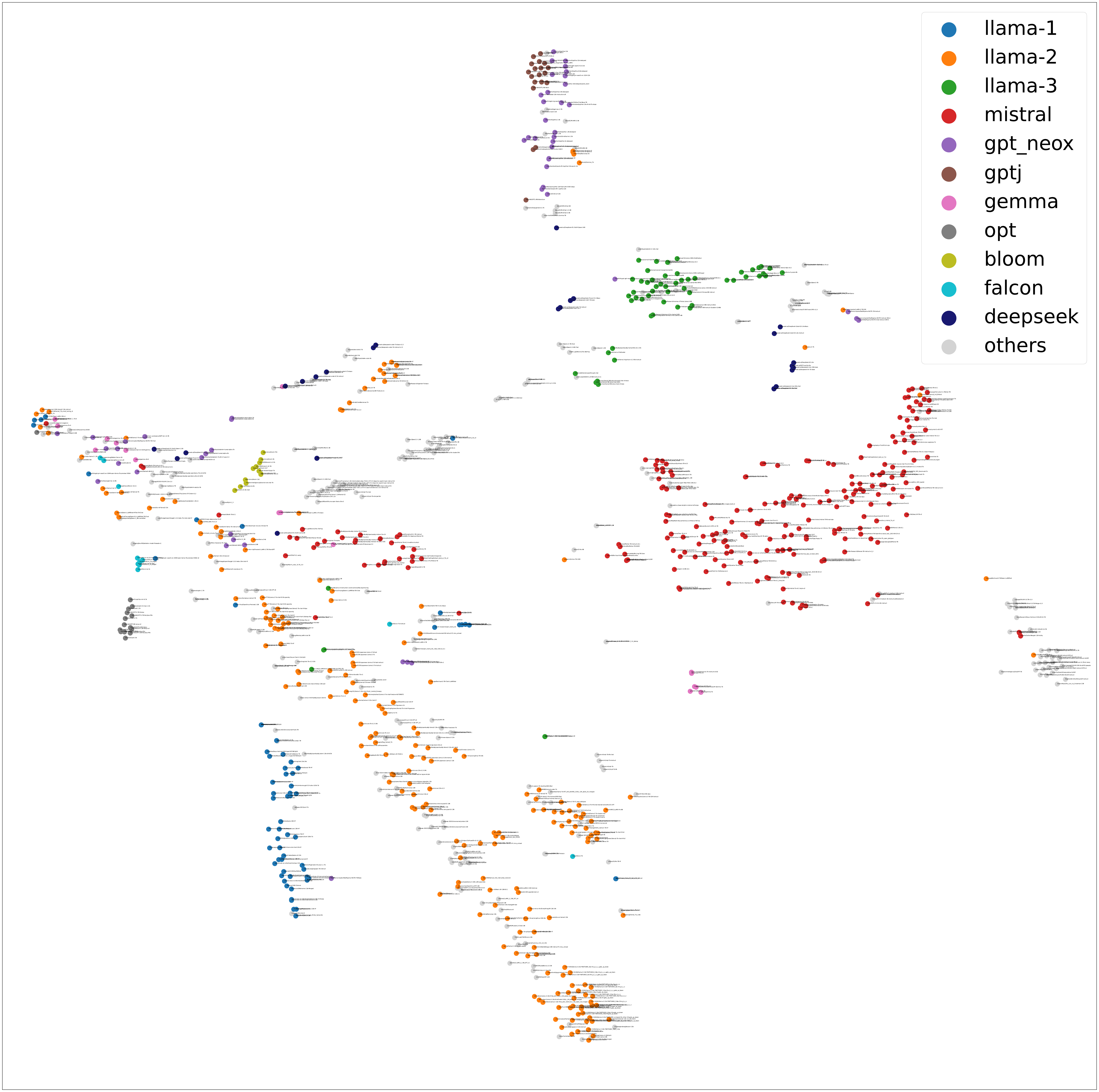}
    \caption{Model map obtained by dimensionality reduction of the log-likelihood matrix $\bm{L}$. Each point on the model map is labeled with the corresponding model name.}
    \label{fig:large-map}
\end{figure*}

\begin{figure*}
    \centering
    \includegraphics[width=\linewidth]{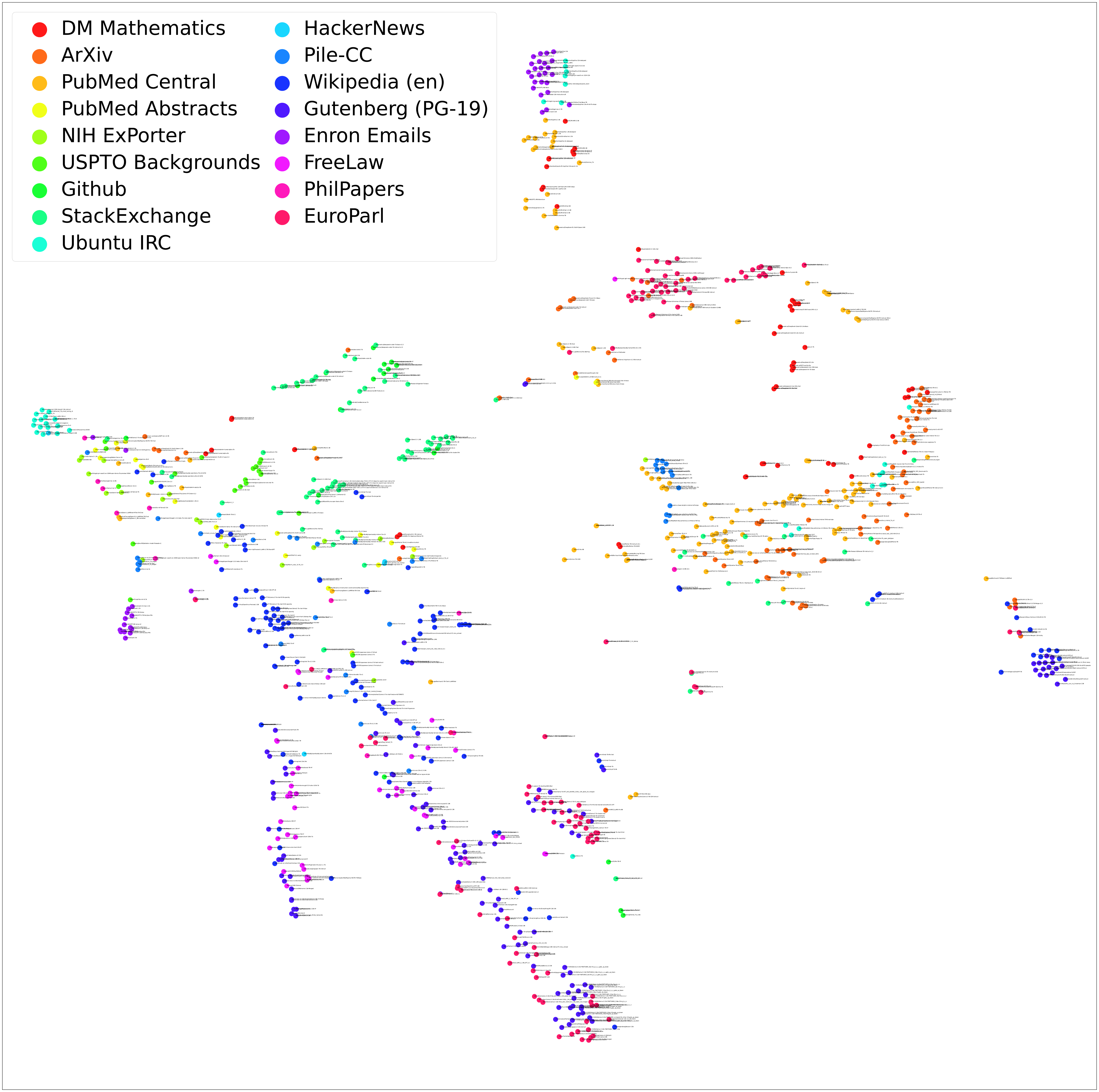}
    \caption{Model map obtained by dimensionality reduction of the log-likelihood matrix $\bm{L}$. Each point on the model map is labeled with the corresponding model name. Colors indicate the model's ``primary text category,'' the text category where the model achieves the highest standardized log-likelihood score among 17 categories.}
    \label{fig:large-map-text}
\end{figure*}

\begin{figure*}
    \centering
    \includegraphics[width=\linewidth]{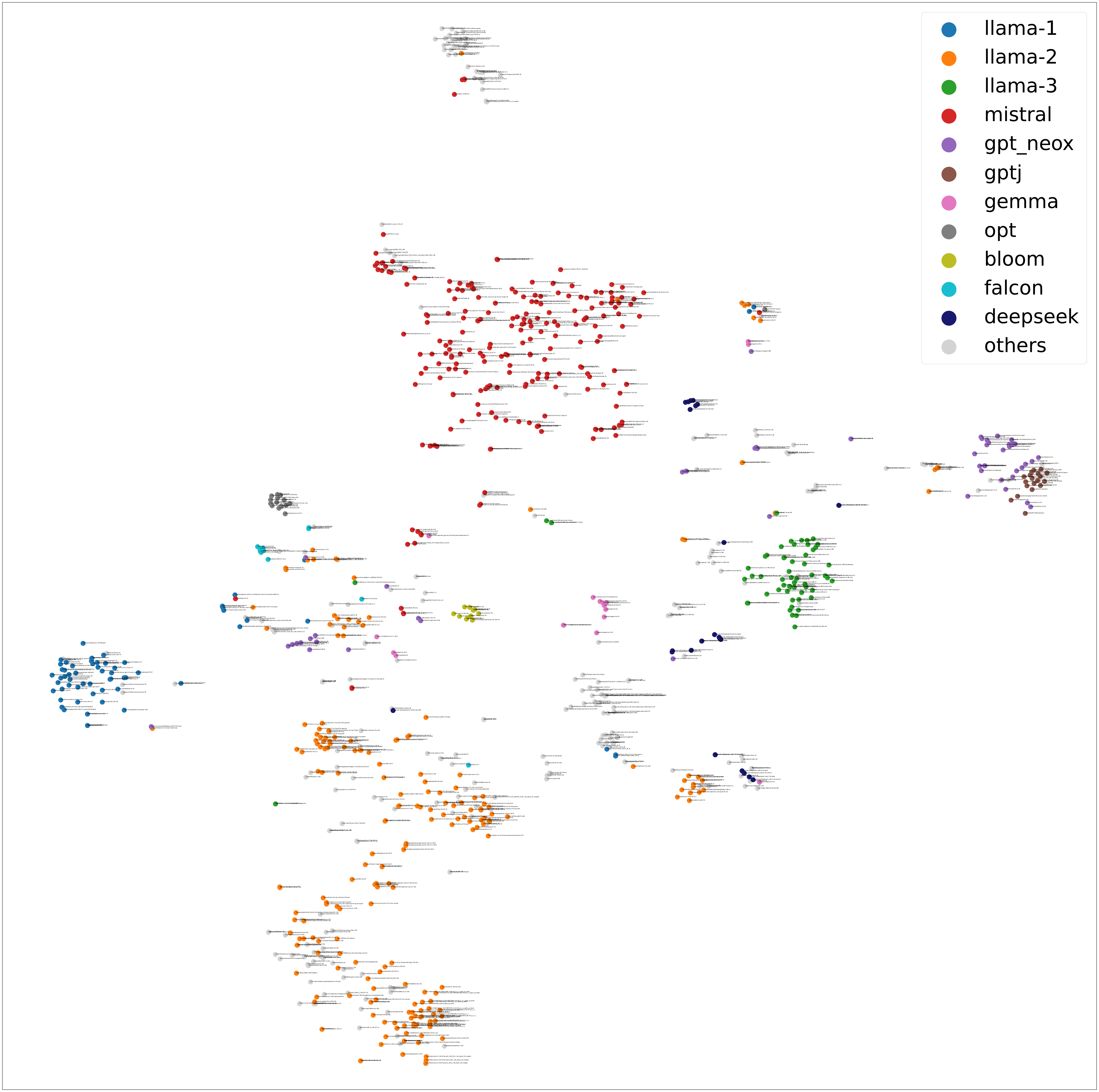}
    \caption{Model map obtained by dimensionality reduction of the double-centered log-likelihood matrix $\bmQ$. Each point on the model map is labeled with the corresponding model name.}
    \label{fig:large-map-Q}
\end{figure*}

\begin{figure*}
    \centering
    \includegraphics[width=\linewidth]{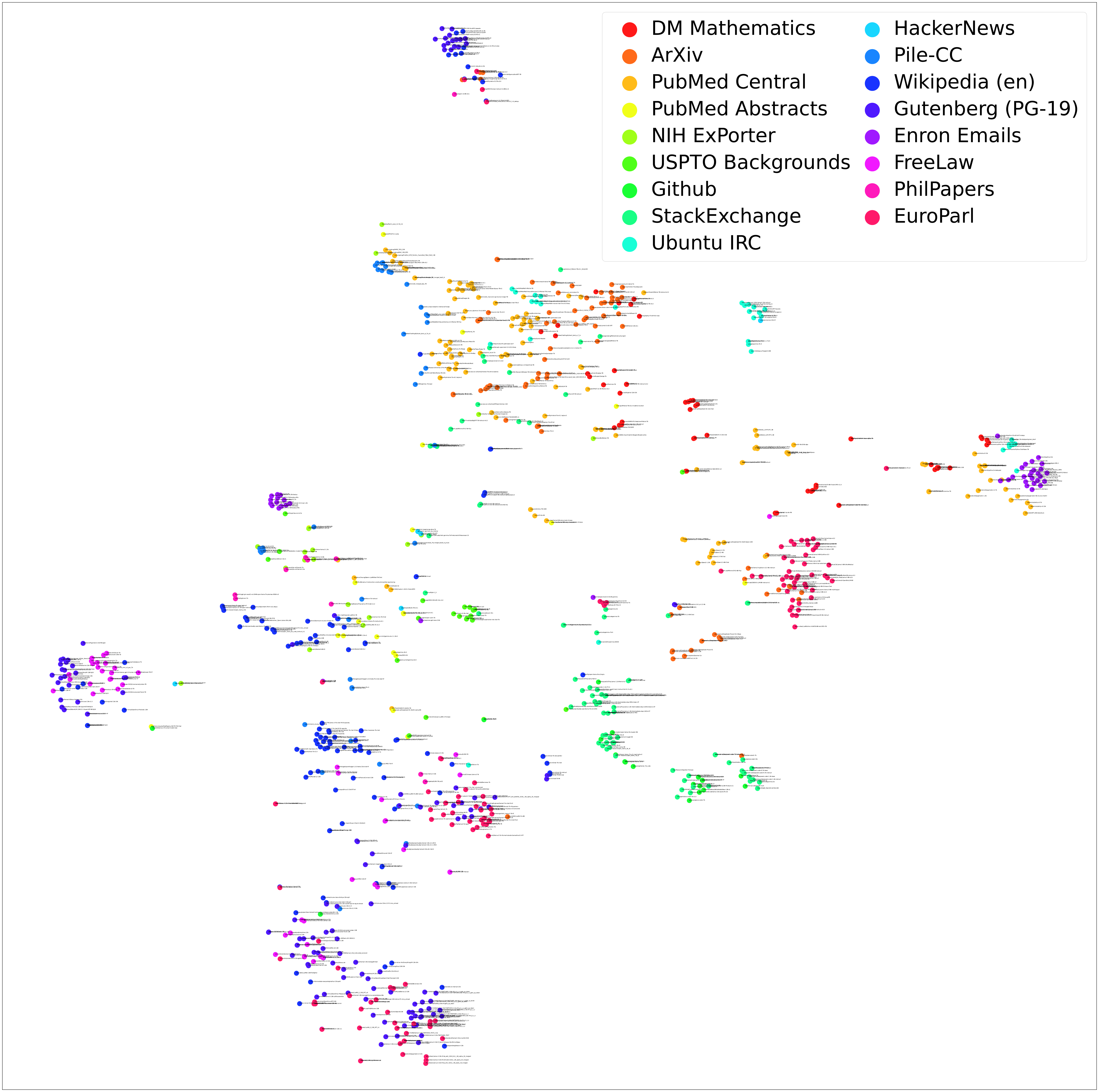}
    \caption{Model map obtained by dimensionality reduction of the double-centered log-likelihood matrix $\bmQ$. Each point on the model map is labeled with the corresponding model name. Colors indicate the model's ``primary text category,'' the text category where the model achieves the highest 
    standardized log-likelihood score among 17 categories.}
    \label{fig:large-map-Q-text}
\end{figure*}

We present figures that display the model names corresponding to each point on the model maps defined in Section~\ref{sec:model-map}. For both the log-likelihood matrix $\bm{L}$ and the double-centered log-likelihood matrix $\bm{Q}$, we provide two types of maps: one colored by model type and another colored by primary text category. Figures~\ref{fig:large-map} and \ref{fig:large-map-text} show the maps obtained by applying dimensionality reduction to $\bm{L}$, while Figures~\ref{fig:large-map-Q} and \ref{fig:large-map-Q-text} show the maps obtained using $\bm{Q}$.

\section{Table of Nearest Neighbor Models}\label{sec:top10-tables}

Table~\ref{tab:nearest-neighbor-16} presents the top 10 nearest neighbors among the 1,018 language models for each of the models highlighted in the top panel of Figure~\ref{fig:model-map-tsne-intro}. Each sub-table corresponds to a specific model of interest, listing its nearest neighbors in descending order of the KL divergence. From this table, we can see that similar models tend to cluster together, exhibiting relatively small KL divergence values. Additionally, the values in parentheses denote the KL divergence measured in bits per byte of text. This conversion makes it easier to interpret the information cost per byte when comparing predictive distributions of different models.

\begin{table*}[p]
\centering
\begin{adjustbox}{width=0.82\linewidth}
\begin{tabular}{@{}lrlr@{}}
\toprule
bigcode/starcoder2-7b & KL [bpb] & codellama/CodeLlama-7b-Instruct-hf & KL [bpb] \\ 
\cmidrule(lr){1-2}\cmidrule(lr){3-4}
bigcode/starcoder2-3b & 1.14 & codellama/CodeLlama-7b-hf & 0.0715 \\ 
stabilityai/stable-code-3b & 2.38 & NousResearch/CodeLlama-7b-hf & 0.0715 \\ 
deepseek-ai/deepseek-coder-6.7b-base & 2.50 & codellama/CodeLlama-13b-Instruct-hf & 0.206 \\ 
EleutherAI/llemma\_7b & 2.81 & TheBloke/CodeLlama-13B-Instruct-fp16 & 0.206 \\ 
deepseek-ai/deepseek-coder-7b-base-v1.5 & 3.01 & Nexusflow/NexusRaven-V2-13B & 0.218 \\ 
deepseek-ai/DeepSeek-Coder-V2-Lite-Base & 3.03 & codellama/CodeLlama-13b-hf & 0.236 \\ 
deepseek-ai/deepseek-coder-7b-instruct-v1.5 & 3.03 & NousResearch/CodeLlama-13b-hf & 0.236 \\ 
deepseek-ai/deepseek-coder-6.7b-instruct & 3.18 & OpenAssistant/codellama-13b-oasst-sft-v10 & 0.326 \\ 
deepseek-ai/DeepSeek-Coder-V2-Lite-Instruct & 3.19 & HiTZ/GoLLIE-7B & 0.335 \\ 
meta-math/MetaMath-Llemma-7B & 3.24 & WhiteRabbitNeo/WhiteRabbitNeo-13B-v1 & 0.386 \\ 
\midrule
deepseek-ai/deepseek-coder-1.3b-base & KL [bpb] & deepseek-ai/deepseek-llm-7b-base & KL [bpb] \\ 
\cmidrule(lr){1-2}\cmidrule(lr){3-4}
deepseek-ai/deepseek-coder-1.3b-instruct & 0.610 & deepseek-ai/deepseek-moe-16b-base & 0.194 \\ 
deepseek-ai/deepseek-coder-6.7b-instruct & 0.761 & deepseek-ai/deepseek-llm-7b-chat & 0.364 \\ 
deepseek-ai/deepseek-coder-6.7b-base & 0.892 & deepseek-ai/deepseek-moe-16b-chat & 0.368 \\ 
bigcode/starcoderbase-1b & 1.185 & deepseek-ai/DeepSeek-V2-Lite & 0.455 \\ 
bigcode/starcoderbase-7b & 1.855 & deepseek-ai/ESFT-vanilla-lite & 0.456 \\ 
NTQAI/Nxcode-CQ-7B-orpo & 1.940 & deepseek-ai/DeepSeek-V2-Lite-Chat & 0.868 \\ 
Qwen/CodeQwen1.5-7B-Chat & 1.954 & mistralai/Mistral-7B-Instruct-v0.1 & 1.356 \\ 
bigcode/starcoder2-3b & 2.757 & statking/zephyr-7b-sft-full-orpo & 1.616 \\ 
Salesforce/codegen-6B-multi & 2.929 & Severian/ANIMA-Phi-Neptune-Mistral-7B & 1.692 \\ 
google/codegemma-2b & 3.139 & sethuiyer/Medichat-Llama3-8B & 1.718 \\ 
\midrule
EleutherAI/gpt-neo-1.3B & KL [bpb] & EleutherAI/pythia-12b & KL [bpb] \\ 
\cmidrule(lr){1-2}\cmidrule(lr){3-4}
EleutherAI/gpt-neo-2.7B & 0.325 & matsuo-lab/weblab-10b & 0.207 \\ 
EleutherAI/pythia-1.4b & 0.429 & h2oai/h2ogpt-oig-oasst1-256-6\_9b & 0.237 \\ 
EleutherAI/pythia-1b-deduped & 0.613 & Salesforce/codegen-6B-nl & 0.260 \\ 
HWERI/pythia-1.4b-deduped-sharegpt & 0.625 & EleutherAI/gpt-j-6b & 0.277 \\ 
beaugogh/pythia-1.4b-deduped-sharegpt & 0.625 & TehVenom/Dolly\_Malion-6b & 0.344 \\ 
EleutherAI/pythia-1.4b-deduped & 0.663 & TehVenom/PPO\_Shygmalion-6b & 0.346 \\ 
PygmalionAI/metharme-1.3b & 0.666 & TehVenom/Dolly\_Shygmalion-6b-Dev\_V8P2 & 0.351 \\ 
RWKV/rwkv-raven-1b5 & 0.761 & TehVenom/PPO\_Pygway-V8p4\_Dev-6b & 0.352 \\ 
databricks/dolly-v2-3b & 0.780 & TehVenom/GPT-J-Pyg\_PPO-6B-Dev-V8p4 & 0.364 \\ 
EleutherAI/pythia-2.8b-deduped & 0.872 & TehVenom/PPO\_Shygmalion-V8p4\_Dev-6b & 0.364 \\ 
\midrule
facebook/opt-6.7b & KL [bpb] & google/codegemma-2b & KL [bpb] \\ 
\cmidrule(lr){1-2}\cmidrule(lr){3-4}
KoboldAI/OPT-6.7B-Erebus & 0.00102 & deepseek-ai/deepseek-coder-1.3b-instruct & 2.46 \\ 
KoboldAI/OPT-6.7B-Nerybus-Mix & 0.117 & bigcode/starcoderbase-1b & 2.69 \\ 
KoboldAI/OPT-6B-nerys-v2 & 0.185 & deepseek-ai/deepseek-coder-1.3b-base & 3.14 \\ 
facebook/opt-13b & 0.292 & bigcode/gpt\_bigcode-santacoder & 3.92 \\ 
KoboldAI/OPT-13B-Nerybus-Mix & 0.341 & deepseek-ai/deepseek-coder-6.7b-instruct & 3.97 \\ 
KoboldAI/OPT-13B-Erebus & 0.353 & Qwen/CodeQwen1.5-7B-Chat & 4.09 \\ 
KoboldAI/OPT-13B-Nerys-v2 & 0.406 & NTQAI/Nxcode-CQ-7B-orpo & 4.11 \\ 
facebook/opt-2.7b & 0.486 & Salesforce/codegen-6B-multi & 4.24 \\ 
KoboldAI/OPT-2.7B-Nerybus-Mix & 0.634 & bigcode/starcoderbase-7b & 4.44 \\ 
KoboldAI/OPT-2.7B-Erebus & 0.663 & deepseek-ai/deepseek-coder-6.7b-base & 4.87 \\ 
\midrule
google/gemma-7b & KL [bpb] & lmsys/vicuna-13b-v1.3 & KL [bpb] \\ 
\cmidrule(lr){1-2}\cmidrule(lr){3-4}
SeaLLMs/SeaLLM-7B-v2.5 & 0.367 & TheBloke/stable-vicuna-13B-HF & 0.290 \\ 
VAGOsolutions/SauerkrautLM-Gemma-7b & 0.383 & Yhyu13/chimera-inst-chat-13b-hf & 0.292 \\ 
lemon-mint/gemma-ko-7b-instruct-v0.62 & 0.511 & junelee/wizard-vicuna-13b & 0.325 \\ 
google/gemma-2b & 0.748 & TheBloke/wizard-vicuna-13B-HF & 0.325 \\ 
google/codegemma-7b & 0.885 & TheBloke/UltraLM-13B-fp16 & 0.326 \\ 
PathFinderKR/Waktaverse-Llama-3-KO-8B-Instruct & 1.209 & NousResearch/Nous-Hermes-13b & 0.328 \\ 
FlagAlpha/Llama3-Chinese-8B-Instruct & 1.214 & project-baize/baize-v2-13b & 0.347 \\ 
FairMind/Llama-3-8B-4bit-UltraChat-Ita & 1.236 & TheBloke/guanaco-13B-HF & 0.347 \\ 
migtissera/Llama-3-8B-Synthia-v3.5 & 1.251 & openaccess-ai-collective/minotaur-13b-fixed & 0.349 \\ 
Orenguteng/Llama-3-8B-Lexi-Uncensored & 1.253 & openaccess-ai-collective/wizard-mega-13b & 0.353 \\ 
\midrule
medalpaca/medalpaca-7b & KL [bpb] & meta-llama/Llama-2-13b-hf & KL [bpb] \\ 
\cmidrule(lr){1-2}\cmidrule(lr){3-4}
TheBloke/guanaco-7B-HF & 0.424 & TaylorAI/Flash-Llama-13B & 1.82e-16 \\ 
eachadea/vicuna-7b-1.1 & 0.499 & TheBloke/Llama-2-13B-fp16 & 3.6e-06 \\ 
TehVenom/Pygmalion-Vicuna-1.1-7b & 0.548 & StudentLLM/Alpagasus-2-13b-QLoRA-merged & 0.00668 \\ 
lmsys/vicuna-7b-v1.3 & 0.556 & CHIH-HUNG/llama-2-13b-FINETUNE2\_TEST\_2.2w & 0.0113 \\ 
jphme/orca\_mini\_v2\_ger\_7b & 0.569 & garage-bAInd/Platypus2-13B & 0.0154 \\ 
ajibawa-2023/Uncensored-Jordan-7B & 0.570 & CHIH-HUNG/llama-2-13b-dolphin\_5w & 0.0213 \\ 
bofenghuang/vigogne-7b-instruct & 0.580 & CHIH-HUNG/llama-2-13b-OpenOrca\_5w & 0.0222 \\ 
TheBloke/airoboros-7b-gpt4-fp16 & 0.582 & CHIH-HUNG/llama-2-13b-FINETUNE4\_3.8w-r16-gate\_up\_down-test1 & 0.0256 \\ 
TheBloke/tulu-7B-fp16 & 0.628 & CHIH-HUNG/llama-2-13b-FINETUNE4\_addto15k\_4.5w-r16-gate\_up\_down & 0.0258 \\ 
TehVenom/Pygmalion\_AlpacaLora-7b & 0.640 & CHIH-HUNG/llama-2-13b-FINETUNE4\_compare15k\_4.5w-r16-gate\_up\_down & 0.0288 \\ 
\midrule
meta-llama/Llama-2-7b-hf & KL [bpb] & meta-llama/Meta-Llama-3-8B & KL [bpb] \\ 
\cmidrule(lr){1-2}\cmidrule(lr){3-4}
ibranze/araproje-llama2-7b-hf & 0 & Undi95/Meta-Llama-3-8B-hf & 4.56e-06 \\ 
TheTravellingEngineer/llama2-7b-chat-hf-v4 & 0 & dfurman/Llama-3-8B-Orpo-v0.1 & 0.0104 \\ 
TheTravellingEngineer/llama2-7b-chat-hf-v2 & 0 & migtissera/Tess-2.0-Llama-3-8B & 0.0428 \\ 
TaylorAI/Flash-Llama-7B & 0 & freewheelin/free-llama3-dpo-v0.2 & 0.101 \\ 
yeen214/test\_llama2\_7b & 0 & jondurbin/bagel-8b-v1.0 & 0.164 \\ 
NewstaR/Starlight-7B & 4.57e-06 & migtissera/Llama-3-8B-Synthia-v3.5 & 0.190 \\ 
Delcos/Mistral-Pygmalion-7b & 0.0220 & nvidia/Llama3-ChatQA-1.5-8B & 0.191 \\ 
elliotthwang/elliott\_Llama-2-7b-hf & 0.0246 & ruslanmv/Medical-Llama3-8B & 0.206 \\ 
garage-bAInd/Platypus2-7B & 0.0338 & FairMind/Llama-3-8B-4bit-UltraChat-Ita & 0.296 \\ 
Lazycuber/L2-7b-Base-Guanaco-Uncensored & 0.0346 & NousResearch/Hermes-2-Theta-Llama-3-8B & 0.330 \\ 
\midrule
meta-math/MetaMath-Mistral-7B & KL [bpb] & mistralai/Mistral-7B-v0.3 & KL [bpb] \\ 
\cmidrule(lr){1-2}\cmidrule(lr){3-4}
Weyaxi/MetaMath-NeuralHermes-2.5-Mistral-7B-Linear & 0.0639 & MaziyarPanahi/Mistral-7B-v0.3 & 3.64e-16 \\ 
Weyaxi/MetaMath-Tulpar-7b-v2-Slerp & 0.121 & mistral-community/Mistral-7B-v0.2 & 0.0107 \\ 
Weyaxi/MetaMath-OpenHermes-2.5-neural-chat-v3-3-Slerp & 0.150 & unsloth/mistral-7b-v0.2 & 0.0107 \\ 
Q-bert/Bumblebee-7B & 0.158 & mistralai/Mistral-7B-v0.1 & 0.0327 \\ 
OpenPipe/mistral-ft-optimized-1227 & 0.163 & Cartinoe5930/Llama2\_init\_Mistral & 0.0442 \\ 
Toten5/Marcoroni-neural-chat-7B-v2 & 0.169 & Locutusque/Hercules-3.1-Mistral-7B & 0.0476 \\ 
ignos/Mistral-T5-7B-v1 & 0.170 & migtissera/Synthia-7B-v3.0 & 0.0496 \\ 
Weyaxi/MetaMath-Chupacabra-7B-v2.01-Slerp & 0.170 & uukuguy/speechless-zephyr-code-functionary-7b & 0.0530 \\ 
Q-bert/Optimus-7B & 0.175 & uukuguy/zephyr-7b-alpha-dare-0.85 & 0.0530 \\ 
Weyaxi/MetaMath-NeuralHermes-2.5-Mistral-7B-Ties & 0.189 & crumb/apricot-wildflower-20 & 0.0607 \\ 
\bottomrule
\end{tabular}
\end{adjustbox}
\caption{
Top 10 nearest neighbors among the 1,018 language models for each model labeled in the top panel of Fig.~\ref{fig:model-map-tsne-intro}.
The values indicate the KL divergence measured in bits per byte (bpb), as defined in Section~\ref{sec:byte-normalized-KL}.
These are computed using formula~(\ref{eq:main-KL-xi}) in Section~\ref{sec:model-text}, by multiplying the original KL divergence by 0.001484.
}
\label{tab:nearest-neighbor-16}
\end{table*}

\begin{figure*}[t!]
    \centering
    \includegraphics[width=0.98\linewidth]{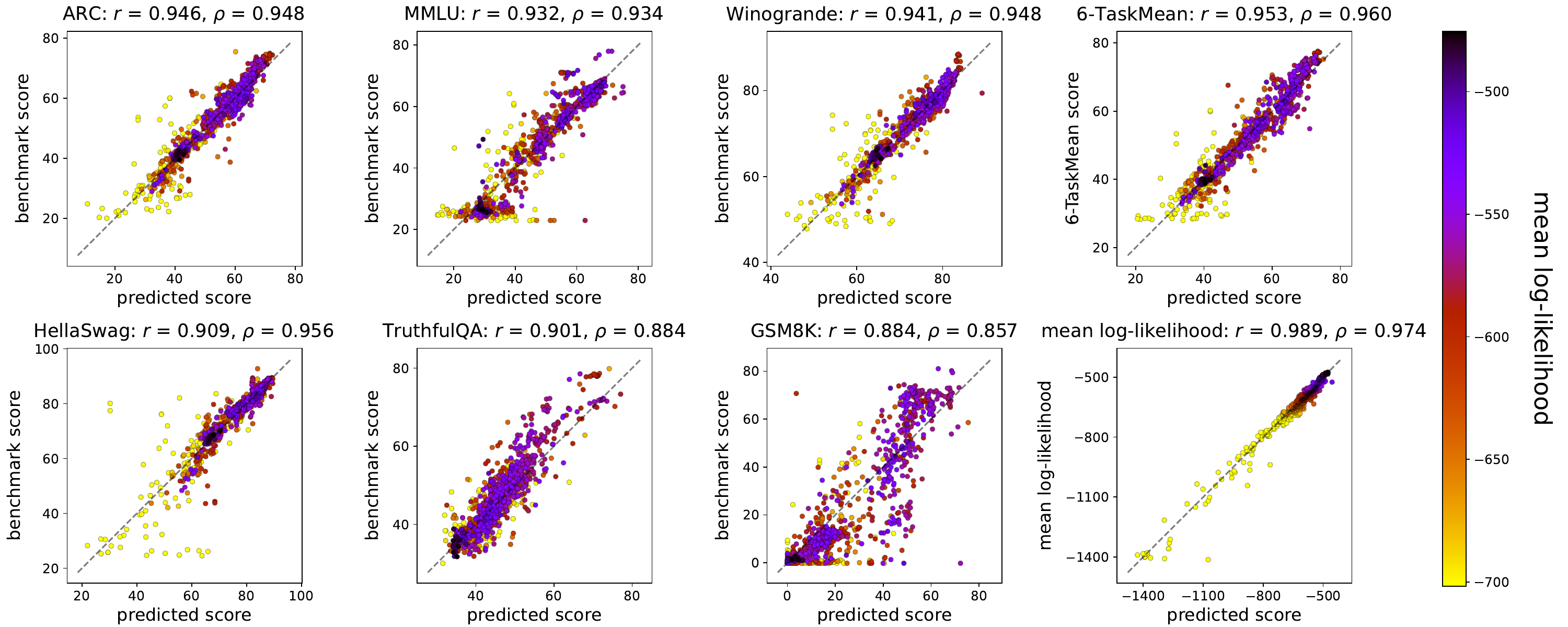}
    \caption{
Scatter plots of predicted scores versus benchmark scores for test sets across the six benchmark tasks (the dashed line indicates the identity line).
Additionally, results for predicting 6-TaskMean (identical to Fig.~\ref{fig:group_5fold_split_average_task}) and the mean log-likelihood are also shown. 
Each point is color-coded by the mean log-likelihood, with higher mean log-likelihood values generally corresponding to higher task scores. 
For better visualization, the color bar range is clipped to the 10th--100th percentile.}
    \label{fig:group_5fold_split_all_tasks}
\end{figure*}

\section{Details of Weight Interpolation}

\label{app:weight-parameter}

In this section, we describe the experimental details concerning the relationship between model coordinates and model weights, as introduced in Section \ref{sec:weight-parameter}.

\paragraph{Constructing the weight grid.}
Let $p_0$ denote the base model, and $p_1$ and $p_2$ denote the fine-tuned models derived from $p_0$. We denote the weight parameter vectors of these models as $W_0$, $W_1$, and $W_2$, respectively. To construct the weight grid, we merged the model weights using the following linear operation:
\begin{equation}
W_{\alpha,\beta} = W_0 + \alpha(W_1 - W_0) + \beta(W_2 - W_0),
\end{equation}
where the merge ratios $\alpha, \beta \in \mathbb{R}$ were chosen from 36 evenly spaced combinations within the interval $[0,1]$: $\{0.0, 0.2, 0.6, 0.8, 1.0\}$. The original models $W_0$, $W_1$, and $W_2$ correspond to $W_{0,0}$, $W_{1,0}$, and $W_{0,1}$, respectively. When $\alpha + \beta \leq 1$, the operation corresponds to linear interpolation between models.
Even among models with the same architecture, the sizes of the embedding/unembedding matrices may differ. In such cases, we truncated or reshaped the weight parameters to match the base model.

\paragraph{Computing model coordinates.}
For each composed model $p_{\alpha, \beta}$ with weight $W_{\alpha, \beta}$, we computed the model coordinates $\bm{q}_{\alpha,\beta}$ following the method described in Section \ref{sec:model-text}. The tokenizer of the base model was used to ensure consistency. The text data consists of all 10,000 texts prepared in Section \ref{sec:text-set}. Additionally, the deterministic algorithms option\footnote{\texttt{torch.use\_deterministic\_algorithms(True)}} was enabled in the implementation to ensure reproducibility.

\paragraph{Selection of models.}
We conducted experiments using two base models: \texttt{Llama-2-7b-hf} and \texttt{mistralai/Mistral-7B-v0.1}.  
For each base model $p_0$, we selected the two most downloaded fine-tuned models available on Hugging Face.
Specifically, when $p_0 = \texttt{Llama-2-7b-hf}$, we set $p_1 = \texttt{vicuna-7b-v1.5}$ and $p_2 = \texttt{Llama-2-7b-chat-hf}$.  
When $p_0 = \texttt{mistralai/Mistral-7B-v0.1}$, we set $p_1 = \texttt{HuggingFaceH4/zephyr-7b-beta}$ and $p_2 = \texttt{mistralai/Mistral-7B-Instruct-v0.1}$.

\paragraph{Visualization.}
\begin{figure}[t!]
    \centering\includegraphics[width=\linewidth]{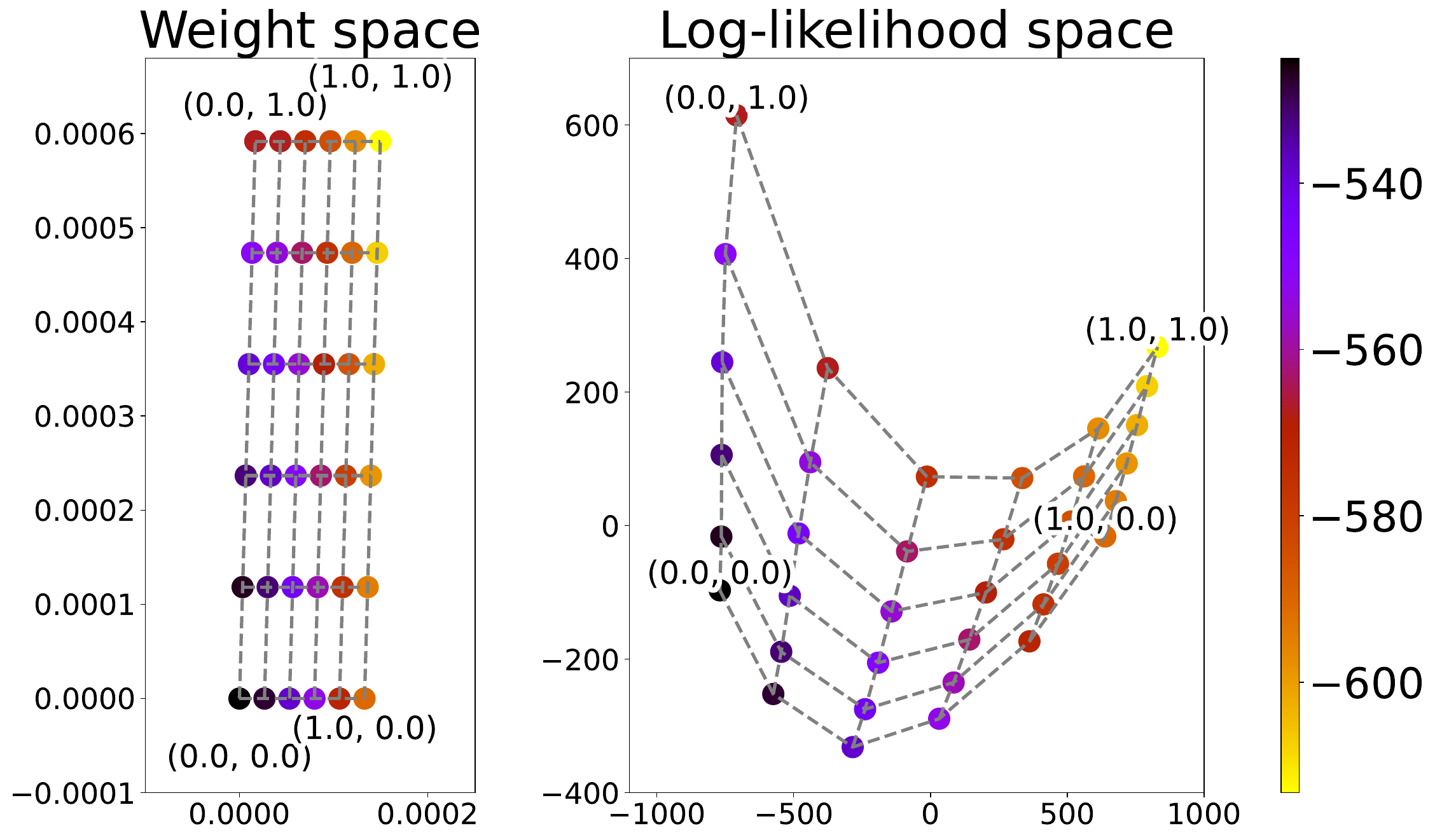}

    \caption{
    Visualization of 36 language models obtained by linearly interpolating pretrained model weights based on \texttt{mistralai/Mistral-7B-v0.1}.  
    Each point is color-coded according to its mean log-likelihood.  
    (Left) Models in the weight parameter space.  
    (Right) Models in the log-likelihood space, represented by the \(\bmq\)-coordinate system.
    }
    
    \label{fig:weight-interpolation-grids-mistral}
\end{figure}
\begin{figure}[t!]
    \centering
    \includegraphics[width=0.98\linewidth]{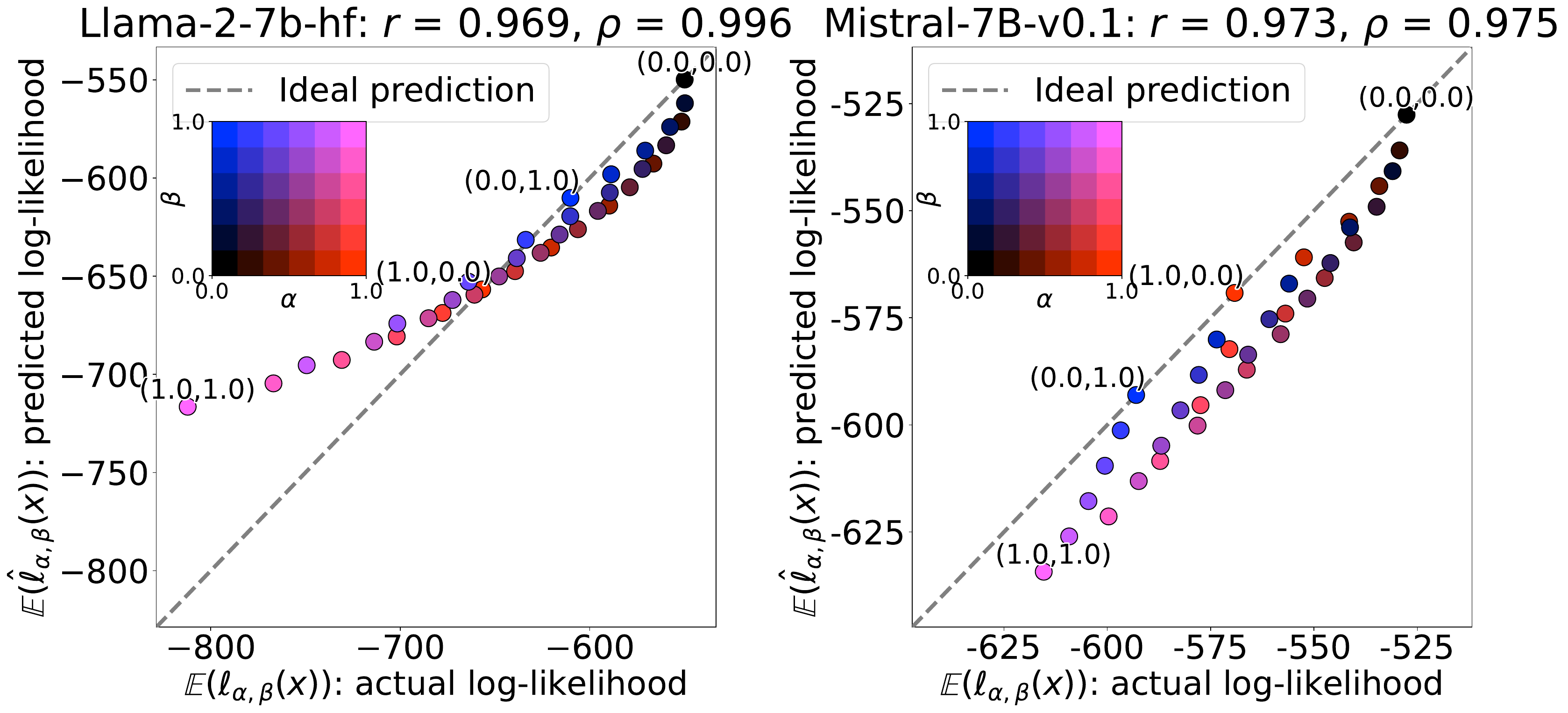}
    \caption{
    Scatter plots comparing the actual and predicted mean log-likelihoods for 36 interpolated models derived from \texttt{Llama-2-7b-hf} (left) and \texttt{Mistral-7B-v0.1} (right).  
    Each point corresponds to a unique combination of merge ratios $(\alpha, \beta)$ and is color-coded accordingly. 
    The dashed line indicates ideal prediction.
    Strong correlations are observed for both model sets: Llama-2 shows $r = 0.969$, $\rho = 0.996$, and Mistral shows $r = 0.973$, $\rho = 0.975$.
    A 2D color map of $(\alpha, \beta)$ values is shown as an inset.
    }
    \label{fig:mean-logp-interpolation}
\end{figure}

Figures \ref{fig:weight-interpolation-grids-llama} and \ref{fig:weight-interpolation-grids-mistral} show the linearly merged models, visualized in both weight space and log-likelihood space. The corners of the grid are labeled with their corresponding $(\alpha, \beta)$ values.

In the left panel, we visualized $W_{\alpha, \beta}$ in the weight space. Since the dimensionality of $W_{\alpha, \beta}$, i.e., the number of model parameters, is extremely high, we employed a 2D projection method using the norms of the difference vectors $r_1 = \left\|W_1 - W_0\right\|_2$, $r_2 = \left\|W_2 - W_0\right\|_2$, and the angle between them, $\phi = \arccos((W_1 - W_0)^\top (W_2 - W_0) / r_1 r_2)$. Each point was placed at $(\alpha r_1 + \beta r_2\cos\phi, \beta r_2\sin\phi)$.

In the right panel, we visualized the model coordinates $\bmq_{\alpha, \beta}$ by Principal Component Analysis (PCA). Each model $p_{\alpha, \beta}$ was mapped onto the $\bmq$-coordinate system to analyze the structure of the interpolated models.

\paragraph{Estimating the mean log-likelihood of interpolated models.}

To assess the potential of predicting task performance from log-likelihood vectors without explicitly computing them for every interpolated model, we evaluated whether the mean log-likelihoods can be estimated by linearly interpolating those of the base models.
Let $\ell_{\alpha,\beta}(x)$ denote the log-likelihood of input $x$ computed by the merged model $p_{\alpha,\beta}$. Let $\ell_0$, $\ell_1$, and $\ell_2$ be the log-likelihoods from the base and fine-tuned models $p_0$, $p_1$, and $p_2$, respectively. We define the linearly interpolated log-likelihood as:
\begin{equation}
\hat{\ell}_{\alpha,\beta} =  \ell_0 + \alpha (\ell_1 - \ell_0) + \beta (\ell_2 - \ell_0).
\end{equation}
Figure~\ref{fig:mean-logp-interpolation} shows a strong correlation between the predicted and actual mean log-likelihoods.
This result suggests that log-likelihood vectors can be approximated by linear interpolation, enabling efficient prediction of model performance without computing log-likelihoods for every model.

\begin{table*}[t!]
\tiny
    \centering
    \begin{tabular}{@{\hspace{0.95em}}lr@{\hspace{0.95em}}r@{\hspace{0.95em}}r@{\hspace{0.95em}}r@{\hspace{0.95em}}r@{\hspace{0.95em}}r@{\hspace{0.95em}}r@{\hspace{0.95em}}r@{\hspace{0.95em}}}
    \toprule
        & ARC & HellaSwag & MMLU & TruthfulQA & Winogrande & GSM8K & Average & mean log-likelihood\\
        \midrule
        Pearson's $r$ & $0.941\pm0.002$ & $0.904\pm0.005$ & $0.924\pm0.006$ & $0.889\pm0.013$ & $0.934\pm0.005$ & $0.864\pm0.019$ & $0.947\pm0.005$ & $0.987\pm0.006$ \\
        Spearman's $\rho$ & $0.944\pm0.005$ & $0.951\pm0.003$ & $0.926\pm0.005$ & $0.875\pm0.003$ & $0.943\pm0.009$ & $0.841\pm0.013$ & $0.956\pm0.004$ & $0.971\pm0.006$ \\
    \bottomrule
    \end{tabular}
\caption{Group 5-fold based on model types using $\bmQ$: Mean and standard deviation of the correlation coefficients between predicted and actual benchmark scores. Coefficients were obtained by ridge regression under five data splits based on model types. Results for predicting 6-TaskMean and the mean log-likelihood are also included.}
    \label{tab:group-result-regression_mean_and_std}
\end{table*}

\begin{table*}[t!]
\tiny
    \centering
    \begin{tabular}{@{\hspace{0.95em}}lr@{\hspace{0.95em}}r@{\hspace{0.95em}}r@{\hspace{0.95em}}r@{\hspace{0.95em}}r@{\hspace{0.95em}}r@{\hspace{0.95em}}r@{\hspace{0.95em}}r@{\hspace{0.95em}}}
    \toprule
        & ARC & HellaSwag & MMLU & TruthfulQA & Winogrande & GSM8K & Average & mean log-likelihood\\
        \midrule
        Pearson's $r$ & $0.968\pm0.001$ & $0.939\pm0.004$ & $0.960\pm0.002$ & $0.952\pm0.001$ & $0.960\pm0.003$ & $0.931\pm0.001$ & $0.973\pm0.001$ & $0.994\pm0.001$ \\
        Spearman's $\rho$ & $0.972\pm0.001$ & $0.970\pm0.002$ & $0.966\pm0.001$ & $0.929\pm0.001$ & $0.969\pm0.001$ & $0.891\pm0.004$ & $0.976\pm0.001$ & $0.990\pm0.000$ \\
    \bottomrule
    \end{tabular}
\caption{Random 5-fold using $\bmQ$: Mean and standard deviation of the correlation coefficients, obtained as in Table~\ref{tab:group-result-regression_mean_and_std}, but using five random data splits instead of splits based on model types.}
    \label{tab:result-regression_mean_and_std}
\end{table*}

\begin{table*}[t!]
\tiny
    \centering
    \begin{tabular}{@{\hspace{0.95em}}lr@{\hspace{0.95em}}r@{\hspace{0.95em}}r@{\hspace{0.95em}}r@{\hspace{0.95em}}r@{\hspace{0.95em}}r@{\hspace{0.95em}}r@{\hspace{0.95em}}r@{\hspace{0.95em}}}
    \toprule
        & ARC & HellaSwag & MMLU & TruthfulQA & Winogrande & GSM8K & Average & mean log-likelihood\\
        \midrule
        Pearson's $r$ & $0.939 \pm 0.002$ & $0.903 \pm 0.007$ & $0.923 \pm 0.006$ & $0.890 \pm 0.013$ & $0.935 \pm 0.006$ & $0.863 \pm 0.018$ & $0.945 \pm 0.005$ & $1.000 \pm 0.000$ \\
        Spearman's $\rho$ & $0.943 \pm 0.005$ & $0.952 \pm 0.003$ & $0.925 \pm 0.005$ & $0.875 \pm 0.003$ & $0.944 \pm 0.008$ & $0.841 \pm 0.013$ & $0.954 \pm 0.004$ & $1.000 \pm 0.000$ \\
    \bottomrule
    \end{tabular}
\caption{Group 5-fold based on model types using $\bmL$: Mean and standard deviation of the correlation coefficients, obtained as in Table~\ref{tab:group-result-regression_mean_and_std}, but using the matrix $\bmL$ instead of the matrix $\bmQ$ in the ridge regression of  (\ref{eq:ridge}) in Section~\ref{sec:predict-performance}.}
    \label{tab:group-result-regression_mean_and_std_using_L}
\end{table*}

\section{Details of Model Performance Prediction} \label{app:model_pred}

This section provides additional details on the prediction of benchmark scores using model coordinates, as discussed in Section~\ref{sec:predict-performance}.

\subsection{Details of ridge regression}\label{app:model_pred_training_details}
As described in Section~\ref{sec:setting_for_ridge}, ridge regression requires setting a regularization strength parameter, $\alpha$. 
To determine $\alpha$ from $\{10^1,\ldots,10^9\}$, we performed a five-fold cross-validation within each training dataset\footnote{The training dataset consisted of four out of the five folds obtained by splitting the dataset for each benchmark task.}.
As a post-processing step, we clipped the predicted scores to the range $[0, 100]$.

For the setting where the target variable $\bm{v}\in\mathbb{R}^K$ was replaced with the mean log-likelihood $(\bar \ell_1,\ldots,\bar \ell_K)\in\mathbb{R}^K$, we searched for $\alpha$ within $\{10^{-4},\ldots,10^4\}$ and did not apply clipping as a post-processing step.

\subsection{Details of prediction results}\label{app:model_pred_result_details}
Figure~\ref{fig:group_5fold_split_all_tasks} shows scatter plots of predicted scores and actual benchmark scores for each benchmark task, as well as for 6-TaskMean and the mean log-likelihood. 
As in Fig.~\ref{fig:group_5fold_split_average_task}, the scatter plots show strong correlations for all six benchmark tasks and for the mean log-likelihood.

To account for randomness, we ran five different data splits based on model types when predicting each benchmark score. 
As explained in Section~\ref{sec:predict-performance}, the final predicted score was the average of these five runs. 
Additionaly, for each split, we computed the correlation coefficients between the predicted scores and the actual benchmark scores. 
Table~\ref{tab:group-result-regression_mean_and_std} presents their mean and standard deviation, and shows a similar trend as Table~\ref{tab:group-result-regression}.

\subsection{Results from different settings}\label{app:model_pred_comp}
To investigate potential data leakage due to model types, Table~\ref{tab:result-regression_mean_and_std} shows the correlation coefficients using five random data splits.
These results demonstrate higher correlation coefficients across all tasks compared to those obtained using splits based on model types (Table~\ref{tab:group-result-regression_mean_and_std}), suggesting that randomly splitting models may unintentionally simplify the prediction task due to leakage from model types.

As shown in~(\ref{eq:ridge}) in Section~\ref{sec:predict-performance}, we performed ridge regression using the double-centered log-likelihood matrix $\bmQ$, derived from the log-likelihood matrix $\bmL$.
To assess the effect of replacing $\bmQ$ with $\bmL$, we repeated the analysis using $\bmL$ and report the resulting means and standard deviations of the correlation coefficients in Table~\ref{tab:group-result-regression_mean_and_std_using_L}.
The results obtained with $\bmL$ (Table~\ref{tab:group-result-regression_mean_and_std_using_L}) show nearly the same trend as those with $\bmQ$ (Table~\ref{tab:group-result-regression_mean_and_std}).
Note, however, that the mean of each row of $\bmL$ equals the mean log‑likelihood itself, so this specific target can be reproduced exactly. Consequently, the mean correlation coefficients are $1.000$ for both Pearson's $r$ and Spearman's $\rho$.

\section{Model List} \label{app:model_list}
Table~\ref{tab:model_list} lists the 1,018 models used in this study, sorted alphabetically by their names.
The BibTeX entries cited for each model were determined through the following procedure. 

First, we extracted the BibTeX entries available in each model's Hugging Face model card\footnote{\url{https://github.com/huggingface/huggingface_hub}.}.
If the BibTeX entry was missing a year of publication, we filled it in with the model's creation date\footnote{\url{https://github.com/sciunto-org/python-bibtexparser}.}. 
Additionally, we generated BibTeX entries using the arXiv IDs found in the model card tags by querying the arXiv API\footnote{\url{https://github.com/lukasschwab/arxiv.py}.}. This process resulted in a set of BibTeX entries for each model. 

Next, we manually checked pairs of different BibTeX entries where the title similarity\footnote{We used Python's \texttt{difflib.SequenceMatcher}.} was high, or the authors matched, to determine whether they corresponded to the same source.
This step allowed us to create groups of BibTeX entries that were considered identical.

Then, for each BibTeX group, we selected a representative entry as follows. 
Within each group, the entry most frequently cited by the models was chosen as the representative. 
If multiple candidates met this criterion, we prioritized BibTeX entries generated from arXiv IDs when available. 
If no such entry existed, we selected the one with the longest string. 

Finally, we replaced each model's BibTeX entry with the representative entry from its corresponding group. 
Any selected BibTeX entry that contained typos or formatting errors was manually corrected based on compilation errors. 
If the author information was incomplete, we corrected it manually by checking the source.

Note that for \texttt{google/codegemma-2b} and \texttt{deepseek-ai/deepseek-llm-7b-base} in Table~\ref{tab:nearest-neighbor-models}, as well as for \texttt{deepseek-ai/deepseek-coder-1.3b-base} and \texttt{mistralai/Mistral-7B-v0.3} in Table~\ref{tab:nearest-neighbor-16}, we manually prepared the BibTeX entries for citation based on their respective sources. We also used the same BibTeX entries for all other models that were considered to be of the same type.

\onecolumn
\input{model_list}

\end{document}

%% file: model_list.tex
{\scriptsize
\begin{longtable}{r@{\hspace{1em}}l@{\hspace{1em}}l@{\hspace{1em}}r@{\hspace{1em}}r@{\hspace{1em}}r@{\hspace{1em}}r@{\hspace{1em}}r}
\caption{List of 1018 models. ``ID'' denotes the alphabetical index; ``Model Name'' denotes the name of the model; ``Model Type'' denotes the classification defined in this paper; ``Size'' denotes the size of the model (B: billion); ``Date'' denotes the date of model creation; ``DLs'' denotes the total number of downloads; $\bar{\ell}_i$ denotes the mean log-likelihood; ``Task'' denotes the mean of the 6 benchmark scores (i.e., 6-TaskMean).}\\

\toprule
ID & Model Name & Model Type & Size & Date & DLs & $\bar{\ell}_i$ & Task\\
\midrule
\endfirsthead
    
\toprule
ID & Model Name & Model Type & Size & Date & DLs & $\bar{\ell}_i$ & Task\\
\midrule
\endhead
1 & \parbox[t]{7.5cm}{aaditya/Llama3-OpenBioLLM-8B~\cite{arxiv:2212.13138,arxiv:2305.09617,arxiv:2303.13375,arxiv:2305.18290,OpenBioLLMs,arxiv:2402.07023}} & llama-3 & 8B & 2024-04-20 & 9308 & -590.14 & 54.06 \\
2 & \parbox[t]{7.5cm}{aari1995/germeo-7b-laser} & mistral & 7B & 2024-01-09 & 2351 & -686.18 & 62.82 \\
3 & \parbox[t]{7.5cm}{abacusai/bigstral-12b-32k} & mistral & 12B & 2024-03-06 & 5216 & -632.41 & 62.17 \\
4 & \parbox[t]{7.5cm}{abacusai/Giraffe-13b-32k-v3} & llama-2 & 13B & 2023-12-06 & 1214 & -546.12 & 57.24 \\
5 & \parbox[t]{7.5cm}{abacusai/Llama-3-Smaug-8B~\cite{arxiv:2402.13228}} & llama-3 & 8B & 2024-04-19 & 12873 & -565.82 & 64.61 \\
6 & \parbox[t]{7.5cm}{abacusai/Slerp-CM-mist-dpo} & mistral & 7B & 2024-01-03 & 4030 & -553.48 & 73.10 \\
7 & \parbox[t]{7.5cm}{abhinand/Llama-3-OpenBioMed-8B-slerp-v0.3} & llama-3 & 8B & 2024-05-02 & 2713 & -558.51 & 62.08 \\
8 & \parbox[t]{7.5cm}{abhinand/tamil-llama-7b-base-v0.1~\cite{arxiv:2311.05845}} & llama-2 & 7B & 2023-11-08 & 1382 & -877.00 & 44.52 \\
\end{longtable}
}

%% file: modelmap2025.bbl
\begin{thebibliography}{280}
\providecommand{\natexlab}[1]{#1}

\bibitem[{{01. AI} et~al.(2025){01. AI}, Young, Chen, Li, Huang, Zhang, Zhang, Wang, Li, Zhu, Chen, Chang, Yu, Liu, Liu, Yue, Yang, Yang, Xie, Huang, Hu, Ren, Niu, Nie, Li, Xu, Liu, Wang, Cai, Gu, Liu, and Dai}]{arxiv:2403.04652}
{01. AI}, Alex Young, Bei Chen, Chao Li, Chengen Huang, Ge~Zhang, Guanwei Zhang, Guoyin Wang, Heng Li, Jiangcheng Zhu, Jianqun Chen, Jing Chang, Kaidong Yu, Peng Liu, Qiang Liu, Shawn Yue, Senbin Yang, Shiming Yang, Wen Xie, and 13 others. 2025.
\newblock \href {https://arxiv.org/abs/2403.04652} {Yi: Open foundation models by 01.ai}.
\newblock \emph{Preprint}, arXiv:2403.04652.

\bibitem[{{42dot Inc.}(2023)}]{42dot2023llm}
{42dot Inc.} 2023.
\newblock \href {https://github.com/42dot/42dot\_LLM} {42dot llm: A series of large language model by 42dot}.

\bibitem[{Acharya(2023)}]{rishiraj2023catppt}
Rishiraj Acharya. 2023.
\newblock \href {https://huggingface.co/rishiraj/CatPPT} {Catppt}.

\bibitem[{{AI@Meta}(2024)}]{llama3modelcard}
{AI@Meta}. 2024.
\newblock \href {https://github.com/meta-llama/llama3/blob/main/MODEL\_CARD.md} {Llama 3 model card}.

\bibitem[{Ainslie et~al.(2023)Ainslie, Lee-Thorp, de~Jong, Zemlyanskiy, Lebrón, and Sanghai}]{arxiv:2305.13245}
Joshua Ainslie, James Lee-Thorp, Michiel de~Jong, Yury Zemlyanskiy, Federico Lebrón, and Sumit Sanghai. 2023.
\newblock \href {https://arxiv.org/abs/2305.13245} {Gqa: Training generalized multi-query transformer models from multi-head checkpoints}.
\newblock \emph{Preprint}, arXiv:2305.13245.

\bibitem[{{AI@Waktaverse}(2024)}]{waktaversellama3modelcard}
{AI@Waktaverse}. 2024.
\newblock \href {https://huggingface.co/PathFinderKR/Waktaverse-Llama-3-KO-8B-Instruct} {Waktaverse llama 3 model card}.

\bibitem[{Almazrouei et~al.(2023)Almazrouei, Alobeidli, Alshamsi, Cappelli, Cojocaru, Debbah, Goffinet, Heslow, Launay, Malartic, Noune, Pannier, and Penedo}]{falcon40b}
Ebtesam Almazrouei, Hamza Alobeidli, Abdulaziz Alshamsi, Alessandro Cappelli, Ruxandra Cojocaru, Merouane Debbah, Etienne Goffinet, Daniel Heslow, Julien Launay, Quentin Malartic, Badreddine Noune, Baptiste Pannier, and Guilherme Penedo. 2023.
\newblock {Falcon-40B}: an open large language model with state-of-the-art performance.

\bibitem[{Alves et~al.(2024)Alves, Pombal, Guerreiro, Martins, Alves, Farajian, Peters, Rei, Fernandes, Agrawal, Colombo, de~Souza, and Martins}]{arxiv:2402.17733}
Duarte~M. Alves, José Pombal, Nuno~M. Guerreiro, Pedro~H. Martins, João Alves, Amin Farajian, Ben Peters, Ricardo Rei, Patrick Fernandes, Sweta Agrawal, Pierre Colombo, José G.~C. de~Souza, and André F.~T. Martins. 2024.
\newblock \href {https://arxiv.org/abs/2402.17733} {Tower: An open multilingual large language model for translation-related tasks}.
\newblock \emph{Preprint}, arXiv:2402.17733.

\bibitem[{Amari(1982)}]{amari1982-10.1214/aos/1176345779}
Shun-Ichi Amari. 1982.
\newblock \href {https://doi.org/10.1214/aos/1176345779} {{Differential Geometry of Curved Exponential Families-Curvatures and Information Loss}}.
\newblock \emph{The Annals of Statistics}, 10(2):357 -- 385.

\bibitem[{Amari(1998)}]{amari1998natural}
Shun-Ichi Amari. 1998.
\newblock Natural gradient works efficiently in learning.
\newblock \emph{Neural computation}, 10:251--276.

\bibitem[{Anand et~al.(2023)Anand, Nussbaum, Duderstadt, Schmidt, and Mulyar}]{gpt4all}
Yuvanesh Anand, Zach Nussbaum, Brandon Duderstadt, Benjamin Schmidt, and Andriy Mulyar. 2023.
\newblock \href {https://github.com/nomic-ai/gpt4all} {Gpt4all: Training an assistant-style chatbot with large scale data distillation from gpt-3.5-turbo}.

\bibitem[{Andonian et~al.(2021)Andonian, Anthony, Biderman, Black, Gali, Gao, Hallahan, Levy-Kramer, Leahy, Nestler, Parker, Pieler, Purohit, Songz, Phil, and Weinbach}]{gpt-neox-library}
Alex Andonian, Quentin Anthony, Stella Biderman, Sid Black, Preetham Gali, Leo Gao, Eric Hallahan, Josh Levy-Kramer, Connor Leahy, Lucas Nestler, Kip Parker, Michael Pieler, Shivanshu Purohit, Tri Songz, Wang Phil, and Samuel Weinbach. 2021.
\newblock \href {https://doi.org/10.5281/zenodo.5879544} {{GPT-NeoX: Large Scale Autoregressive Language Modeling in PyTorch}}.

\bibitem[{Asai et~al.(2023)Asai, Wu, Wang, Sil, and Hajishirzi}]{arxiv:2310.11511}
Akari Asai, Zeqiu Wu, Yizhong Wang, Avirup Sil, and Hannaneh Hajishirzi. 2023.
\newblock \href {https://arxiv.org/abs/2310.11511} {Self-rag: Learning to retrieve, generate, and critique through self-reflection}.
\newblock \emph{Preprint}, arXiv:2310.11511.

\bibitem[{Austin et~al.(2021)Austin, Odena, Nye, Bosma, Michalewski, Dohan, Jiang, Cai, Terry, Le, and Sutton}]{arxiv:2108.07732}
Jacob Austin, Augustus Odena, Maxwell Nye, Maarten Bosma, Henryk Michalewski, David Dohan, Ellen Jiang, Carrie Cai, Michael Terry, Quoc Le, and Charles Sutton. 2021.
\newblock \href {https://arxiv.org/abs/2108.07732} {Program synthesis with large language models}.
\newblock \emph{Preprint}, arXiv:2108.07732.

\bibitem[{Azerbayev et~al.(2024)Azerbayev, Schoelkopf, Paster, Santos, McAleer, Jiang, Deng, Biderman, and Welleck}]{arxiv:2310.10631}
Zhangir Azerbayev, Hailey Schoelkopf, Keiran Paster, Marco~Dos Santos, Stephen McAleer, Albert~Q. Jiang, Jia Deng, Stella Biderman, and Sean Welleck. 2024.
\newblock \href {https://arxiv.org/abs/2310.10631} {Llemma: An open language model for mathematics}.
\newblock \emph{Preprint}, arXiv:2310.10631.

\bibitem[{Ba et~al.(2016)Ba, Kiros, and Hinton}]{arxiv:1607.06450}
Jimmy~Lei Ba, Jamie~Ryan Kiros, and Geoffrey~E. Hinton. 2016.
\newblock \href {https://arxiv.org/abs/1607.06450} {Layer normalization}.
\newblock \emph{Preprint}, arXiv:1607.06450.

\bibitem[{Bai et~al.(2023)Bai, Bai, Chu, Cui, Dang, Deng, Fan, Ge, Han, Huang, Hui, Ji, Li, Lin, Lin, Liu, Liu, Lu, Lu, Ma, Men, Ren, Ren, Tan, Tan, Tu, Wang, Wang, Wang, Wu, Xu, Xu, Yang, Yang, Yang, Yang, Yao, Yu, Yuan, Yuan, Zhang, Zhang, Zhang, Zhang, Zhou, Zhou, Zhou, and Zhu}]{arxiv:2309.16609}
Jinze Bai, Shuai Bai, Yunfei Chu, Zeyu Cui, Kai Dang, Xiaodong Deng, Yang Fan, Wenbin Ge, Yu~Han, Fei Huang, Binyuan Hui, Luo Ji, Mei Li, Junyang Lin, Runji Lin, Dayiheng Liu, Gao Liu, Chengqiang Lu, Keming Lu, and 29 others. 2023.
\newblock \href {https://arxiv.org/abs/2309.16609} {Qwen technical report}.
\newblock \emph{Preprint}, arXiv:2309.16609.

\bibitem[{Balachandran(2023)}]{arxiv:2311.05845}
Abhinand Balachandran. 2023.
\newblock \href {https://arxiv.org/abs/2311.05845} {Tamil-llama: A new tamil language model based on llama 2}.
\newblock \emph{Preprint}, arXiv:2311.05845.

\bibitem[{Barndorff-Nielsen(2014)}]{barndorff2014information}
Ole Barndorff-Nielsen. 2014.
\newblock \emph{Information and exponential families: in statistical theory}.
\newblock John Wiley \& Sons.

\bibitem[{Basile et~al.(2023)Basile, Musacchio, Polignano, Siciliani, Fiameni, and Semeraro}]{arxiv:2312.09993}
Pierpaolo Basile, Elio Musacchio, Marco Polignano, Lucia Siciliani, Giuseppe Fiameni, and Giovanni Semeraro. 2023.
\newblock \href {https://arxiv.org/abs/2312.09993} {Llamantino: Llama 2 models for effective text generation in italian language}.
\newblock \emph{Preprint}, arXiv:2312.09993.

\bibitem[{Bavarian et~al.(2022)Bavarian, Jun, Tezak, Schulman, McLeavey, Tworek, and Chen}]{arxiv:2207.14255}
Mohammad Bavarian, Heewoo Jun, Nikolas Tezak, John Schulman, Christine McLeavey, Jerry Tworek, and Mark Chen. 2022.
\newblock \href {https://arxiv.org/abs/2207.14255} {Efficient training of language models to fill in the middle}.
\newblock \emph{Preprint}, arXiv:2207.14255.

\bibitem[{Beeching et~al.(2023)Beeching, Fourrier, Habib, Han, Lambert, Rajani, Sanseviero, Tunstall, and Wolf}]{open-llm-leaderboard}
Edward Beeching, Clémentine Fourrier, Nathan Habib, Sheon Han, Nathan Lambert, Nazneen Rajani, Omar Sanseviero, Lewis Tunstall, and Thomas Wolf. 2023.
\newblock \href {https://huggingface.co/spaces/HuggingFaceH4/open\_llm\_leaderboard} {Open llm leaderboard}.

\bibitem[{Beltagy et~al.(2020)Beltagy, Peters, and Cohan}]{arxiv:2004.05150}
Iz~Beltagy, Matthew~E. Peters, and Arman Cohan. 2020.
\newblock \href {https://arxiv.org/abs/2004.05150} {Longformer: The long-document transformer}.
\newblock \emph{Preprint}, arXiv:2004.05150.

\bibitem[{Bereska and Gavves(2024)}]{bereska2024mechanistic}
Leonard Bereska and Stratis Gavves. 2024.
\newblock \href {https://openreview.net/forum?id=ePUVetPKu6} {Mechanistic interpretability for {AI} safety - a review}.
\newblock \emph{Transactions on Machine Learning Research}.
\newblock Survey Certification, Expert Certification.

\bibitem[{Biderman et~al.(2022)Biderman, Bicheno, and Gao}]{arxiv:2201.07311}
Stella Biderman, Kieran Bicheno, and Leo Gao. 2022.
\newblock \href {https://arxiv.org/abs/2201.07311} {Datasheet for the pile}.
\newblock \emph{Preprint}, arXiv:2201.07311.

\bibitem[{Biderman et~al.(2023)Biderman, Schoelkopf, Anthony, Bradley, O'Brien, Hallahan, Khan, Purohit, Prashanth, Raff, Skowron, Sutawika, and van~der Wal}]{arxiv:2304.01373}
Stella Biderman, Hailey Schoelkopf, Quentin Anthony, Herbie Bradley, Kyle O'Brien, Eric Hallahan, Mohammad~Aflah Khan, Shivanshu Purohit, USVSN~Sai Prashanth, Edward Raff, Aviya Skowron, Lintang Sutawika, and Oskar van~der Wal. 2023.
\newblock \href {https://arxiv.org/abs/2304.01373} {Pythia: A suite for analyzing large language models across training and scaling}.
\newblock \emph{Preprint}, arXiv:2304.01373.

\bibitem[{Bisk et~al.(2019)Bisk, Zellers, Bras, Gao, and Choi}]{arxiv:1911.11641}
Yonatan Bisk, Rowan Zellers, Ronan~Le Bras, Jianfeng Gao, and Yejin Choi. 2019.
\newblock \href {https://arxiv.org/abs/1911.11641} {Piqa: Reasoning about physical commonsense in natural language}.
\newblock \emph{Preprint}, arXiv:1911.11641.

\bibitem[{Black et~al.(2022)Black, Biderman, Hallahan, Anthony, Gao, Golding, He, Leahy, McDonell, Phang, Pieler, Prashanth, Purohit, Reynolds, Tow, Wang, and Weinbach}]{arxiv:2204.06745}
Sid Black, Stella Biderman, Eric Hallahan, Quentin Anthony, Leo Gao, Laurence Golding, Horace He, Connor Leahy, Kyle McDonell, Jason Phang, Michael Pieler, USVSN~Sai Prashanth, Shivanshu Purohit, Laria Reynolds, Jonathan Tow, Ben Wang, and Samuel Weinbach. 2022.
\newblock \href {https://arxiv.org/abs/2204.06745} {Gpt-neox-20b: An open-source autoregressive language model}.
\newblock \emph{Preprint}, arXiv:2204.06745.

\bibitem[{Black et~al.(2021)Black, Leo, Wang, Leahy, and Biderman}]{gpt-neo}
Sid Black, Gao Leo, Phil Wang, Connor Leahy, and Stella Biderman. 2021.
\newblock \href {https://doi.org/10.5281/zenodo.5297715} {{GPT-Neo: Large Scale Autoregressive Language Modeling with Mesh-Tensorflow}}.
\newblock {If you use this software, please cite it using these metadata.}

\bibitem[{Borg and Groenen(2005)}]{borg2005modern}
Ingwer Borg and Patrick J.~F. Groenen. 2005.
\newblock \href {https://doi.org/10.1007/0-387-28981-X} {\emph{Modern Multidimensional Scaling: Theory and Applications}}, 2 edition.
\newblock Springer Series in Statistics. Springer, New York, NY.

\bibitem[{Brown et~al.(2020)Brown, Mann, Ryder, Subbiah, Kaplan, Dhariwal, Neelakantan, Shyam, Sastry, Askell, Agarwal, Herbert-Voss, Krueger, Henighan, Child, Ramesh, Ziegler, Wu, Winter, Hesse, Chen, Sigler, Litwin, Gray, Chess, Clark, Berner, McCandlish, Radford, Sutskever, and Amodei}]{arxiv:2005.14165}
Tom~B. Brown, Benjamin Mann, Nick Ryder, Melanie Subbiah, Jared Kaplan, Prafulla Dhariwal, Arvind Neelakantan, Pranav Shyam, Girish Sastry, Amanda Askell, Sandhini Agarwal, Ariel Herbert-Voss, Gretchen Krueger, Tom Henighan, Rewon Child, Aditya Ramesh, Daniel~M. Ziegler, Jeffrey Wu, Clemens Winter, and 12 others. 2020.
\newblock \href {https://arxiv.org/abs/2005.14165} {Language models are few-shot learners}.
\newblock \emph{Preprint}, arXiv:2005.14165.

\bibitem[{{CeADAR}(2023)}]{ceadar-2023}
{CeADAR}. 2023.
\newblock \href {https://doi.org/10.57967/hf/1405} {Financeconnect-13b (revision 5f7841d)}.

\bibitem[{Chaudhary(2023)}]{codealpaca}
Sahil Chaudhary. 2023.
\newblock \href {https://github.com/sahil280114/codealpaca} {Code alpaca: An instruction-following llama model for code generation}.

\bibitem[{Chen et~al.(2024{\natexlab{a}})Chen, Li, Yan, Wang, Gunaratna, Yadav, Tang, Srinivasan, Zhou, Huang, and Jin}]{arxiv:2307.08701}
Lichang Chen, Shiyang Li, Jun Yan, Hai Wang, Kalpa Gunaratna, Vikas Yadav, Zheng Tang, Vijay Srinivasan, Tianyi Zhou, Heng Huang, and Hongxia Jin. 2024{\natexlab{a}}.
\newblock \href {https://arxiv.org/abs/2307.08701} {Alpagasus: Training a better alpaca with fewer data}.
\newblock \emph{Preprint}, arXiv:2307.08701.

\bibitem[{Chen et~al.(2021)Chen, Tworek, Jun, Yuan, de~Oliveira~Pinto, Kaplan, Edwards, Burda, Joseph, Brockman, Ray, Puri, Krueger, Petrov, Khlaaf, Sastry, Mishkin, Chan, Gray, Ryder, Pavlov, Power, Kaiser, Bavarian, Winter, Tillet, Such, Cummings, Plappert, Chantzis, Barnes, Herbert-Voss, Guss, Nichol, Paino, Tezak, Tang, Babuschkin, Balaji, Jain, Saunders, Hesse, Carr, Leike, Achiam, Misra, Morikawa, Radford, Knight, Brundage, Murati, Mayer, Welinder, McGrew, Amodei, McCandlish, Sutskever, and Zaremba}]{arxiv:2107.03374}
Mark Chen, Jerry Tworek, Heewoo Jun, Qiming Yuan, Henrique~Ponde de~Oliveira~Pinto, Jared Kaplan, Harri Edwards, Yuri Burda, Nicholas Joseph, Greg Brockman, Alex Ray, Raul Puri, Gretchen Krueger, Michael Petrov, Heidy Khlaaf, Girish Sastry, Pamela Mishkin, Brooke Chan, Scott Gray, and 39 others. 2021.
\newblock \href {https://arxiv.org/abs/2107.03374} {Evaluating large language models trained on code}.
\newblock \emph{Preprint}, arXiv:2107.03374.

\bibitem[{Chen et~al.(2023{\natexlab{a}})Chen, Liang, Huang, Real, Wang, Liu, Pham, Dong, Luong, Hsieh, Lu, and Le}]{arxiv:2302.06675}
Xiangning Chen, Chen Liang, Da~Huang, Esteban Real, Kaiyuan Wang, Yao Liu, Hieu Pham, Xuanyi Dong, Thang Luong, Cho-Jui Hsieh, Yifeng Lu, and Quoc~V. Le. 2023{\natexlab{a}}.
\newblock \href {https://arxiv.org/abs/2302.06675} {Symbolic discovery of optimization algorithms}.
\newblock \emph{Preprint}, arXiv:2302.06675.

\bibitem[{Chen et~al.(2024{\natexlab{b}})Chen, Qian, Tang, Lai, Liu, Han, and Jia}]{arxiv:2309.12307}
Yukang Chen, Shengju Qian, Haotian Tang, Xin Lai, Zhijian Liu, Song Han, and Jiaya Jia. 2024{\natexlab{b}}.
\newblock \href {https://arxiv.org/abs/2309.12307} {Longlora: Efficient fine-tuning of long-context large language models}.
\newblock \emph{Preprint}, arXiv:2309.12307.

\bibitem[{Chen et~al.(2023{\natexlab{b}})Chen, Yu, Qian, Tang, Lai, Liu, Han, and Jia}]{long-alpaca}
Yukang Chen, Shaozuo Yu, Shengju Qian, Haotian Tang, Xin Lai, Zhijian Liu, Song Han, and Jiaya Jia. 2023{\natexlab{b}}.
\newblock \href {https://github.com/dvlab-research/LongLoRA} {Long alpaca: Long-context instruction-following models}.

\bibitem[{Chiang et~al.(2023)Chiang, Li, Lin, Sheng, Wu, Zhang, Zheng, Zhuang, Zhuang, Gonzalez, Stoica, and Xing}]{vicuna2023}
Wei-Lin Chiang, Zhuohan Li, Zi~Lin, Ying Sheng, Zhanghao Wu, Hao Zhang, Lianmin Zheng, Siyuan Zhuang, Yonghao Zhuang, Joseph~E. Gonzalez, Ion Stoica, and Eric~P. Xing. 2023.
\newblock \href {https://vicuna.lmsys.org} {Vicuna: An open-source chatbot impressing gpt-4 with 90\%* chatgpt quality}.

\bibitem[{Chiang et~al.(2024)Chiang, Zheng, Sheng, Angelopoulos, Li, Li, Zhang, Zhu, Jordan, Gonzalez, and Stoica}]{chiang2024chatbot}
Wei-Lin Chiang, Lianmin Zheng, Ying Sheng, Anastasios~Nikolas Angelopoulos, Tianle Li, Dacheng Li, Hao Zhang, Banghua Zhu, Michael Jordan, Joseph~E. Gonzalez, and Ion Stoica. 2024.
\newblock \href {https://arxiv.org/abs/2403.04132} {Chatbot arena: An open platform for evaluating llms by human preference}.
\newblock \emph{Preprint}, arXiv:2403.04132.

\bibitem[{chiliu(2023)}]{mamba-gpt-3b-v4}
chiliu. 2023.
\newblock \href {https://huggingface.co/CobraMamba/mamba-gpt-3b-v4} {Mamba-gpt-3b-v4}.

\bibitem[{Chollet(2019)}]{arxiv:1911.01547}
François Chollet. 2019.
\newblock \href {https://arxiv.org/abs/1911.01547} {On the measure of intelligence}.
\newblock \emph{Preprint}, arXiv:1911.01547.

\bibitem[{Clark et~al.(2019)Clark, Lee, Chang, Kwiatkowski, Collins, and Toutanova}]{arxiv:1905.10044}
Christopher Clark, Kenton Lee, Ming-Wei Chang, Tom Kwiatkowski, Michael Collins, and Kristina Toutanova. 2019.
\newblock \href {https://arxiv.org/abs/1905.10044} {Boolq: Exploring the surprising difficulty of natural yes/no questions}.
\newblock \emph{Preprint}, arXiv:1905.10044.

\bibitem[{Clark et~al.(2018)Clark, Cowhey, Etzioni, Khot, Sabharwal, Schoenick, and Tafjord}]{arxiv:1803.05457}
Peter Clark, Isaac Cowhey, Oren Etzioni, Tushar Khot, Ashish Sabharwal, Carissa Schoenick, and Oyvind Tafjord. 2018.
\newblock \href {https://arxiv.org/abs/1803.05457} {Think you have solved question answering? try arc, the ai2 reasoning challenge}.
\newblock \emph{Preprint}, arXiv:1803.05457.

\bibitem[{Cobbe et~al.(2021)Cobbe, Kosaraju, Bavarian, Chen, Jun, Kaiser, Plappert, Tworek, Hilton, Nakano, Hesse, and Schulman}]{arxiv:2110.14168}
Karl Cobbe, Vineet Kosaraju, Mohammad Bavarian, Mark Chen, Heewoo Jun, Lukasz Kaiser, Matthias Plappert, Jerry Tworek, Jacob Hilton, Reiichiro Nakano, Christopher Hesse, and John Schulman. 2021.
\newblock \href {https://arxiv.org/abs/2110.14168} {Training verifiers to solve math word problems}.
\newblock \emph{Preprint}, arXiv:2110.14168.

\bibitem[{{CodeGemma Team} et~al.(2024){CodeGemma Team}, Zhao, Hui, Howland, Nguyen, Zuo, Hu, Choquette-Choo, Shen, Kelley, Bansal, Vilnis, Wirth, Michel, Choy, Joshi, Kumar, Hashmi, Agrawal, Gong, Fine, Warkentin, Hartman, Ni, Korevec, Schaefer, and Huffman}]{arxiv:2406.11409}
{CodeGemma Team}, Heri Zhao, Jeffrey Hui, Joshua Howland, Nam Nguyen, Siqi Zuo, Andrea Hu, Christopher~A. Choquette-Choo, Jingyue Shen, Joe Kelley, Kshitij Bansal, Luke Vilnis, Mateo Wirth, Paul Michel, Peter Choy, Pratik Joshi, Ravin Kumar, Sarmad Hashmi, Shubham Agrawal, and 8 others. 2024.
\newblock \href {https://arxiv.org/abs/2406.11409} {Codegemma: Open code models based on gemma}.
\newblock \emph{Preprint}, arXiv:2406.11409.

\bibitem[{Conover et~al.(2023)Conover, Hayes, Mathur, Xie, Wan, Shah, Ghodsi, Wendell, Zaharia, and Xin}]{DatabricksBlog2023DollyV2}
Mike Conover, Matt Hayes, Ankit Mathur, Jianwei Xie, Jun Wan, Sam Shah, Ali Ghodsi, Patrick Wendell, Matei Zaharia, and Reynold Xin. 2023.
\newblock \href {https://www.databricks.com/blog/2023/04/12/dolly-first-open-commercially-viable-instruction-tuned-llm} {Free dolly: Introducing the world's first truly open instruction-tuned llm}.

\bibitem[{Cui et~al.(2024)Cui, Yuan, Ding, Yao, He, Zhu, Ni, Xie, Xie, Lin, Liu, and Sun}]{arxiv:2310.01377}
Ganqu Cui, Lifan Yuan, Ning Ding, Guanming Yao, Bingxiang He, Wei Zhu, Yuan Ni, Guotong Xie, Ruobing Xie, Yankai Lin, Zhiyuan Liu, and Maosong Sun. 2024.
\newblock \href {https://arxiv.org/abs/2310.01377} {Ultrafeedback: Boosting language models with scaled ai feedback}.
\newblock \emph{Preprint}, arXiv:2310.01377.

\bibitem[{Dai et~al.(2024)Dai, Deng, Zhao, Xu, Gao, Chen, Li, Zeng, Yu, Wu, Xie, Li, Huang, Luo, Ruan, Sui, and Liang}]{arxiv:2401.06066}
Damai Dai, Chengqi Deng, Chenggang Zhao, R.~X. Xu, Huazuo Gao, Deli Chen, Jiashi Li, Wangding Zeng, Xingkai Yu, Y.~Wu, Zhenda Xie, Y.~K. Li, Panpan Huang, Fuli Luo, Chong Ruan, Zhifang Sui, and Wenfeng Liang. 2024.
\newblock \href {https://arxiv.org/abs/2401.06066} {Deepseekmoe: Towards ultimate expert specialization in mixture-of-experts language models}.
\newblock \emph{Preprint}, arXiv:2401.06066.

\bibitem[{Dale et~al.(2023)Dale, Lian, Goodson, Wang, Pentland, Cook, Vong, and "Teknium"}]{dale2023llongorca13b}
Alpin Dale, Wing Lian, Bleys Goodson, Guan Wang, Eugene Pentland, Austin Cook, Chanvichet Vong, and "Teknium". 2023.
\newblock \href {https://huggingface.co/Open-Orca/LlongOrca-7B-16k} {Llongorca13b: Llama2-13b model instruct-tuned for long context on filtered openorcav1 gpt-4 dataset}.

\bibitem[{Dao(2023)}]{dao2023flashattention2}
Tri Dao. 2023.
\newblock Flash{A}ttention-2: Faster attention with better parallelism and work partitioning.

\bibitem[{Dao et~al.(2022)Dao, Fu, Ermon, Rudra, and Ré}]{arxiv:2205.14135}
Tri Dao, Daniel~Y. Fu, Stefano Ermon, Atri Rudra, and Christopher Ré. 2022.
\newblock \href {https://arxiv.org/abs/2205.14135} {Flashattention: Fast and memory-efficient exact attention with io-awareness}.
\newblock \emph{Preprint}, arXiv:2205.14135.

\bibitem[{De et~al.(2024)De, Smith, Fernando, Botev, Cristian-Muraru, Gu, Haroun, Berrada, Chen, Srinivasan, Desjardins, Doucet, Budden, Teh, Pascanu, Freitas, and Gulcehre}]{arxiv:2402.19427}
Soham De, Samuel~L. Smith, Anushan Fernando, Aleksandar Botev, George Cristian-Muraru, Albert Gu, Ruba Haroun, Leonard Berrada, Yutian Chen, Srivatsan Srinivasan, Guillaume Desjardins, Arnaud Doucet, David Budden, Yee~Whye Teh, Razvan Pascanu, Nando~De Freitas, and Caglar Gulcehre. 2024.
\newblock \href {https://arxiv.org/abs/2402.19427} {Griffin: Mixing gated linear recurrences with local attention for efficient language models}.
\newblock \emph{Preprint}, arXiv:2402.19427.

\bibitem[{{DeciAI Research Team}(2023)}]{DeciFoundationModels}
{DeciAI Research Team}. 2023.
\newblock \href {https://huggingface.co/Deci/DeciLM-7B-instruct} {Decilm-7b-instruct}.

\bibitem[{{DeepSeek-AI} et~al.(2024{\natexlab{a}}){DeepSeek-AI}, Bi, Chen, Chen, Chen, Dai, Deng, Ding, Dong, Du, Fu, Gao, Gao, Gao, Ge, Guan, Guo, Guo, Hao, Hao, He, Hu, Huang, Li, Li, Li, Li, Li, Liang, Lin, Liu, Liu, Liu, Liu, Liu, Liu, Lu, Lu, Luo, Ma, Nie, Pei, Piao, Qiu, Qu, Ren, Ren, Ruan, Sha, Shao, Song, Su, Sun, Sun, Tang, Wang, Wang, Wang, Wang, Wang, Wu, Wu, Xie, Xie, Xie, Xiong, Xu, Xu, Xu, Yang, You, Yu, Yu, Zhang, Zhang, Zhang, Zhang, Zhang, Zhang, Zhang, Zhang, Zhao, Zhao, Zhou, Zhou, Zhu, and Zou}]{arxiv:2401.02954}
{DeepSeek-AI}, Xiao Bi, Deli Chen, Guanting Chen, Shanhuang Chen, Damai Dai, Chengqi Deng, Honghui Ding, Kai Dong, Qiushi Du, Zhe Fu, Huazuo Gao, Kaige Gao, Wenjun Gao, Ruiqi Ge, Kang Guan, Daya Guo, Jianzhong Guo, Guangbo Hao, and 68 others. 2024{\natexlab{a}}.
\newblock \href {https://arxiv.org/abs/2401.02954} {Deepseek llm: Scaling open-source language models with longtermism}.
\newblock \emph{Preprint}, arXiv:2401.02954.

\bibitem[{{DeepSeek-AI} et~al.(2025){DeepSeek-AI}, Guo, Yang, Zhang, Song, Zhang, Xu, Zhu, Ma, Wang, Bi, Zhang, Yu, Wu, Wu, Gou, Shao, Li, Gao, Liu, Xue, Wang, Wu, Feng, Lu, Zhao, Deng, Zhang, Ruan, Dai, Chen, Ji, Li, Lin, Dai, Luo, Hao, Chen, Li, Zhang, Bao, Xu, Wang, Ding, Xin, Gao, Qu, Li, Guo, Li, Wang, Chen, Yuan, Qiu, Li, Cai, Ni, Liang, Chen, Dong, Hu, Gao, Guan, Huang, Yu, Wang, Zhang, Zhao, Wang, Zhang, Xu, Xia, Zhang, Zhang, Tang, Li, Wang, Li, Tian, Huang, Zhang, Wang, Chen, Du, Ge, Zhang, Pan, Wang, Chen, Jin, Chen, Lu, Zhou, Chen, Ye, Wang, Yu, Zhou, Pan, Li, Zhou, Wu, Ye, Yun, Pei, Sun, Wang, Zeng, Zhao, Liu, Liang, Gao, Yu, Zhang, Xiao, An, Liu, Wang, Chen, Nie, Cheng, Liu, Xie, Liu, Yang, Li, Su, Lin, Li, Jin, Shen, Chen, Sun, Wang, Song, Zhou, Wang, Shan, Li, Wang, Wei, Zhang, Xu, Li, Zhao, Sun, Wang, Yu, Zhang, Shi, Xiong, He, Piao, Wang, Tan, Ma, Liu, Guo, Ou, Wang, Gong, Zou, He, Xiong, Luo, You, Liu, Zhou, Zhu, Xu, Huang, Li, Zheng, Zhu, Ma, Tang, Zha, Yan, Ren, Ren, Sha, Fu, Xu, Xie,
  Zhang, Hao, Ma, Yan, Wu, Gu, Zhu, Liu, Li, Xie, Song, Pan, Huang, Xu, Zhang, and Zhang}]{arxiv:2501.12948}
{DeepSeek-AI}, Daya Guo, Dejian Yang, Haowei Zhang, Junxiao Song, Ruoyu Zhang, Runxin Xu, Qihao Zhu, Shirong Ma, Peiyi Wang, Xiao Bi, Xiaokang Zhang, Xingkai Yu, Yu~Wu, Z.~F. Wu, Zhibin Gou, Zhihong Shao, Zhuoshu Li, Ziyi Gao, and 181 others. 2025.
\newblock \href {https://arxiv.org/abs/2501.12948} {Deepseek-r1: Incentivizing reasoning capability in llms via reinforcement learning}.
\newblock \emph{Preprint}, arXiv:2501.12948.

\bibitem[{{DeepSeek-AI} et~al.(2024{\natexlab{b}}){DeepSeek-AI}, Liu, Feng, Wang, Wang, Liu, Zhao, Dengr, Ruan, Dai, Guo, Yang, Chen, Ji, Li, Lin, Luo, Hao, Chen, Li, Zhang, Xu, Yang, Zhang, Ding, Xin, Gao, Li, Qu, Cai, Liang, Guo, Ni, Li, Chen, Yuan, Qiu, Song, Dong, Gao, Guan, Wang, Zhang, Xu, Xia, Zhao, Zhang, Li, Wang, Zhang, Zhang, Tang, Li, Tian, Huang, Wang, Zhang, Zhu, Chen, Du, Chen, Jin, Ge, Pan, Xu, Chen, Li, Lu, Zhou, Chen, Wu, Ye, Ma, Wang, Zhou, Yu, Zhou, Zheng, Wang, Pei, Yuan, Sun, Xiao, Zeng, An, Liu, Liang, Gao, Zhang, Li, Jin, Wang, Bi, Liu, Wang, Shen, Chen, Chen, Nie, Sun, Wang, Liu, Xie, Yu, Song, Zhou, Yang, Lu, Su, Wu, Li, Wei, Zhu, Xu, Huang, Li, Zhao, Sun, Li, Wang, Zheng, Zhang, Xiong, Zhao, He, Tang, Piao, Dong, Tan, Liu, Wang, Guo, Zhu, Wang, Zou, Zha, Ma, Yan, You, Liu, Ren, Ren, Sha, Fu, Huang, Zhang, Xie, Hao, Shao, Wen, Xu, Zhang, Li, Wang, Gu, Li, and Xie}]{arxiv:2405.04434}
{DeepSeek-AI}, Aixin Liu, Bei Feng, Bin Wang, Bingxuan Wang, Bo~Liu, Chenggang Zhao, Chengqi Dengr, Chong Ruan, Damai Dai, Daya Guo, Dejian Yang, Deli Chen, Dongjie Ji, Erhang Li, Fangyun Lin, Fuli Luo, Guangbo Hao, Guanting Chen, and 138 others. 2024{\natexlab{b}}.
\newblock \href {https://arxiv.org/abs/2405.04434} {Deepseek-v2: A strong, economical, and efficient mixture-of-experts language model}.
\newblock \emph{Preprint}, arXiv:2405.04434.

\bibitem[{Dettmers et~al.(2022)Dettmers, Lewis, Shleifer, and Zettlemoyer}]{arxiv:2110.02861}
Tim Dettmers, Mike Lewis, Sam Shleifer, and Luke Zettlemoyer. 2022.
\newblock \href {https://arxiv.org/abs/2110.02861} {8-bit optimizers via block-wise quantization}.
\newblock \emph{Preprint}, arXiv:2110.02861.

\bibitem[{Dettmers et~al.(2023)Dettmers, Pagnoni, Holtzman, and Zettlemoyer}]{arxiv:2305.14314}
Tim Dettmers, Artidoro Pagnoni, Ari Holtzman, and Luke Zettlemoyer. 2023.
\newblock \href {https://arxiv.org/abs/2305.14314} {Qlora: Efficient finetuning of quantized llms}.
\newblock \emph{Preprint}, arXiv:2305.14314.

\bibitem[{Devine(2024)}]{arxiv:2405.12612}
Peter Devine. 2024.
\newblock \href {https://arxiv.org/abs/2405.12612} {Tagengo: A multilingual chat dataset}.
\newblock \emph{Preprint}, arXiv:2405.12612.

\bibitem[{Dhamala et~al.(2021)Dhamala, Sun, Kumar, Krishna, Pruksachatkun, Chang, and Gupta}]{arxiv:2101.11718}
Jwala Dhamala, Tony Sun, Varun Kumar, Satyapriya Krishna, Yada Pruksachatkun, Kai-Wei Chang, and Rahul Gupta. 2021.
\newblock \href {https://arxiv.org/abs/2101.11718} {Bold: Dataset and metrics for measuring biases in open-ended language generation}.
\newblock \emph{Preprint}, arXiv:2101.11718.

\bibitem[{Ding et~al.(2023)Ding, Chen, Xu, Qin, Zheng, Hu, Liu, Sun, and Zhou}]{arxiv:2305.14233}
Ning Ding, Yulin Chen, Bokai Xu, Yujia Qin, Zhi Zheng, Shengding Hu, Zhiyuan Liu, Maosong Sun, and Bowen Zhou. 2023.
\newblock \href {https://arxiv.org/abs/2305.14233} {Enhancing chat language models by scaling high-quality instructional conversations}.
\newblock \emph{Preprint}, arXiv:2305.14233.

\bibitem[{Do et~al.(2023)Do, Le, and Nguyen}]{allbyai2023ToRoLaMa}
Duy~Quang Do, Hoang Le, and Duc~Thang Nguyen. 2023.
\newblock \href {https://github.com/allbyai/ToRoLaMa} {Torolama: The vietnamese instruction-following and chat model}.

\bibitem[{Efron(1978)}]{efron1978geometry}
Bradley Efron. 1978.
\newblock The geometry of exponential families.
\newblock \emph{The Annals of Statistics}, 6:362--376.

\bibitem[{Efron(2022)}]{efron2022exponential}
Bradley Efron. 2022.
\newblock \emph{Exponential Families in Theory and Practice}.
\newblock Cambridge University Press.

\bibitem[{Ethayarajh et~al.(2023)Ethayarajh, Xu, Jurafsky, and Kiela}]{ethayarajh2023halos}
Kawin Ethayarajh, Winnie Xu, Dan Jurafsky, and Douwe Kiela. 2023.
\newblock \href {https://github.com/ContextualAI/HALOs/blob/main/assets/report.pdf} {Human-centered loss functions (halos)}.

\bibitem[{Fang et~al.(2025)Fang, Wang, Liu, Zhang, Jegelka, Gao, Ding, and Wang}]{fang2025wrong}
Lizhe Fang, Yifei Wang, Zhaoyang Liu, Chenheng Zhang, Stefanie Jegelka, Jinyang Gao, Bolin Ding, and Yisen Wang. 2025.
\newblock \href {https://openreview.net/forum?id=fL4qWkSmtM} {What is wrong with perplexity for long-context language modeling?}
\newblock In \emph{Proceedings of the Thirteenth International Conference on Learning Representations (ICLR)}.

\bibitem[{Fourrier et~al.(2024)Fourrier, Habib, Lozovskaya, Szafer, and Wolf}]{open-llm-leaderboard-v2}
Clémentine Fourrier, Nathan Habib, Alina Lozovskaya, Konrad Szafer, and Thomas Wolf. 2024.
\newblock Open llm leaderboard v2.
\newblock \url{https://huggingface.co/spaces/open-llm-leaderboard/open_llm_leaderboard}.

\bibitem[{Gallego(2024)}]{arxiv:2404.00495}
Victor Gallego. 2024.
\newblock \href {https://arxiv.org/abs/2404.00495} {Configurable safety tuning of language models with synthetic preference data}.
\newblock \emph{Preprint}, arXiv:2404.00495.

\bibitem[{Gao et~al.(2020)Gao, Biderman, Black, Golding, Hoppe, Foster, Phang, He, Thite, Nabeshima, Presser, and Leahy}]{arxiv:2101.00027}
Leo Gao, Stella Biderman, Sid Black, Laurence Golding, Travis Hoppe, Charles Foster, Jason Phang, Horace He, Anish Thite, Noa Nabeshima, Shawn Presser, and Connor Leahy. 2020.
\newblock \href {https://arxiv.org/abs/2101.00027} {The pile: An 800gb dataset of diverse text for language modeling}.
\newblock \emph{Preprint}, arXiv:2101.00027.

\bibitem[{Gao et~al.(2021{\natexlab{a}})Gao, Tow, Biderman, Black, DiPofi, Foster, Golding, Hsu, McDonell, Muennighoff, Phang, Reynolds, Tang, Thite, Wang, Wang, and Zou}]{eval-harness}
Leo Gao, Jonathan Tow, Stella Biderman, Sid Black, Anthony DiPofi, Charles Foster, Laurence Golding, Jeffrey Hsu, Kyle McDonell, Niklas Muennighoff, Jason Phang, Laria Reynolds, Eric Tang, Anish Thite, Ben Wang, Kevin Wang, and Andy Zou. 2021{\natexlab{a}}.
\newblock \href {https://doi.org/10.5281/zenodo.5371628} {A framework for few-shot language model evaluation}.

\bibitem[{Gao et~al.(2021{\natexlab{b}})Gao, Yao, and Chen}]{gao2021simcse}
Tianyu Gao, Xingcheng Yao, and Danqi Chen. 2021{\natexlab{b}}.
\newblock {SimCSE}: Simple contrastive learning of sentence embeddings.
\newblock In \emph{Empirical Methods in Natural Language Processing (EMNLP)}.

\bibitem[{Gehman et~al.(2020)Gehman, Gururangan, Sap, Choi, and Smith}]{arxiv:2009.11462}
Samuel Gehman, Suchin Gururangan, Maarten Sap, Yejin Choi, and Noah~A. Smith. 2020.
\newblock \href {https://arxiv.org/abs/2009.11462} {Realtoxicityprompts: Evaluating neural toxic degeneration in language models}.
\newblock \emph{Preprint}, arXiv:2009.11462.

\bibitem[{{Gemini Team} et~al.(2024){Gemini Team}, Anil, Borgeaud, Alayrac, Yu, Soricut, Schalkwyk, Dai, Hauth, Millican, Silver, Johnson, Antonoglou, Schrittwieser, Glaese, Chen, Pitler, Lillicrap, Lazaridou, Firat, Molloy, Isard, Barham, Hennigan, Lee, Viola, Reynolds, Xu, Doherty, Collins, Meyer, Rutherford, Moreira, Ayoub, Goel, Krawczyk, Du, Chi, Cheng, Ni, Shah, Kane, Chan, Faruqui, Severyn, Lin, Li, Cheng, Ittycheriah, Mahdieh, Chen, Sun, Tran, Bagri, Lakshminarayanan, Liu, Orban, Güra, Zhou, Song, Boffy, Ganapathy, Zheng, Choe, Ágoston Weisz, Zhu, Lu, Gopal, Kahn, Kula, Pitman, Shah, Taropa, Merey, Baeuml, Chen, Shafey, Zhang, Sercinoglu, Tucker, Piqueras, Krikun, Barr, Savinov, Danihelka, Roelofs, White, Andreassen, von Glehn, Yagati, Kazemi, Gonzalez, Khalman, Sygnowski, Frechette, Smith, Culp, Proleev, Luan, Chen, Lottes, Schucher, Lebron, Rrustemi, Clay, Crone, Kocisky, Zhao, Perz, Yu, Howard, Bloniarz, Rae, Lu, Sifre, Maggioni, Alcober, Garrette, Barnes, Thakoor, Austin, Barth-Maron, Wong,
  Joshi, Chaabouni, Fatiha, Ahuja, Tomar, Senter, Chadwick, Kornakov, Attaluri, Iturrate, Liu, Li, Cogan, Chen, Jia, Gu, Zhang, Grimstad, Hartman, Garcia, Pillai, Devlin, Laskin, de~Las~Casas, Valter, Tao, Blanco, Badia, Reitter, Chen, Brennan, Rivera, Brin, Iqbal, Surita, Labanowski, Rao, Winkler, Parisotto, Gu, Olszewska, Addanki, Miech, Louis, Teplyashin, Brown, Catt, Balaguer, Xiang, Wang, Ashwood, Briukhov, Webson, Ganapathy, Sanghavi, Kannan, Chang, Stjerngren, Djolonga, Sun, Bapna, Aitchison, Pejman, Michalewski, Yu, Wang, Love, Ahn, Bloxwich, Han, Humphreys, Sellam, Bradbury, Godbole, Samangooei, Damoc, Kaskasoli, Arnold, Vasudevan, Agrawal, Riesa, Lepikhin, Tanburn, Srinivasan, Lim, Hodkinson, Shyam, Ferret, Hand, Garg, Paine, Li, Li, Giang, Neitz, Abbas, York, Reid, Cole, Chowdhery, Das, Rogozińska, Nikolaev, Sprechmann, Nado, Zilka, Prost, He, Monteiro, Mishra, Welty, Newlan, Jia, Allamanis, Hu, de~Liedekerke, Gilmer, Saroufim, Rijhwani, Hou, Shrivastava, Baddepudi, Goldin, Ozturel, Cassirer, Xu,
  Sohn, Sachan, Amplayo, Swanson, Petrova, Narayan, Guez, Brahma, Landon, Patel, Zhao, Villela, Wang, Jia, Rahtz, Giménez, Yeung, Keeling, Georgiev, Mincu, Wu, Haykal, Saputro, Vodrahalli, Qin, Cankara, Sharma, Fernando, Hawkins, Neyshabur, Kim, Hutter, Agrawal, Castro-Ros, van~den Driessche, Wang, Yang, yiin Chang, Komarek, McIlroy, Lučić, Zhang, Farhan, Sharman, Natsev, Michel, Bansal, Qiao, Cao, Shakeri, Butterfield, Chung, Rubenstein, Agrawal, Mensch, Soparkar, Lenc, Chung, Pope, Maggiore, Kay, Jhakra, Wang, Maynez, Phuong, Tobin, Tacchetti, Trebacz, Robinson, Katariya, Riedel, Bailey, Xiao, Ghelani, Aroyo, Slone, Houlsby, Xiong, Yang, Gribovskaya, Adler, Wirth, Lee, Li, Kagohara, Pavagadhi, Bridgers, Bortsova, Ghemawat, Ahmed, Liu, Powell, Bolina, Iinuma, Zablotskaia, Besley, Chung, Dozat, Comanescu, Si, Greer, Su, Polacek, Kaufman, Tokumine, Hu, Buchatskaya, Miao, Elhawaty, Siddhant, Tomasev, Xing, Greer, Miller, Ashraf, Roy, Zhang, Ma, Filos, Besta, Blevins, Klimenko, Yeh, Changpinyo, Mu, Chang,
  Pajarskas, Muir, Cohen, Lan, Haridasan, Marathe, Hansen, Douglas, Samuel, Wang, Austin, Lan, Jiang, Chiu, Lorenzo, Sjösund, Cevey, Gleicher, Avrahami, Boral, Srinivasan, Selo, May, Aisopos, Hussenot, Soares, Baumli, Chang, Recasens, Caine, Pritzel, Pavetic, Pardo, Gergely, Frye, Ramasesh, Horgan, Badola, Kassner, Roy, Dyer, Campos, Tomala, Tang, Badawy, White, Mustafa, Lang, Jindal, Vikram, Gong, Caelles, Hemsley, Thornton, Feng, Stokowiec, Zheng, Thacker, Çağlar Ünlü, Zhang, Saleh, Svensson, Bileschi, Patil, Anand, Ring, Tsihlas, Vezer, Selvi, Shevlane, Rodriguez, Kwiatkowski, Daruki, Rong, Dafoe, FitzGerald, Gu-Lemberg, Khan, Hendricks, Pellat, Feinberg, Cobon-Kerr, Sainath, Rauh, Hashemi, Ives, Hasson, Noland, Cao, Byrd, Hou, Wang, Sottiaux, Paganini, Lespiau, Moufarek, Hassan, Shivakumar, van Amersfoort, Mandhane, Joshi, Goyal, Tung, Brock, Sheahan, Misra, Li, Rakićević, Dehghani, Liu, Mittal, Oh, Noury, Sezener, Huot, Lamm, Cao, Chen, Mudgal, Stella, Brooks, Vasudevan, Liu, Chain, Melinkeri,
  Cohen, Wang, Seymore, Zubkov, Goel, Yue, Krishnakumaran, Albert, Hurley, Sano, Mohananey, Joughin, Filonov, Kępa, Eldawy, Lim, Rishi, Badiezadegan, Bos, Chang, Jain, Padmanabhan, Puttagunta, Krishna, Baker, Kalb, Bedapudi, Kurzrok, Lei, Yu, Litvin, Zhou, Wu, Sobell, Siciliano, Papir, Neale, Bragagnolo, Toor, Chen, Anklin, Wang, Feng, Gholami, Ling, Liu, Walter, Moghaddam, Kishore, Adamek, Mercado, Mallinson, Wandekar, Cagle, Ofek, Garrido, Lombriser, Mukha, Sun, Mohammad, Matak, Qian, Peswani, Janus, Yuan, Schelin, David, Garg, He, Duzhyi, Älgmyr, Lottaz, Li, Yadav, Xu, Chinien, Shivanna, Chuklin, Li, Spadine, Wolfe, Mohamed, Das, Dai, He, von Dincklage, Upadhyay, Maurya, Chi, Krause, Salama, Rabinovitch, M, Selvan, Dektiarev, Ghiasi, Guven, Gupta, Liu, Sharma, Shtacher, Paul, Akerlund, Aubet, Huang, Zhu, Zhu, Teixeira, Fritze, Bertolini, Marinescu, Bölle, Paulus, Gupta, Latkar, Chang, Sanders, Wilson, Wu, Tan, Thiet, Doshi, Lall, Mishra, Chen, Luong, Benjamin, Lee, Andrejczuk, Rabiej, Ranjan, Styrc,
  Yin, Simon, Harriott, Bansal, Robsky, Bacon, Greene, Mirylenka, Zhou, Sarvana, Goyal, Andermatt, Siegler, Horn, Israel, Pongetti, Chen, Selvatici, Silva, Wang, Tolins, Guu, Yogev, Cai, Agostini, Shah, Nguyen, Donnaile, Pereira, Friso, Stambler, Kurzrok, Kuang, Romanikhin, Geller, Yan, Jang, Lee, Fica, Malmi, Tan, Banica, Balle, Pham, Huang, Avram, Shi, Singh, Hidey, Ahuja, Saxena, Dooley, Potharaju, O'Neill, Gokulchandran, Foley, Zhao, Dusenberry, Liu, Mehta, Kotikalapudi, Safranek-Shrader, Goodman, Kessinger, Globen, Kolhar, Gorgolewski, Ibrahim, Song, Eichenbaum, Brovelli, Potluri, Lahoti, Baetu, Ghorbani, Chen, Crawford, Pal, Sridhar, Gurita, Mujika, Petrovski, Cedoz, Li, Chen, Santo, Goyal, Punjabi, Kappaganthu, Kwak, LV, Velury, Choudhury, Hall, Shah, Figueira, Thomas, Lu, Zhou, Kumar, Jurdi, Chikkerur, Ma, Yu, Kwak, Ähdel, Rajayogam, Choma, Liu, Barua, Ji, Park, Hellendoorn, Bailey, Bilal, Zhou, Khatir, Sutton, Rzadkowski, Macintosh, Shagin, Medina, Liang, Zhou, Shah, Bi, Dankovics, Banga, Lehmann,
  Bredesen, Lin, Hoffmann, Lai, Chung, Yang, Balani, Bražinskas, Sozanschi, Hayes, Alcalde, Makarov, Chen, Stella, Snijders, Mandl, Kärrman, Nowak, Wu, Dyck, Vaidyanathan, R, Mallet, Rudominer, Johnston, Mittal, Udathu, Christensen, Verma, Irving, Santucci, Elsayed, Davoodi, Georgiev, Tenney, Hua, Cideron, Leurent, Alnahlawi, Georgescu, Wei, Zheng, Scandinaro, Jiang, Snoek, Sundararajan, Wang, Ontiveros, Karo, Cole, Rajashekhar, Tumeh, Ben-David, Jain, Uesato, Datta, Bunyan, Wu, Zhang, Stanczyk, Zhang, Steiner, Naskar, Azzam, Johnson, Paszke, Chiu, Elias, Mohiuddin, Muhammad, Miao, Lee, Vieillard, Park, Zhang, Stanway, Garmon, Karmarkar, Dong, Lee, Kumar, Zhou, Evens, Isaac, Irving, Loper, Fink, Arkatkar, Chen, Shafran, Petrychenko, Chen, Jia, Levskaya, Zhu, Grabowski, Mao, Magni, Yao, Snaider, Casagrande, Palmer, Suganthan, Castaño, Giannoumis, Kim, Rybiński, Sreevatsa, Prendki, Soergel, Goedeckemeyer, Gierke, Jafari, Gaba, Wiesner, Wright, Wei, Vashisht, Kulizhskaya, Hoover, Le, Li, Iwuanyanwu, Liu,
  Ramirez, Khorlin, Cui, LIN, Wu, Aguilar, Pallo, Chakladar, Perng, Abellan, Zhang, Dasgupta, Kushman, Penchev, Repina, Wu, van~der Weide, Ponnapalli, Kaplan, Simsa, Li, Dousse, Yang, Piper, Ie, Pasumarthi, Lintz, Vijayakumar, Andor, Valenzuela, Lui, Paduraru, Peng, Lee, Zhang, Greene, Nguyen, Kurylowicz, Hardin, Dixon, Janzer, Choo, Feng, Zhang, Singhal, Du, McKinnon, Antropova, Bolukbasi, Keller, Reid, Finchelstein, Raad, Crocker, Hawkins, Dadashi, Gaffney, Franko, Bulanova, Leblond, Chung, Askham, Cobo, Xu, Fischer, Xu, Sorokin, Alberti, Lin, Evans, Dimitriev, Forbes, Banarse, Tung, Omernick, Bishop, Sterneck, Jain, Xia, Amid, Piccinno, Wang, Banzal, Mankowitz, Polozov, Krakovna, Brown, Bateni, Duan, Firoiu, Thotakuri, Natan, Geist, tan Girgin, Li, Ye, Roval, Tojo, Kwong, Lee-Thorp, Yew, Sinopalnikov, Ramos, Mellor, Sharma, Wu, Miller, Sonnerat, Vnukov, Greig, Beattie, Caveness, Bai, Eisenschlos, Korchemniy, Tsai, Jasarevic, Kong, Dao, Zheng, Liu, Yang, Zhu, Teh, Sanmiya, Gladchenko, Trdin, Toyama, Rosen,
  Tavakkol, Xue, Elkind, Woodman, Carpenter, Papamakarios, Kemp, Kafle, Grunina, Sinha, Talbert, Wu, Owusu-Afriyie, Du, Thornton, Pont-Tuset, Narayana, Li, Fatehi, Wieting, Ajmeri, Uria, Ko, Knight, Héliou, Niu, Gu, Pang, Li, Levine, Stolovich, Santamaria-Fernandez, Goenka, Yustalim, Strudel, Elqursh, Deck, Lee, Li, Levin, Hoffmann, Holtmann-Rice, Bachem, Arora, Koh, Yeganeh, Põder, Tariq, Sun, Ionita, Seyedhosseini, Tafti, Liu, Gulati, Liu, Ye, Chrzaszcz, Wang, Sethi, Li, Brown, Singh, Fan, Parisi, Stanton, Koverkathu, Choquette-Choo, Li, Lu, Ittycheriah, Shroff, Varadarajan, Bahargam, Willoughby, Gaddy, Desjardins, Cornero, Robenek, Mittal, Albrecht, Shenoy, Moiseev, Jacobsson, Ghaffarkhah, Rivière, Walton, Crepy, Parrish, Zhou, Farabet, Radebaugh, Srinivasan, van~der Salm, Fidjeland, Scellato, Latorre-Chimoto, Klimczak-Plucińska, Bridson, de~Cesare, Hudson, Mendolicchio, Walker, Morris, Mauger, Guseynov, Reid, Odoom, Loher, Cotruta, Yenugula, Grewe, Petrushkina, Duerig, Sanchez, Yadlowsky, Shen,
  Globerson, Webb, Dua, Li, Bhupatiraju, Hurt, Qureshi, Agarwal, Shani, Eyal, Khare, Belle, Wang, Tekur, Kale, Wei, Sang, Saeta, Liechty, Sun, Zhao, Lee, Nayak, Fritz, Vuyyuru, Aslanides, Vyas, Wicke, Ma, Eltyshev, Martin, Cate, Manyika, Amiri, Kim, Xiong, Kang, Luisier, Tripuraneni, Madras, Guo, Waters, Wang, Ainslie, Baldridge, Zhang, Pruthi, Bauer, Yang, Mansour, Gelman, Xu, Polovets, Liu, Cai, Chen, Sheng, Xue, Ozair, Angermueller, Li, Sinha, Wang, Wiesinger, Koukoumidis, Tian, Iyer, Gurumurthy, Goldenson, Shah, Blake, Yu, Urbanowicz, Palomaki, Fernando, Durden, Mehta, Momchev, Rahimtoroghi, Georgaki, Raul, Ruder, Redshaw, Lee, Zhou, Jalan, Li, Hechtman, Schuh, Nasr, Milan, Mikulik, Franco, Green, Nguyen, Kelley, Mahendru, Hu, Howland, Vargas, Hui, Bansal, Rao, Ghiya, Wang, Ye, Sarr, Preston, Elish, Li, Kaku, Gupta, Pasupat, Juan, Someswar, M., Chen, Amini, Fabrikant, Chu, Dong, Muthal, Buthpitiya, Jauhari, Hua, Khandelwal, Hitron, Ren, Rinaldi, Drath, Dabush, Jiang, Godhia, Sachs, Chen, Fan, Taitelbaum,
  Noga, Dai, Wang, Liang, Hamer, Ferng, Elkind, Atias, Lee, Listík, Carlen, van~de Kerkhof, Pikus, Zaher, Müller, Zykova, Stefanec, Gatsko, Hirnschall, Sethi, Xu, Ahuja, Tsai, Stefanoiu, Feng, Dhandhania, Katyal, Gupta, Parulekar, Pitta, Zhao, Bhatia, Bhavnani, Alhadlaq, Li, Danenberg, Tu, Pine, Filippova, Ghosh, Limonchik, Urala, Lanka, Clive, Sun, Li, Wu, Hongtongsak, Li, Thakkar, Omarov, Majmundar, Alverson, Kucharski, Patel, Jain, Zabelin, Pelagatti, Kohli, Kumar, Kim, Sankar, Shah, Ramachandruni, Zeng, Bariach, Weidinger, Vu, Andreev, He, Hui, Kashem, Subramanya, Hsiao, Hassabis, Kavukcuoglu, Sadovsky, Le, Strohman, Wu, Petrov, Dean, and Vinyals}]{arxiv:2312.11805}
{Gemini Team}, Rohan Anil, Sebastian Borgeaud, Jean-Baptiste Alayrac, Jiahui Yu, Radu Soricut, Johan Schalkwyk, Andrew~M. Dai, Anja Hauth, Katie Millican, David Silver, Melvin Johnson, Ioannis Antonoglou, Julian Schrittwieser, Amelia Glaese, Jilin Chen, Emily Pitler, Timothy Lillicrap, Angeliki Lazaridou, and 1331 others. 2024.
\newblock \href {https://arxiv.org/abs/2312.11805} {Gemini: A family of highly capable multimodal models}.
\newblock \emph{Preprint}, arXiv:2312.11805.

\bibitem[{Geng and Liu(2023)}]{openlm2023openllama}
Xinyang Geng and Hao Liu. 2023.
\newblock \href {https://github.com/openlm-research/open\_llama} {Openllama: An open reproduction of llama}.

\bibitem[{Gou et~al.(2024)Gou, Shao, Gong, Shen, Yang, Huang, Duan, and Chen}]{arxiv:2309.17452}
Zhibin Gou, Zhihong Shao, Yeyun Gong, Yelong Shen, Yujiu Yang, Minlie Huang, Nan Duan, and Weizhu Chen. 2024.
\newblock \href {https://arxiv.org/abs/2309.17452} {Tora: A tool-integrated reasoning agent for mathematical problem solving}.
\newblock \emph{Preprint}, arXiv:2309.17452.

\bibitem[{Granziol et~al.(2021)Granziol, Zohren, and Roberts}]{arxiv:2006.09092}
Diego Granziol, Stefan Zohren, and Stephen Roberts. 2021.
\newblock \href {https://arxiv.org/abs/2006.09092} {Learning rates as a function of batch size: A random matrix theory approach to neural network training}.
\newblock \emph{Preprint}, arXiv:2006.09092.

\bibitem[{{Griffin Team} et~al.(2024){Griffin Team}, De, Smith, Fernando, Botev, Muraru, Haroun, and et~al.}]{recurrentgemma-2024}
{Griffin Team}, Soham De, Samuel~L Smith, Anushan Fernando, Alex Botev, George-Christian Muraru, Ruba Haroun, and Leonard~Berrada et~al. 2024.
\newblock Recurrentgemma.

\bibitem[{Groeneveld et~al.(2024)Groeneveld, Beltagy, Walsh, Bhagia, Kinney, Tafjord, Jha, Ivison, Magnusson, Wang, Arora, Atkinson, Authur, Chandu, Cohan, Dumas, Elazar, Gu, Hessel, Khot, Merrill, Morrison, Muennighoff, Naik, Nam, Peters, Pyatkin, Ravichander, Schwenk, Shah, Smith, Strubell, Subramani, Wortsman, Dasigi, Lambert, Richardson, Zettlemoyer, Dodge, Lo, Soldaini, Smith, and Hajishirzi}]{arxiv:2402.00838}
Dirk Groeneveld, Iz~Beltagy, Pete Walsh, Akshita Bhagia, Rodney Kinney, Oyvind Tafjord, Ananya~Harsh Jha, Hamish Ivison, Ian Magnusson, Yizhong Wang, Shane Arora, David Atkinson, Russell Authur, Khyathi~Raghavi Chandu, Arman Cohan, Jennifer Dumas, Yanai Elazar, Yuling Gu, Jack Hessel, and 24 others. 2024.
\newblock \href {https://arxiv.org/abs/2402.00838} {Olmo: Accelerating the science of language models}.
\newblock \emph{Preprint}, arXiv:2402.00838.

\bibitem[{Guo et~al.(2024)Guo, Zhu, Yang, Xie, Dong, Zhang, Chen, Bi, Wu, Li, Luo, Xiong, and Liang}]{arxiv:2401.14196}
Daya Guo, Qihao Zhu, Dejian Yang, Zhenda Xie, Kai Dong, Wentao Zhang, Guanting Chen, Xiao Bi, Y.~Wu, Y.~K. Li, Fuli Luo, Yingfei Xiong, and Wenfeng Liang. 2024.
\newblock \href {https://arxiv.org/abs/2401.14196} {Deepseek-coder: When the large language model meets programming -- the rise of code intelligence}.
\newblock \emph{Preprint}, arXiv:2401.14196.

\bibitem[{Harries(2023)}]{orca-mini-v2-ger-7b}
Jan~Philipp Harries. 2023.
\newblock \href {https://huggingface.co/jphme/orca\_mini\_v2\_ger\_7b} {orca\_mini\_v2\_ger\_7b: An explain tuned llama-7b model based on orca mini v2 and adapted to german language}.

\bibitem[{Hartvigsen et~al.(2022)Hartvigsen, Gabriel, Palangi, Sap, Ray, and Kamar}]{arxiv:2203.09509}
Thomas Hartvigsen, Saadia Gabriel, Hamid Palangi, Maarten Sap, Dipankar Ray, and Ece Kamar. 2022.
\newblock \href {https://arxiv.org/abs/2203.09509} {Toxigen: A large-scale machine-generated dataset for adversarial and implicit hate speech detection}.
\newblock \emph{Preprint}, arXiv:2203.09509.

\bibitem[{Hendrycks et~al.(2021{\natexlab{a}})Hendrycks, Burns, Basart, Zou, Mazeika, Song, and Steinhardt}]{arxiv:2009.03300}
Dan Hendrycks, Collin Burns, Steven Basart, Andy Zou, Mantas Mazeika, Dawn Song, and Jacob Steinhardt. 2021{\natexlab{a}}.
\newblock \href {https://arxiv.org/abs/2009.03300} {Measuring massive multitask language understanding}.
\newblock \emph{Preprint}, arXiv:2009.03300.

\bibitem[{Hendrycks et~al.(2021{\natexlab{b}})Hendrycks, Burns, Kadavath, Arora, Basart, Tang, Song, and Steinhardt}]{arxiv:2103.03874}
Dan Hendrycks, Collin Burns, Saurav Kadavath, Akul Arora, Steven Basart, Eric Tang, Dawn Song, and Jacob Steinhardt. 2021{\natexlab{b}}.
\newblock \href {https://arxiv.org/abs/2103.03874} {Measuring mathematical problem solving with the math dataset}.
\newblock \emph{Preprint}, arXiv:2103.03874.

\bibitem[{Henry et~al.(2020)Henry, Dachapally, Pawar, and Chen}]{arxiv:2010.04245}
Alex Henry, Prudhvi~Raj Dachapally, Shubham Pawar, and Yuxuan Chen. 2020.
\newblock \href {https://arxiv.org/abs/2010.04245} {Query-key normalization for transformers}.
\newblock \emph{Preprint}, arXiv:2010.04245.

\bibitem[{Hong et~al.(2024)Hong, Lee, and Thorne}]{arxiv:2403.07691}
Jiwoo Hong, Noah Lee, and James Thorne. 2024.
\newblock \href {https://arxiv.org/abs/2403.07691} {Orpo: Monolithic preference optimization without reference model}.
\newblock \emph{Preprint}, arXiv:2403.07691.

\bibitem[{Horwitz et~al.(2025)Horwitz, Kurer, Kahana, Amar, and Hoshen}]{horwitz2025charting}
Eliahu Horwitz, Nitzan Kurer, Jonathan Kahana, Liel Amar, and Yedid Hoshen. 2025.
\newblock \href {https://arxiv.org/abs/2503.10633} {Charting and navigating hugging face's model atlas}.
\newblock \emph{Preprint}, arXiv:2503.10633.

\bibitem[{Houlsby et~al.(2019)Houlsby, Giurgiu, Jastrzebski, Morrone, de~Laroussilhe, Gesmundo, Attariyan, and Gelly}]{arxiv:1902.00751}
Neil Houlsby, Andrei Giurgiu, Stanislaw Jastrzebski, Bruna Morrone, Quentin de~Laroussilhe, Andrea Gesmundo, Mona Attariyan, and Sylvain Gelly. 2019.
\newblock \href {https://arxiv.org/abs/1902.00751} {Parameter-efficient transfer learning for nlp}.
\newblock \emph{Preprint}, arXiv:1902.00751.

\bibitem[{Hsu et~al.(2024)Hsu, Liu, Liao, Hsu, Chen, and Shiu}]{arxiv:2403.02712}
Chan-Jan Hsu, Chang-Le Liu, Feng-Ting Liao, Po-Chun Hsu, Yi-Chang Chen, and Da-Shan Shiu. 2024.
\newblock \href {https://arxiv.org/abs/2403.02712} {Breeze-7b technical report}.
\newblock \emph{Preprint}, arXiv:2403.02712.

\bibitem[{Hu et~al.(2022)Hu, Shen, Wallis, Allen-Zhu, Li, Wang, Wang, and Chen}]{hu2022lora}
Edward~J Hu, Yelong Shen, Phillip Wallis, Zeyuan Allen-Zhu, Yuanzhi Li, Shean Wang, Lu~Wang, and Weizhu Chen. 2022.
\newblock \href {https://openreview.net/forum?id=nZeVKeeFYf9} {Lo{RA}: Low-rank adaptation of large language models}.
\newblock In \emph{International Conference on Learning Representations}.

\bibitem[{{IDEA-CCNL}(2021)}]{Fengshenbang-LM}
{IDEA-CCNL}. 2021.
\newblock \href {https://github.com/IDEA-CCNL/Fengshenbang-LM} {Fengshenbang-lm}.

\bibitem[{Ilharco et~al.(2023)Ilharco, Ribeiro, Wortsman, Schmidt, Hajishirzi, and Farhadi}]{ilharco2023editing}
Gabriel Ilharco, Marco~Tulio Ribeiro, Mitchell Wortsman, Ludwig Schmidt, Hannaneh Hajishirzi, and Ali Farhadi. 2023.
\newblock \href {https://openreview.net/forum?id=6t0Kwf8-jrj} {Editing models with task arithmetic}.
\newblock In \emph{The Eleventh International Conference on Learning Representations}.

\bibitem[{"interstellarninja" et~al.(2024)"interstellarninja", "Teknium", "theemozilla", "karan4d", and "huemin\_art"}]{Hermes-2-Pro-Mistral-7B}
"interstellarninja", "Teknium", "theemozilla", "karan4d", and "huemin\_art". 2024.
\newblock \href {https://huggingface.co/NousResearch/Hermes-2-Pro-Mistral-7B} {Hermes-2-pro-mistral-7b}.

\bibitem[{Iyer et~al.(2023)Iyer, Lin, Pasunuru, Mihaylov, Simig, Yu, Shuster, Wang, Liu, Koura, Li, O'Horo, Pereyra, Wang, Dewan, Celikyilmaz, Zettlemoyer, and Stoyanov}]{arxiv:2212.12017}
Srinivasan Iyer, Xi~Victoria Lin, Ramakanth Pasunuru, Todor Mihaylov, Daniel Simig, Ping Yu, Kurt Shuster, Tianlu Wang, Qing Liu, Punit~Singh Koura, Xian Li, Brian O'Horo, Gabriel Pereyra, Jeff Wang, Christopher Dewan, Asli Celikyilmaz, Luke Zettlemoyer, and Ves Stoyanov. 2023.
\newblock \href {https://arxiv.org/abs/2212.12017} {Opt-iml: Scaling language model instruction meta learning through the lens of generalization}.
\newblock \emph{Preprint}, arXiv:2212.12017.

\bibitem[{Jain et~al.(2023)Jain, yeh Chiang, Wen, Kirchenbauer, Chu, Somepalli, Bartoldson, Kailkhura, Schwarzschild, Saha, Goldblum, Geiping, and Goldstein}]{arxiv:2310.05914}
Neel Jain, Ping yeh Chiang, Yuxin Wen, John Kirchenbauer, Hong-Min Chu, Gowthami Somepalli, Brian~R. Bartoldson, Bhavya Kailkhura, Avi Schwarzschild, Aniruddha Saha, Micah Goldblum, Jonas Geiping, and Tom Goldstein. 2023.
\newblock \href {https://arxiv.org/abs/2310.05914} {Neftune: Noisy embeddings improve instruction finetuning}.
\newblock \emph{Preprint}, arXiv:2310.05914.

\bibitem[{Jiang et~al.(2023)Jiang, Sablayrolles, Mensch, Bamford, Chaplot, de~las Casas, Bressand, Lengyel, Lample, Saulnier, Lavaud, Lachaux, Stock, Scao, Lavril, Wang, Lacroix, and Sayed}]{arxiv:2310.06825}
Albert~Q. Jiang, Alexandre Sablayrolles, Arthur Mensch, Chris Bamford, Devendra~Singh Chaplot, Diego de~las Casas, Florian Bressand, Gianna Lengyel, Guillaume Lample, Lucile Saulnier, Lélio~Renard Lavaud, Marie-Anne Lachaux, Pierre Stock, Teven~Le Scao, Thibaut Lavril, Thomas Wang, Timothée Lacroix, and William~El Sayed. 2023.
\newblock \href {https://arxiv.org/abs/2310.06825} {Mistral 7b}.
\newblock \emph{Preprint}, arXiv:2310.06825.

\bibitem[{Jiang et~al.(2024)Jiang, Li, Zhang, Huang, Lin, and Chen}]{arxiv:2310.00752}
Dongfu Jiang, Yishan Li, Ge~Zhang, Wenhao Huang, Bill~Yuchen Lin, and Wenhu Chen. 2024.
\newblock \href {https://arxiv.org/abs/2310.00752} {Tigerscore: Towards building explainable metric for all text generation tasks}.
\newblock \emph{Preprint}, arXiv:2310.00752.

\bibitem[{Joshi et~al.(2017)Joshi, Choi, Weld, and Zettlemoyer}]{arxiv:1705.03551}
Mandar Joshi, Eunsol Choi, Daniel~S. Weld, and Luke Zettlemoyer. 2017.
\newblock \href {https://arxiv.org/abs/1705.03551} {Triviaqa: A large scale distantly supervised challenge dataset for reading comprehension}.
\newblock \emph{Preprint}, arXiv:1705.03551.

\bibitem[{Kim et~al.(2024{\natexlab{a}})Kim, Kim, Song, Kim, Kim, Kim, and Park}]{arxiv:2403.19270}
Dahyun Kim, Yungi Kim, Wonho Song, Hyeonwoo Kim, Yunsu Kim, Sanghoon Kim, and Chanjun Park. 2024{\natexlab{a}}.
\newblock \href {https://arxiv.org/abs/2403.19270} {sdpo: Don't use your data all at once}.
\newblock \emph{Preprint}, arXiv:2403.19270.

\bibitem[{Kim et~al.(2024{\natexlab{b}})Kim, Park, Kim, Lee, Song, Kim, Kim, Kim, Lee, Kim, Ahn, Yang, Lee, Park, Gim, Cha, Lee, and Kim}]{arxiv:2312.15166}
Dahyun Kim, Chanjun Park, Sanghoon Kim, Wonsung Lee, Wonho Song, Yunsu Kim, Hyeonwoo Kim, Yungi Kim, Hyeonju Lee, Jihoo Kim, Changbae Ahn, Seonghoon Yang, Sukyung Lee, Hyunbyung Park, Gyoungjin Gim, Mikyoung Cha, Hwalsuk Lee, and Sunghun Kim. 2024{\natexlab{b}}.
\newblock \href {https://arxiv.org/abs/2312.15166} {Solar 10.7b: Scaling large language models with simple yet effective depth up-scaling}.
\newblock \emph{Preprint}, arXiv:2312.15166.

\bibitem[{Kim et~al.(2022)Kim, Jang, Kwon, and Davis}]{arxiv:2204.04541}
Dohyeong Kim, Myeongjun Jang, Deuk~Sin Kwon, and Eric Davis. 2022.
\newblock \href {https://arxiv.org/abs/2204.04541} {Kobest: Korean balanced evaluation of significant tasks}.
\newblock \emph{Preprint}, arXiv:2204.04541.

\bibitem[{Kim et~al.(2024{\natexlab{c}})Kim, Choi, and Jeong}]{arxiv:2402.14714}
Seungduk Kim, Seungtaek Choi, and Myeongho Jeong. 2024{\natexlab{c}}.
\newblock \href {https://arxiv.org/abs/2402.14714} {Efficient and effective vocabulary expansion towards multilingual large language models}.
\newblock \emph{Preprint}, arXiv:2402.14714.

\bibitem[{Ko et~al.(2023)Ko, Yang, Ryu, Choi, Yang, Hyun, Park, and Park}]{arxiv:2306.02254}
Hyunwoong Ko, Kichang Yang, Minho Ryu, Taekyoon Choi, Seungmu Yang, Jiwung Hyun, Sungho Park, and Kyubyong Park. 2023.
\newblock \href {https://arxiv.org/abs/2306.02254} {A technical report for polyglot-ko: Open-source large-scale korean language models}.
\newblock \emph{Preprint}, arXiv:2306.02254.

\bibitem[{Kojima et~al.(2023)Kojima, Gu, Reid, Matsuo, and Iwasawa}]{arxiv:2205.11916}
Takeshi Kojima, Shixiang~Shane Gu, Machel Reid, Yutaka Matsuo, and Yusuke Iwasawa. 2023.
\newblock \href {https://arxiv.org/abs/2205.11916} {Large language models are zero-shot reasoners}.
\newblock \emph{Preprint}, arXiv:2205.11916.

\bibitem[{Komatsuzaki et~al.(2023)Komatsuzaki, Puigcerver, Lee-Thorp, Ruiz, Mustafa, Ainslie, Tay, Dehghani, and Houlsby}]{arxiv:2212.05055}
Aran Komatsuzaki, Joan Puigcerver, James Lee-Thorp, Carlos~Riquelme Ruiz, Basil Mustafa, Joshua Ainslie, Yi~Tay, Mostafa Dehghani, and Neil Houlsby. 2023.
\newblock \href {https://arxiv.org/abs/2212.05055} {Sparse upcycling: Training mixture-of-experts from dense checkpoints}.
\newblock \emph{Preprint}, arXiv:2212.05055.

\bibitem[{Kweon et~al.(2024)Kweon, Kim, Kim, Im, Cho, Bae, Oh, Lee, Moon, You, Baek, Han, Jung, Jo, and Choi}]{arxiv:2309.00237}
Sunjun Kweon, Junu Kim, Jiyoun Kim, Sujeong Im, Eunbyeol Cho, Seongsu Bae, Jungwoo Oh, Gyubok Lee, Jong~Hak Moon, Seng~Chan You, Seungjin Baek, Chang~Hoon Han, Yoon~Bin Jung, Yohan Jo, and Edward Choi. 2024.
\newblock \href {https://arxiv.org/abs/2309.00237} {Publicly shareable clinical large language model built on synthetic clinical notes}.
\newblock \emph{Preprint}, arXiv:2309.00237.

\bibitem[{Köpf et~al.(2023)Köpf, Kilcher, von Rütte, Anagnostidis, Tam, Stevens, Barhoum, Duc, Stanley, Nagyfi, ES, Suri, Glushkov, Dantuluri, Maguire, Schuhmann, Nguyen, and Mattick}]{arxiv:2304.07327}
Andreas Köpf, Yannic Kilcher, Dimitri von Rütte, Sotiris Anagnostidis, Zhi-Rui Tam, Keith Stevens, Abdullah Barhoum, Nguyen~Minh Duc, Oliver Stanley, Richárd Nagyfi, Shahul ES, Sameer Suri, David Glushkov, Arnav Dantuluri, Andrew Maguire, Christoph Schuhmann, Huu Nguyen, and Alexander Mattick. 2023.
\newblock \href {https://arxiv.org/abs/2304.07327} {Openassistant conversations -- democratizing large language model alignment}.
\newblock \emph{Preprint}, arXiv:2304.07327.

\bibitem[{Labrak et~al.(2024)Labrak, Bazoge, Morin, Gourraud, Rouvier, and Dufour}]{arxiv:2402.10373}
Yanis Labrak, Adrien Bazoge, Emmanuel Morin, Pierre-Antoine Gourraud, Mickael Rouvier, and Richard Dufour. 2024.
\newblock \href {https://arxiv.org/abs/2402.10373} {Biomistral: A collection of open-source pretrained large language models for medical domains}.
\newblock \emph{Preprint}, arXiv:2402.10373.

\bibitem[{Lacoste et~al.(2019)Lacoste, Luccioni, Schmidt, and Dandres}]{arxiv:1910.09700}
Alexandre Lacoste, Alexandra Luccioni, Victor Schmidt, and Thomas Dandres. 2019.
\newblock \href {https://arxiv.org/abs/1910.09700} {Quantifying the carbon emissions of machine learning}.
\newblock \emph{Preprint}, arXiv:1910.09700.

\bibitem[{Lee et~al.(2024)Lee, Hunter, and Ruiz}]{arxiv:2308.07317}
Ariel~N. Lee, Cole~J. Hunter, and Nataniel Ruiz. 2024.
\newblock \href {https://arxiv.org/abs/2308.07317} {Platypus: Quick, cheap, and powerful refinement of llms}.
\newblock \emph{Preprint}, arXiv:2308.07317.

\bibitem[{Lee et~al.(2023)Lee, Hunter, Ruiz, Goodson, Lian, Wang, Pentland, Cook, Vong, and "Teknium"}]{hunterlee2023orcaplaty1}
Ariel~N. Lee, Cole~J. Hunter, Nataniel Ruiz, Bleys Goodson, Wing Lian, Guan Wang, Eugene Pentland, Austin Cook, Chanvichet Vong, and "Teknium". 2023.
\newblock \href {https://huggingface.co/Open-Orca/OpenOrca-Platypus2-13B} {Openorcaplatypus: Llama2-13b model instruct-tuned on filtered openorcav1 gpt-4 dataset and merged with divergent stem and logic dataset model}.

\bibitem[{Lee(2023)}]{l.-junbum-2023}
Junbum Lee. 2023.
\newblock \href {https://doi.org/10.57967/hf/1098} {llama-2-ko-7b (revision 4a9993e)}.

\bibitem[{Lee(2024{\natexlab{a}})}]{llama3koen}
Junbum Lee. 2024{\natexlab{a}}.
\newblock \href {https://huggingface.co/beomi/Llama-3-KoEn-8B} {Llama-3-koen}.

\bibitem[{Lee(2024{\natexlab{b}})}]{lee-junbum-2024}
Junbum Lee. 2024{\natexlab{b}}.
\newblock \href {https://doi.org/10.57967/hf/1708} {Yi-ko-6b (revision 205083a)}.

\bibitem[{Lee and Ahn(2024)}]{llama3cs-layertuning}
Jungwon Lee and Seungjun Ahn. 2024.
\newblock \href {https://huggingface.co/DBCM/Llama-3-instruction-constructionsafety-layertuning} {Llama-3-instruction-constructionsafety-layertuning}.

\bibitem[{Levine et~al.(2020)Levine, Kumar, Tucker, and Fu}]{arxiv:2005.01643}
Sergey Levine, Aviral Kumar, George Tucker, and Justin Fu. 2020.
\newblock \href {https://arxiv.org/abs/2005.01643} {Offline reinforcement learning: Tutorial, review, and perspectives on open problems}.
\newblock \emph{Preprint}, arXiv:2005.01643.

\bibitem[{Li et~al.(2023{\natexlab{a}})Li, Allal, Zi, Muennighoff, Kocetkov, Mou, Marone, Akiki, Li, Chim, Liu, Zheltonozhskii, Zhuo, Wang, Dehaene, Davaadorj, Lamy-Poirier, Monteiro, Shliazhko, Gontier, Meade, Zebaze, Yee, Umapathi, Zhu, Lipkin, Oblokulov, Wang, Murthy, Stillerman, Patel, Abulkhanov, Zocca, Dey, Zhang, Fahmy, Bhattacharyya, Yu, Singh, Luccioni, Villegas, Kunakov, Zhdanov, Romero, Lee, Timor, Ding, Schlesinger, Schoelkopf, Ebert, Dao, Mishra, Gu, Robinson, Anderson, Dolan-Gavitt, Contractor, Reddy, Fried, Bahdanau, Jernite, Ferrandis, Hughes, Wolf, Guha, von Werra, and de~Vries}]{arxiv:2305.06161}
Raymond Li, Loubna~Ben Allal, Yangtian Zi, Niklas Muennighoff, Denis Kocetkov, Chenghao Mou, Marc Marone, Christopher Akiki, Jia Li, Jenny Chim, Qian Liu, Evgenii Zheltonozhskii, Terry~Yue Zhuo, Thomas Wang, Olivier Dehaene, Mishig Davaadorj, Joel Lamy-Poirier, João Monteiro, Oleh Shliazhko, and 48 others. 2023{\natexlab{a}}.
\newblock \href {https://arxiv.org/abs/2305.06161} {Starcoder: may the source be with you!}
\newblock \emph{Preprint}, arXiv:2305.06161.

\bibitem[{Li et~al.(2023{\natexlab{b}})Li, Liu, Bian, Fang, Huang, Liu, Wang, and You}]{arxiv:2110.14883}
Shenggui Li, Hongxin Liu, Zhengda Bian, Jiarui Fang, Haichen Huang, Yuliang Liu, Boxiang Wang, and Yang You. 2023{\natexlab{b}}.
\newblock \href {https://arxiv.org/abs/2110.14883} {Colossal-ai: A unified deep learning system for large-scale parallel training}.
\newblock \emph{Preprint}, arXiv:2110.14883.

\bibitem[{Li et~al.(2023{\natexlab{c}})Li, Bubeck, Eldan, Giorno, Gunasekar, and Lee}]{arxiv:2309.05463}
Yuanzhi Li, Sébastien Bubeck, Ronen Eldan, Allie~Del Giorno, Suriya Gunasekar, and Yin~Tat Lee. 2023{\natexlab{c}}.
\newblock \href {https://arxiv.org/abs/2309.05463} {Textbooks are all you need ii: phi-1.5 technical report}.
\newblock \emph{Preprint}, arXiv:2309.05463.

\bibitem[{Li et~al.(2023{\natexlab{d}})Li, Li, Zhang, Dan, Jiang, and Zhang}]{arxiv:2303.14070}
Yunxiang Li, Zihan Li, Kai Zhang, Ruilong Dan, Steve Jiang, and You Zhang. 2023{\natexlab{d}}.
\newblock \href {https://arxiv.org/abs/2303.14070} {Chatdoctor: A medical chat model fine-tuned on a large language model meta-ai (llama) using medical domain knowledge}.
\newblock \emph{Preprint}, arXiv:2303.14070.

\bibitem[{Lian et~al.(2023{\natexlab{a}})Lian, Goodson, Pentland, Cook, Vong, and "Teknium"}]{OpenOrca-Preview1}
Wing Lian, Bleys Goodson, Eugene Pentland, Austin Cook, Chanvichet Vong, and "Teknium". 2023{\natexlab{a}}.
\newblock \href {https://huggingface.co/Open-Orca/OpenOrca-Preview1-13B} {Openorca\_preview1: A llama-13b model fine-tuned on small portion of openorcav1 dataset}.

\bibitem[{Lian et~al.(2023{\natexlab{b}})Lian, Goodson, Wang, Pentland, Cook, Vong, and "Teknium"}]{lian2023jackalope}
Wing Lian, Bleys Goodson, Guan Wang, Eugene Pentland, Austin Cook, Chanvichet Vong, and "Teknium". 2023{\natexlab{b}}.
\newblock \href {https://huggingface.co/openaccess-ai-collective/jackalope-7b} {Jackalope 7b: Mistral-7b model multi-turn chat tuned on filtered openorcav1 gpt-4 dataset}.

\bibitem[{Lian et~al.(2023{\natexlab{c}})Lian, Goodson, Wang, Pentland, Cook, Vong, and "Teknium"}]{lian2023llongorca7b}
Wing Lian, Bleys Goodson, Guan Wang, Eugene Pentland, Austin Cook, Chanvichet Vong, and "Teknium". 2023{\natexlab{c}}.
\newblock \href {https://huggingface.co/Open-Orca/LlongOrca-7B-16k} {Llongorca7b: Llama2-7b model instruct-tuned for long context on filtered openorcav1 gpt-4 dataset}.

\bibitem[{Lian et~al.(2023{\natexlab{d}})Lian, Goodson, Wang, Pentland, Cook, Vong, and "Teknium"}]{lian2023mistralorca1}
Wing Lian, Bleys Goodson, Guan Wang, Eugene Pentland, Austin Cook, Chanvichet Vong, and "Teknium". 2023{\natexlab{d}}.
\newblock \href {https://huggingface.co/Open-Orca/Mistral-7B-OpenOrca} {Mistralorca: Mistral-7b model instruct-tuned on filtered openorcav1 gpt-4 dataset}.

\bibitem[{Lian et~al.(2023{\natexlab{e}})Lian, Goodson, Wang, Pentland, Cook, Vong, and "Teknium"}]{lian2023mistralslimorca1}
Wing Lian, Bleys Goodson, Guan Wang, Eugene Pentland, Austin Cook, Chanvichet Vong, and "Teknium". 2023{\natexlab{e}}.
\newblock \href {https://huggingface.co/Open-Orca/Mistral-7B-SlimOrca} {Mistralslimorca: Mistral-7b model instruct-tuned on filtered, corrected, openorcav1 gpt-4 dataset}.

\bibitem[{Lian et~al.(2023{\natexlab{f}})Lian, Wang, Goodson, Pentland, Cook, Vong, and "Teknium"}]{SlimOrca}
Wing Lian, Guan Wang, Bleys Goodson, Eugene Pentland, Austin Cook, Chanvichet Vong, and "Teknium". 2023{\natexlab{f}}.
\newblock \href {https://huggingface.co/Open-Orca/SlimOrca} {Slimorca: An open dataset of gpt-4 augmented flan reasoning traces, with verification}.

\bibitem[{Lian et~al.(2023{\natexlab{g}})Lian, Wang, Goodson, Pentland, Cook, Vong, "Teknium", and Hoos}]{SlimOrcaDedup}
Wing Lian, Guan Wang, Bleys Goodson, Eugene Pentland, Austin Cook, Chanvichet Vong, "Teknium", and Nathan Hoos. 2023{\natexlab{g}}.
\newblock \href {https://huggingface.co/datasets/Open-Orca/SlimOrca-Dedup/} {Slimorca dedup: A deduplicated subset of slimorca}.

\bibitem[{Lin et~al.(2022{\natexlab{a}})Lin, Hilton, and Evans}]{arxiv:2109.07958}
Stephanie Lin, Jacob Hilton, and Owain Evans. 2022{\natexlab{a}}.
\newblock \href {https://arxiv.org/abs/2109.07958} {Truthfulqa: Measuring how models mimic human falsehoods}.
\newblock \emph{Preprint}, arXiv:2109.07958.

\bibitem[{Lin et~al.(2022{\natexlab{b}})Lin, Mihaylov, Artetxe, Wang, Chen, Simig, Ott, Goyal, Bhosale, Du, Pasunuru, Shleifer, Koura, Chaudhary, O'Horo, Wang, Zettlemoyer, Kozareva, Diab, Stoyanov, and Li}]{arxiv:2112.10668}
Xi~Victoria Lin, Todor Mihaylov, Mikel Artetxe, Tianlu Wang, Shuohui Chen, Daniel Simig, Myle Ott, Naman Goyal, Shruti Bhosale, Jingfei Du, Ramakanth Pasunuru, Sam Shleifer, Punit~Singh Koura, Vishrav Chaudhary, Brian O'Horo, Jeff Wang, Luke Zettlemoyer, Zornitsa Kozareva, Mona Diab, and 2 others. 2022{\natexlab{b}}.
\newblock \href {https://arxiv.org/abs/2112.10668} {Few-shot learning with multilingual language models}.
\newblock \emph{Preprint}, arXiv:2112.10668.

\bibitem[{Liu et~al.(2025)Liu, Yan, Zaharia, and Abbeel}]{arxiv:2402.08268}
Hao Liu, Wilson Yan, Matei Zaharia, and Pieter Abbeel. 2025.
\newblock \href {https://arxiv.org/abs/2402.08268} {World model on million-length video and language with blockwise ringattention}.
\newblock \emph{Preprint}, arXiv:2402.08268.

\bibitem[{Liu et~al.(2023{\natexlab{a}})Liu, Xie, Li, and Ma}]{liu2023same}
Hong Liu, Sang~Michael Xie, Zhiyuan Li, and Tengyu Ma. 2023{\natexlab{a}}.
\newblock \href {https://proceedings.mlr.press/v202/liu23ao.html} {Same pre-training loss, better downstream: Implicit bias matters for language models}.
\newblock In \emph{Proceedings of the 40th International Conference on Machine Learning (ICML)}, volume 202 of \emph{Proceedings of Machine Learning Research}, pages 22208--22242. PMLR.

\bibitem[{Liu et~al.(2023{\natexlab{b}})Liu, Lin, Hewitt, Paranjape, Bevilacqua, Petroni, and Liang}]{arxiv:2307.03172}
Nelson~F. Liu, Kevin Lin, John Hewitt, Ashwin Paranjape, Michele Bevilacqua, Fabio Petroni, and Percy Liang. 2023{\natexlab{b}}.
\newblock \href {https://arxiv.org/abs/2307.03172} {Lost in the middle: How language models use long contexts}.
\newblock \emph{Preprint}, arXiv:2307.03172.

\bibitem[{Liu et~al.(2024)Liu, Ping, Roy, Xu, Lee, Shoeybi, and Catanzaro}]{arxiv:2401.10225}
Zihan Liu, Wei Ping, Rajarshi Roy, Peng Xu, Chankyu Lee, Mohammad Shoeybi, and Bryan Catanzaro. 2024.
\newblock \href {https://arxiv.org/abs/2401.10225} {Chatqa: Surpassing gpt-4 on conversational qa and rag}.
\newblock \emph{Preprint}, arXiv:2401.10225.

\bibitem[{Longpre et~al.(2023)Longpre, Hou, Vu, Webson, Chung, Tay, Zhou, Le, Zoph, Wei, and Roberts}]{arxiv:2301.13688}
Shayne Longpre, Le~Hou, Tu~Vu, Albert Webson, Hyung~Won Chung, Yi~Tay, Denny Zhou, Quoc~V. Le, Barret Zoph, Jason Wei, and Adam Roberts. 2023.
\newblock \href {https://arxiv.org/abs/2301.13688} {The flan collection: Designing data and methods for effective instruction tuning}.
\newblock \emph{Preprint}, arXiv:2301.13688.

\bibitem[{Lozhkov et~al.(2024)Lozhkov, Li, Allal, Cassano, Lamy-Poirier, Tazi, Tang, Pykhtar, Liu, Wei, Liu, Tian, Kocetkov, Zucker, Belkada, Wang, Liu, Abulkhanov, Paul, Li, Li, Risdal, Li, Zhu, Zhuo, Zheltonozhskii, Dade, Yu, Krauß, Jain, Su, He, Dey, Abati, Chai, Muennighoff, Tang, Oblokulov, Akiki, Marone, Mou, Mishra, Gu, Hui, Dao, Zebaze, Dehaene, Patry, Xu, McAuley, Hu, Scholak, Paquet, Robinson, Anderson, Chapados, Patwary, Tajbakhsh, Jernite, Ferrandis, Zhang, Hughes, Wolf, Guha, von Werra, and de~Vries}]{arxiv:2402.19173}
Anton Lozhkov, Raymond Li, Loubna~Ben Allal, Federico Cassano, Joel Lamy-Poirier, Nouamane Tazi, Ao~Tang, Dmytro Pykhtar, Jiawei Liu, Yuxiang Wei, Tianyang Liu, Max Tian, Denis Kocetkov, Arthur Zucker, Younes Belkada, Zijian Wang, Qian Liu, Dmitry Abulkhanov, Indraneil Paul, and 47 others. 2024.
\newblock \href {https://arxiv.org/abs/2402.19173} {Starcoder 2 and the stack v2: The next generation}.
\newblock \emph{Preprint}, arXiv:2402.19173.

\bibitem[{Luo et~al.(2025)Luo, Sun, Xu, Zhao, Lou, Tao, Geng, Lin, Chen, Tang, and Zhang}]{arxiv:2308.09583}
Haipeng Luo, Qingfeng Sun, Can Xu, Pu~Zhao, Jianguang Lou, Chongyang Tao, Xiubo Geng, Qingwei Lin, Shifeng Chen, Yansong Tang, and Dongmei Zhang. 2025.
\newblock \href {https://arxiv.org/abs/2308.09583} {Wizardmath: Empowering mathematical reasoning for large language models via reinforced evol-instruct}.
\newblock \emph{Preprint}, arXiv:2308.09583.

\bibitem[{Luo et~al.(2023)Luo, Xu, Zhao, Sun, Geng, Hu, Tao, Ma, Lin, and Jiang}]{arxiv:2306.08568}
Ziyang Luo, Can Xu, Pu~Zhao, Qingfeng Sun, Xiubo Geng, Wenxiang Hu, Chongyang Tao, Jing Ma, Qingwei Lin, and Daxin Jiang. 2023.
\newblock \href {https://arxiv.org/abs/2306.08568} {Wizardcoder: Empowering code large language models with evol-instruct}.
\newblock \emph{Preprint}, arXiv:2306.08568.

\bibitem[{Lv et~al.(2023)Lv, Ding, Qin, Liu, and Sun}]{lv-etal-2023-parameter}
Xingtai Lv, Ning Ding, Yujia Qin, Zhiyuan Liu, and Maosong Sun. 2023.
\newblock \href {https://doi.org/10.18653/v1/2023.acl-short.24} {Parameter-efficient weight ensembling facilitates task-level knowledge transfer}.
\newblock In \emph{Proceedings of the 61st Annual Meeting of the Association for Computational Linguistics (Volume 2: Short Papers)}, pages 270--282, Toronto, Canada. Association for Computational Linguistics.

\bibitem[{{Manuel Romero}(2023)}]{manuel-romero-2023}
{Manuel Romero}. 2023.
\newblock \href {https://doi.org/10.57967/hf/0931} {llama-2-coder-7b (revision d30d193)}.

\bibitem[{Mathur(2023{\natexlab{a}})}]{orca-mini-v2-7b}
Pankaj Mathur. 2023{\natexlab{a}}.
\newblock \href {https://huggingface.co/psmathur/orca\_mini\_v2\_7b} {orca\_mini\_v2\_7b: An explain tuned llama-7b model on uncensored wizardlm, alpaca, \& dolly datasets}.

\bibitem[{Mathur(2023{\natexlab{b}})}]{orca-mini-v3-13b}
Pankaj Mathur. 2023{\natexlab{b}}.
\newblock \href {https://huggingface.co/psmathur/orca\_mini\_v3\_13b} {orca\_mini\_v3\_13b: An orca style llama2-70b model}.

\bibitem[{Mathur(2023{\natexlab{c}})}]{orca-mini-v3-7b}
Pankaj Mathur. 2023{\natexlab{c}}.
\newblock \href {https://huggingface.co/psmathur/orca\_mini\_v3\_7b} {orca\_mini\_v3\_7b: An explain tuned llama2-7b model}.

\bibitem[{Melamed et~al.(2024)Melamed, McCabe, Wakhare, Kim, Huang, and Boix-Adser{\`a}}]{melamed-etal-2024-prompts}
Rimon Melamed, Lucas~Hurley McCabe, Tanay Wakhare, Yejin Kim, H.~Howie Huang, and Enric Boix-Adser{\`a}. 2024.
\newblock \href {https://doi.org/10.18653/v1/2024.emnlp-main.4} {Prompts have evil twins}.
\newblock In \emph{Proceedings of the 2024 Conference on Empirical Methods in Natural Language Processing}, pages 46--74, Miami, Florida, USA. Association for Computational Linguistics.

\bibitem[{Meyn and Tweedie(2009)}]{Meyn_Tweedie_Glynn_2009}
Sean Meyn and Richard~L. Tweedie. 2009.
\newblock \emph{Markov Chains and Stochastic Stability}, 2nd edition.
\newblock Cambridge Mathematical Library. Cambridge University Press.

\bibitem[{Mihaylov et~al.(2018)Mihaylov, Clark, Khot, and Sabharwal}]{arxiv:1809.02789}
Todor Mihaylov, Peter Clark, Tushar Khot, and Ashish Sabharwal. 2018.
\newblock \href {https://arxiv.org/abs/1809.02789} {Can a suit of armor conduct electricity? a new dataset for open book question answering}.
\newblock \emph{Preprint}, arXiv:1809.02789.

\bibitem[{Ming(2023)}]{textgen}
Xu~Ming. 2023.
\newblock \href {https://github.com/shibing624/textgen} {textgen: Implementation of language model finetune}.

\bibitem[{Mitra et~al.(2023)Mitra, Corro, Mahajan, Codas, Simoes, Agarwal, Chen, Razdaibiedina, Jones, Aggarwal, Palangi, Zheng, Rosset, Khanpour, and Awadallah}]{arxiv:2311.11045}
Arindam Mitra, Luciano~Del Corro, Shweti Mahajan, Andres Codas, Clarisse Simoes, Sahaj Agarwal, Xuxi Chen, Anastasia Razdaibiedina, Erik Jones, Kriti Aggarwal, Hamid Palangi, Guoqing Zheng, Corby Rosset, Hamed Khanpour, and Ahmed Awadallah. 2023.
\newblock \href {https://arxiv.org/abs/2311.11045} {Orca 2: Teaching small language models how to reason}.
\newblock \emph{Preprint}, arXiv:2311.11045.

\bibitem[{Mitsuda et~al.(2024)Mitsuda, Chen, Wakatsuki, and Sawada}]{rinna-llama-3-youko-8b}
Koh Mitsuda, Xinqi Chen, Toshiaki Wakatsuki, and Kei Sawada. 2024.
\newblock \href {https://huggingface.co/rinna/llama-3-youko-8b} {rinna/llama-3-youko-8b}.

\bibitem[{{MosaicML NLP Team}(2023{\natexlab{a}})}]{Introducing-MPT-30B:Raising-the-bar-for-open-source-foundation-models}
{MosaicML NLP Team}. 2023{\natexlab{a}}.
\newblock \href {https://www.databricks.com/blog/mpt-30b} {Introducing mpt-30b: Raising the bar for open-source foundation models}.
\newblock Accessed: 2023-06-22.

\bibitem[{{MosaicML NLP Team}(2023{\natexlab{b}})}]{MosaicML2023Introducing}
{MosaicML NLP Team}. 2023{\natexlab{b}}.
\newblock \href {https://www.databricks.com/blog/mpt-7b} {Introducing mpt-7b: A new standard for open-source, commercially usable llms}.
\newblock Accessed: 2023-03-28.

\bibitem[{Muennighoff et~al.(2025)Muennighoff, Su, Wang, Yang, Wei, Yu, Singh, and Kiela}]{arxiv:2402.09906}
Niklas Muennighoff, Hongjin Su, Liang Wang, Nan Yang, Furu Wei, Tao Yu, Amanpreet Singh, and Douwe Kiela. 2025.
\newblock \href {https://arxiv.org/abs/2402.09906} {Generative representational instruction tuning}.
\newblock \emph{Preprint}, arXiv:2402.09906.

\bibitem[{Muennighoff et~al.(2023)Muennighoff, Wang, Sutawika, Roberts, Biderman, Scao, Bari, Shen, Yong, Schoelkopf, Tang, Radev, Aji, Almubarak, Albanie, Alyafeai, Webson, Raff, and Raffel}]{arxiv:2211.01786}
Niklas Muennighoff, Thomas Wang, Lintang Sutawika, Adam Roberts, Stella Biderman, Teven~Le Scao, M~Saiful Bari, Sheng Shen, Zheng-Xin Yong, Hailey Schoelkopf, Xiangru Tang, Dragomir Radev, Alham~Fikri Aji, Khalid Almubarak, Samuel Albanie, Zaid Alyafeai, Albert Webson, Edward Raff, and Colin Raffel. 2023.
\newblock \href {https://arxiv.org/abs/2211.01786} {Crosslingual generalization through multitask finetuning}.
\newblock \emph{Preprint}, arXiv:2211.01786.

\bibitem[{Mukherjee et~al.(2023)Mukherjee, Mitra, Jawahar, Agarwal, Palangi, and Awadallah}]{arxiv:2306.02707}
Subhabrata Mukherjee, Arindam Mitra, Ganesh Jawahar, Sahaj Agarwal, Hamid Palangi, and Ahmed Awadallah. 2023.
\newblock \href {https://arxiv.org/abs/2306.02707} {Orca: Progressive learning from complex explanation traces of gpt-4}.
\newblock \emph{Preprint}, arXiv:2306.02707.

\bibitem[{Murias(2023)}]{unacybertron7b}
Xavier Murias. 2023.
\newblock \href {https://huggingface.co/fblgit/una-cybertron-7b-v2-bf16} {Cybertron: Uniform neural alignment}.

\bibitem[{{Nexusflow.ai team}(2023)}]{nexusraven}
{Nexusflow.ai team}. 2023.
\newblock \href {https://nexusflow.ai/blogs/ravenv2} {Nexusraven-v2: Surpassing gpt-4 for zero-shot function calling}.

\bibitem[{Nguyen et~al.(2024)Nguyen, Zhang, Li, Aljunied, Hu, Shen, Chia, Li, Wang, Tan, Cheng, Chen, Deng, Yang, Liu, Zhang, and Bing}]{arxiv:2312.00738}
Xuan-Phi Nguyen, Wenxuan Zhang, Xin Li, Mahani Aljunied, Zhiqiang Hu, Chenhui Shen, Yew~Ken Chia, Xingxuan Li, Jianyu Wang, Qingyu Tan, Liying Cheng, Guanzheng Chen, Yue Deng, Sen Yang, Chaoqun Liu, Hang Zhang, and Lidong Bing. 2024.
\newblock \href {https://arxiv.org/abs/2312.00738} {Seallms -- large language models for southeast asia}.
\newblock \emph{Preprint}, arXiv:2312.00738.

\bibitem[{Nijkamp et~al.(2023)Nijkamp, Pang, Hayashi, Tu, Wang, Zhou, Savarese, and Xiong}]{arxiv:2203.13474}
Erik Nijkamp, Bo~Pang, Hiroaki Hayashi, Lifu Tu, Huan Wang, Yingbo Zhou, Silvio Savarese, and Caiming Xiong. 2023.
\newblock \href {https://arxiv.org/abs/2203.13474} {Codegen: An open large language model for code with multi-turn program synthesis}.
\newblock \emph{Preprint}, arXiv:2203.13474.

\bibitem[{Nori et~al.(2023)Nori, King, McKinney, Carignan, and Horvitz}]{arxiv:2303.13375}
Harsha Nori, Nicholas King, Scott~Mayer McKinney, Dean Carignan, and Eric Horvitz. 2023.
\newblock \href {https://arxiv.org/abs/2303.13375} {Capabilities of gpt-4 on medical challenge problems}.
\newblock \emph{Preprint}, arXiv:2303.13375.

\bibitem[{Ociepa et~al.(2024{\natexlab{a}})Ociepa, Flis, Wróbel, Gwoździej, {SpeakLeash Team}, and {Cyfronet Team}}]{Introducing-Bielik-7B-v0.1:Polish-Language-Model}
Krzysztof Ociepa, Łukasz Flis, Krzysztof Wróbel, Adrian Gwoździej, {SpeakLeash Team}, and {Cyfronet Team}. 2024{\natexlab{a}}.
\newblock \href {https://huggingface.co/speakleash/Bielik-7B-v0.1} {Introducing bielik-7b-v0.1: Polish language model}.
\newblock Accessed: 2024-04-01.

\bibitem[{Ociepa et~al.(2024{\natexlab{b}})Ociepa, Flis, Wróbel, Kondracki, {SpeakLeash Team}, and {Cyfronet Team}}]{Bielik7Bv01}
Krzysztof Ociepa, Łukasz Flis, Krzysztof Wróbel, Sebastian Kondracki, {SpeakLeash Team}, and {Cyfronet Team}. 2024{\natexlab{b}}.
\newblock \href {https://huggingface.co/speakleash/Bielik-7B-Instruct-v0.1} {Introducing bielik-7b-instruct-v0.1: Instruct polish language model}.
\newblock Accessed: 2024-04-01.

\bibitem[{Ociepa et~al.(2024{\natexlab{c}})Ociepa, Łukasz Flis, Wróbel, Gwoździej, and Kinas}]{arxiv:2410.18565}
Krzysztof Ociepa, Łukasz Flis, Krzysztof Wróbel, Adrian Gwoździej, and Remigiusz Kinas. 2024{\natexlab{c}}.
\newblock \href {https://arxiv.org/abs/2410.18565} {Bielik 7b v0.1: A polish language model -- development, insights, and evaluation}.
\newblock \emph{Preprint}, arXiv:2410.18565.

\bibitem[{{OpenAI} et~al.(2024){OpenAI}, Achiam, Adler, Agarwal, Ahmad, Akkaya, Aleman, Almeida, Altenschmidt, Altman, Anadkat, Avila, Babuschkin, Balaji, Balcom, Baltescu, Bao, Bavarian, Belgum, Bello, Berdine, Bernadett-Shapiro, Berner, Bogdonoff, Boiko, Boyd, Brakman, Brockman, Brooks, Brundage, Button, Cai, Campbell, Cann, Carey, Carlson, Carmichael, Chan, Chang, Chantzis, Chen, Chen, Chen, Chen, Chen, Chess, Cho, Chu, Chung, Cummings, Currier, Dai, Decareaux, Degry, Deutsch, Deville, Dhar, Dohan, Dowling, Dunning, Ecoffet, Eleti, Eloundou, Farhi, Fedus, Felix, Fishman, Forte, Fulford, Gao, Georges, Gibson, Goel, Gogineni, Goh, Gontijo-Lopes, Gordon, Grafstein, Gray, Greene, Gross, Gu, Guo, Hallacy, Han, Harris, He, Heaton, Heidecke, Hesse, Hickey, Hickey, Hoeschele, Houghton, Hsu, Hu, Hu, Huizinga, Jain, Jain, Jang, Jiang, Jiang, Jin, Jin, Jomoto, Jonn, Jun, Kaftan, Łukasz Kaiser, Kamali, Kanitscheider, Keskar, Khan, Kilpatrick, Kim, Kim, Kim, Kirchner, Kiros, Knight, Kokotajlo, Łukasz Kondraciuk,
  Kondrich, Konstantinidis, Kosic, Krueger, Kuo, Lampe, Lan, Lee, Leike, Leung, Levy, Li, Lim, Lin, Lin, Litwin, Lopez, Lowe, Lue, Makanju, Malfacini, Manning, Markov, Markovski, Martin, Mayer, Mayne, McGrew, McKinney, McLeavey, McMillan, McNeil, Medina, Mehta, Menick, Metz, Mishchenko, Mishkin, Monaco, Morikawa, Mossing, Mu, Murati, Murk, Mély, Nair, Nakano, Nayak, Neelakantan, Ngo, Noh, Ouyang, O'Keefe, Pachocki, Paino, Palermo, Pantuliano, Parascandolo, Parish, Parparita, Passos, Pavlov, Peng, Perelman, de~Avila Belbute~Peres, Petrov, de~Oliveira~Pinto, Michael, Pokorny, Pokrass, Pong, Powell, Power, Power, Proehl, Puri, Radford, Rae, Ramesh, Raymond, Real, Rimbach, Ross, Rotsted, Roussez, Ryder, Saltarelli, Sanders, Santurkar, Sastry, Schmidt, Schnurr, Schulman, Selsam, Sheppard, Sherbakov, Shieh, Shoker, Shyam, Sidor, Sigler, Simens, Sitkin, Slama, Sohl, Sokolowsky, Song, Staudacher, Such, Summers, Sutskever, Tang, Tezak, Thompson, Tillet, Tootoonchian, Tseng, Tuggle, Turley, Tworek, Uribe, Vallone,
  Vijayvergiya, Voss, Wainwright, Wang, Wang, Wang, Ward, Wei, Weinmann, Welihinda, Welinder, Weng, Weng, Wiethoff, Willner, Winter, Wolrich, Wong, Workman, Wu, Wu, Wu, Xiao, Xu, Yoo, Yu, Yuan, Zaremba, Zellers, Zhang, Zhang, Zhao, Zheng, Zhuang, Zhuk, and Zoph}]{arxiv:2303.08774}
{OpenAI}, Josh Achiam, Steven Adler, Sandhini Agarwal, Lama Ahmad, Ilge Akkaya, Florencia~Leoni Aleman, Diogo Almeida, Janko Altenschmidt, Sam Altman, Shyamal Anadkat, Red Avila, Igor Babuschkin, Suchir Balaji, Valerie Balcom, Paul Baltescu, Haiming Bao, Mohammad Bavarian, Jeff Belgum, and 262 others. 2024.
\newblock \href {https://arxiv.org/abs/2303.08774} {Gpt-4 technical report}.
\newblock \emph{Preprint}, arXiv:2303.08774.

\bibitem[{Oyama et~al.(2023)Oyama, Yokoi, and Shimodaira}]{oyama-etal-2023-norm}
Momose Oyama, Sho Yokoi, and Hidetoshi Shimodaira. 2023.
\newblock \href {https://doi.org/10.18653/v1/2023.emnlp-main.131} {Norm of word embedding encodes information gain}.
\newblock In \emph{Proceedings of the 2023 Conference on Empirical Methods in Natural Language Processing}, pages 2108--2130, Singapore. Association for Computational Linguistics.

\bibitem[{Pal and Sankarasubbu(2024{\natexlab{a}})}]{arxiv:2402.07023}
Ankit Pal and Malaikannan Sankarasubbu. 2024{\natexlab{a}}.
\newblock \href {https://arxiv.org/abs/2402.07023} {Gemini goes to med school: Exploring the capabilities of multimodal large language models on medical challenge problems \& hallucinations}.
\newblock \emph{Preprint}, arXiv:2402.07023.

\bibitem[{Pal and Sankarasubbu(2024{\natexlab{b}})}]{OpenBioLLMs}
Ankit Pal and Malaikannan Sankarasubbu. 2024{\natexlab{b}}.
\newblock \href {https://huggingface.co/aaditya/OpenBioLLM-Llama3-70B} {Openbiollms: Advancing open-source large language models for healthcare and life sciences}.

\bibitem[{Pal et~al.(2024)Pal, Karkhanis, Dooley, Roberts, Naidu, and White}]{arxiv:2402.13228}
Arka Pal, Deep Karkhanis, Samuel Dooley, Manley Roberts, Siddartha Naidu, and Colin White. 2024.
\newblock \href {https://arxiv.org/abs/2402.13228} {Smaug: Fixing failure modes of preference optimisation with dpo-positive}.
\newblock \emph{Preprint}, arXiv:2402.13228.

\bibitem[{Parrish et~al.(2022)Parrish, Chen, Nangia, Padmakumar, Phang, Thompson, Htut, and Bowman}]{arxiv:2110.08193}
Alicia Parrish, Angelica Chen, Nikita Nangia, Vishakh Padmakumar, Jason Phang, Jana Thompson, Phu~Mon Htut, and Samuel~R. Bowman. 2022.
\newblock \href {https://arxiv.org/abs/2110.08193} {Bbq: A hand-built bias benchmark for question answering}.
\newblock \emph{Preprint}, arXiv:2110.08193.

\bibitem[{Pekelis et~al.(2024)Pekelis, Feil, Moret, Huang, and Peng}]{gradientlongcontextllama3}
Leonid Pekelis, Michael Feil, Forrest Moret, Mark Huang, and Tiffany Peng. 2024.
\newblock \href {https://doi.org/10.57967/hf/3372} {Llama 3 gradient: A series of long context models}.

\bibitem[{Penedo et~al.(2023)Penedo, Malartic, Hesslow, Cojocaru, Cappelli, Alobeidli, Pannier, Almazrouei, and Launay}]{arxiv:2306.01116}
Guilherme Penedo, Quentin Malartic, Daniel Hesslow, Ruxandra Cojocaru, Alessandro Cappelli, Hamza Alobeidli, Baptiste Pannier, Ebtesam Almazrouei, and Julien Launay. 2023.
\newblock \href {https://arxiv.org/abs/2306.01116} {The refinedweb dataset for falcon llm: Outperforming curated corpora with web data, and web data only}.
\newblock \emph{Preprint}, arXiv:2306.01116.

\bibitem[{Peng et~al.(2023{\natexlab{a}})Peng, Li, He, Galley, and Gao}]{arxiv:2304.03277}
Baolin Peng, Chunyuan Li, Pengcheng He, Michel Galley, and Jianfeng Gao. 2023{\natexlab{a}}.
\newblock \href {https://arxiv.org/abs/2304.03277} {Instruction tuning with gpt-4}.
\newblock \emph{Preprint}, arXiv:2304.03277.

\bibitem[{Peng et~al.(2023{\natexlab{b}})Peng, Quesnelle, Fan, and Shippole}]{arxiv:2309.00071}
Bowen Peng, Jeffrey Quesnelle, Honglu Fan, and Enrico Shippole. 2023{\natexlab{b}}.
\newblock \href {https://arxiv.org/abs/2309.00071} {Yarn: Efficient context window extension of large language models}.
\newblock \emph{Preprint}, arXiv:2309.00071.

\bibitem[{Pinnaparaju et~al.(2024)Pinnaparaju, Adithyan, Phung, Tow, Baicoianu, and Cooper}]{stable-code-3b}
Nikhil Pinnaparaju, Reshinth Adithyan, Duy Phung, Jonathan Tow, James Baicoianu, and Nathan Cooper. 2024.
\newblock \href {https://huggingface.co/stabilityai/stable-code-3b} {Stable code 3b}.

\bibitem[{Pipatanakul et~al.(2023)Pipatanakul, Jirabovonvisut, Manakul, Sripaisarnmongkol, Patomwong, Chokchainant, and Tharnpipitchai}]{arxiv:2312.13951}
Kunat Pipatanakul, Phatrasek Jirabovonvisut, Potsawee Manakul, Sittipong Sripaisarnmongkol, Ruangsak Patomwong, Pathomporn Chokchainant, and Kasima Tharnpipitchai. 2023.
\newblock \href {https://arxiv.org/abs/2312.13951} {Typhoon: Thai large language models}.
\newblock \emph{Preprint}, arXiv:2312.13951.

\bibitem[{Polignano et~al.(2024)Polignano, Basile, and Semeraro}]{arxiv:2405.07101}
Marco Polignano, Pierpaolo Basile, and Giovanni Semeraro. 2024.
\newblock \href {https://arxiv.org/abs/2405.07101} {Advanced natural-based interaction for the italian language: Llamantino-3-anita}.
\newblock \emph{Preprint}, arXiv:2405.07101.

\bibitem[{Pradeep et~al.(2023)Pradeep, Sharifymoghaddam, and Lin}]{arxiv:2309.15088}
Ronak Pradeep, Sahel Sharifymoghaddam, and Jimmy Lin. 2023.
\newblock \href {https://arxiv.org/abs/2309.15088} {Rankvicuna: Zero-shot listwise document reranking with open-source large language models}.
\newblock \emph{Preprint}, arXiv:2309.15088.

\bibitem[{Press et~al.(2022)Press, Smith, and Lewis}]{arxiv:2108.12409}
Ofir Press, Noah~A. Smith, and Mike Lewis. 2022.
\newblock \href {https://arxiv.org/abs/2108.12409} {Train short, test long: Attention with linear biases enables input length extrapolation}.
\newblock \emph{Preprint}, arXiv:2108.12409.

\bibitem[{Rafailov et~al.(2024)Rafailov, Sharma, Mitchell, Ermon, Manning, and Finn}]{arxiv:2305.18290}
Rafael Rafailov, Archit Sharma, Eric Mitchell, Stefano Ermon, Christopher~D. Manning, and Chelsea Finn. 2024.
\newblock \href {https://arxiv.org/abs/2305.18290} {Direct preference optimization: Your language model is secretly a reward model}.
\newblock \emph{Preprint}, arXiv:2305.18290.

\bibitem[{Rajbhandari et~al.(2020)Rajbhandari, Rasley, Ruwase, and He}]{arxiv:1910.02054}
Samyam Rajbhandari, Jeff Rasley, Olatunji Ruwase, and Yuxiong He. 2020.
\newblock \href {https://arxiv.org/abs/1910.02054} {Zero: Memory optimizations toward training trillion parameter models}.
\newblock \emph{Preprint}, arXiv:1910.02054.

\bibitem[{Rozière et~al.(2024)Rozière, Gehring, Gloeckle, Sootla, Gat, Tan, Adi, Liu, Sauvestre, Remez, Rapin, Kozhevnikov, Evtimov, Bitton, Bhatt, Ferrer, Grattafiori, Xiong, Défossez, Copet, Azhar, Touvron, Martin, Usunier, Scialom, and Synnaeve}]{arxiv:2308.12950}
Baptiste Rozière, Jonas Gehring, Fabian Gloeckle, Sten Sootla, Itai Gat, Xiaoqing~Ellen Tan, Yossi Adi, Jingyu Liu, Romain Sauvestre, Tal Remez, Jérémy Rapin, Artyom Kozhevnikov, Ivan Evtimov, Joanna Bitton, Manish Bhatt, Cristian~Canton Ferrer, Aaron Grattafiori, Wenhan Xiong, Alexandre Défossez, and 7 others. 2024.
\newblock \href {https://arxiv.org/abs/2308.12950} {Code llama: Open foundation models for code}.
\newblock \emph{Preprint}, arXiv:2308.12950.

\bibitem[{Rudinger et~al.(2018)Rudinger, Naradowsky, Leonard, and Durme}]{arxiv:1804.09301}
Rachel Rudinger, Jason Naradowsky, Brian Leonard, and Benjamin~Van Durme. 2018.
\newblock \href {https://arxiv.org/abs/1804.09301} {Gender bias in coreference resolution}.
\newblock \emph{Preprint}, arXiv:1804.09301.

\bibitem[{Sainz et~al.(2024)Sainz, García-Ferrero, Agerri, de~Lacalle, Rigau, and Agirre}]{arxiv:2310.03668}
Oscar Sainz, Iker García-Ferrero, Rodrigo Agerri, Oier~Lopez de~Lacalle, German Rigau, and Eneko Agirre. 2024.
\newblock \href {https://arxiv.org/abs/2310.03668} {Gollie: Annotation guidelines improve zero-shot information-extraction}.
\newblock \emph{Preprint}, arXiv:2310.03668.

\bibitem[{Sakaguchi et~al.(2019)Sakaguchi, Bras, Bhagavatula, and Choi}]{arxiv:1907.10641}
Keisuke Sakaguchi, Ronan~Le Bras, Chandra Bhagavatula, and Yejin Choi. 2019.
\newblock \href {https://arxiv.org/abs/1907.10641} {Winogrande: An adversarial winograd schema challenge at scale}.
\newblock \emph{Preprint}, arXiv:1907.10641.

\bibitem[{Sap et~al.(2019)Sap, Rashkin, Chen, LeBras, and Choi}]{arxiv:1904.09728}
Maarten Sap, Hannah Rashkin, Derek Chen, Ronan LeBras, and Yejin Choi. 2019.
\newblock \href {https://arxiv.org/abs/1904.09728} {Socialiqa: Commonsense reasoning about social interactions}.
\newblock \emph{Preprint}, arXiv:1904.09728.

\bibitem[{Sasaki et~al.(2023{\natexlab{a}})Sasaki, Hirakawa, Horie, and Nakamura}]{ELYZA-japanese-Llama-2-7b}
Akira Sasaki, Masato Hirakawa, Shintaro Horie, and Tomoaki Nakamura. 2023{\natexlab{a}}.
\newblock \href {https://huggingface.co/elyza/ELYZA-japanese-Llama-2-7b} {Elyza-japanese-llama-2-7b}.

\bibitem[{Sasaki et~al.(2023{\natexlab{b}})Sasaki, Hirakawa, Horie, Nakamura, Passaglia, and Oba}]{elyzallama2023}
Akira Sasaki, Masato Hirakawa, Shintaro Horie, Tomoaki Nakamura, Sam Passaglia, and Daisuke Oba. 2023{\natexlab{b}}.
\newblock \href {https://huggingface.co/elyza/ELYZA-japanese-Llama-2-13b} {Elyza-japanese-llama-2-13b}.

\bibitem[{Sato(2022)}]{DBLP:conf/naacl/Sato22}
Ryoma Sato. 2022.
\newblock \href {https://doi.org/10.18653/V1/2022.NAACL-MAIN.157} {Word tour: One-dimensional word embeddings via the traveling salesman problem}.
\newblock In \emph{Proceedings of the 2022 Conference of the North American Chapter of the Association for Computational Linguistics: Human Language Technologies, {NAACL} 2022, Seattle, WA, United States, July 10-15, 2022}, pages 2166--2172. Association for Computational Linguistics.

\bibitem[{Sawada et~al.(2024)Sawada, Zhao, Shing, Mitsui, Kaga, Hono, Wakatsuki, and Mitsuda}]{arxiv:2404.01657}
Kei Sawada, Tianyu Zhao, Makoto Shing, Kentaro Mitsui, Akio Kaga, Yukiya Hono, Toshiaki Wakatsuki, and Koh Mitsuda. 2024.
\newblock \href {https://arxiv.org/abs/2404.01657} {Release of pre-trained models for the japanese language}.
\newblock \emph{Preprint}, arXiv:2404.01657.

\bibitem[{Shao et~al.(2024)Shao, Wang, Zhu, Xu, Song, Bi, Zhang, Zhang, Li, Wu, and Guo}]{arxiv:2402.03300}
Zhihong Shao, Peiyi Wang, Qihao Zhu, Runxin Xu, Junxiao Song, Xiao Bi, Haowei Zhang, Mingchuan Zhang, Y.~K. Li, Y.~Wu, and Daya Guo. 2024.
\newblock \href {https://arxiv.org/abs/2402.03300} {Deepseekmath: Pushing the limits of mathematical reasoning in open language models}.
\newblock \emph{Preprint}, arXiv:2402.03300.

\bibitem[{Sharma et~al.(2023)Sharma, Ash, and Misra}]{arxiv:2312.13558}
Pratyusha Sharma, Jordan~T. Ash, and Dipendra Misra. 2023.
\newblock \href {https://arxiv.org/abs/2312.13558} {The truth is in there: Improving reasoning in language models with layer-selective rank reduction}.
\newblock \emph{Preprint}, arXiv:2312.13558.

\bibitem[{Shazeer(2019)}]{arxiv:1911.02150}
Noam Shazeer. 2019.
\newblock \href {https://arxiv.org/abs/1911.02150} {Fast transformer decoding: One write-head is all you need}.
\newblock \emph{Preprint}, arXiv:1911.02150.

\bibitem[{Shazeer(2020)}]{arxiv:2002.05202}
Noam Shazeer. 2020.
\newblock \href {https://arxiv.org/abs/2002.05202} {Glu variants improve transformer}.
\newblock \emph{Preprint}, arXiv:2002.05202.

\bibitem[{Shi et~al.(2024)Shi, Min, Lomeli, Zhou, Li, Szilvasy, James, Lin, Smith, Zettlemoyer, Yih, and Lewis}]{arxiv:2310.10638}
Weijia Shi, Sewon Min, Maria Lomeli, Chunting Zhou, Margaret Li, Gergely Szilvasy, Rich James, Xi~Victoria Lin, Noah~A. Smith, Luke Zettlemoyer, Scott Yih, and Mike Lewis. 2024.
\newblock \href {https://arxiv.org/abs/2310.10638} {In-context pretraining: Language modeling beyond document boundaries}.
\newblock \emph{Preprint}, arXiv:2310.10638.

\bibitem[{Shimodaira(1993)}]{Shimodaira1993modelmap}
Hidetoshi Shimodaira. 1993.
\newblock \href {https://ismrepo.ism.ac.jp/records/31997} {A model search technique based on confidence set and map of models [モデルの信頼集合と地図によるモデル探索] (in japanese)}.
\newblock \emph{Proceedings of the Institute of Statistical Mathematics}, 41(2):131--147.

\bibitem[{Shimodaira(2001)}]{shimodaira2001multiple}
Hidetoshi Shimodaira. 2001.
\newblock Multiple comparisons of log-likelihoods and combining nonnested models with applications to phylogenetic tree selection.
\newblock \emph{Communications in Statistics-Theory and Methods}, 30(8-9):1751--1772.

\bibitem[{Shimodaira and Cao(1998)}]{shimodaira1998graphical}
Hidetoshi Shimodaira and Ying Cao. 1998.
\newblock A graphical technique for model selection diagnosis.
\newblock \emph{The Institute of Statistical Mathematics Research Memorandum}, 680.

\bibitem[{Shimodaira and Hasegawa(2005)}]{Shimodaira2005}
Hidetoshi Shimodaira and Masami Hasegawa. 2005.
\newblock \href {https://doi.org/10.1007/0-387-27733-1_17} {\emph{Assessing the Uncertainty in Phylogenetic Inference}}, pages 463--493.
\newblock Springer New York, New York, NY.

\bibitem[{Shimodaira and Terada(2019)}]{shimodaira2019selective}
Hidetoshi Shimodaira and Yoshikazu Terada. 2019.
\newblock \href {https://doi.org/10.3389/fevo.2019.00174} {Selective inference for testing trees and edges in phylogenetics}.
\newblock \emph{Frontiers in ecology and evolution}, 7:174.

\bibitem[{Shoeybi et~al.(2020)Shoeybi, Patwary, Puri, LeGresley, Casper, and Catanzaro}]{arxiv:1909.08053}
Mohammad Shoeybi, Mostofa Patwary, Raul Puri, Patrick LeGresley, Jared Casper, and Bryan Catanzaro. 2020.
\newblock \href {https://arxiv.org/abs/1909.08053} {Megatron-lm: Training multi-billion parameter language models using model parallelism}.
\newblock \emph{Preprint}, arXiv:1909.08053.

\bibitem[{Singhal et~al.(2022)Singhal, Azizi, Tu, Mahdavi, Wei, Chung, Scales, Tanwani, Cole-Lewis, Pfohl, Payne, Seneviratne, Gamble, Kelly, Scharli, Chowdhery, Mansfield, y~Arcas, Webster, Corrado, Matias, Chou, Gottweis, Tomasev, Liu, Rajkomar, Barral, Semturs, Karthikesalingam, and Natarajan}]{arxiv:2212.13138}
Karan Singhal, Shekoofeh Azizi, Tao Tu, S.~Sara Mahdavi, Jason Wei, Hyung~Won Chung, Nathan Scales, Ajay Tanwani, Heather Cole-Lewis, Stephen Pfohl, Perry Payne, Martin Seneviratne, Paul Gamble, Chris Kelly, Nathaneal Scharli, Aakanksha Chowdhery, Philip Mansfield, Blaise~Aguera y~Arcas, Dale Webster, and 11 others. 2022.
\newblock \href {https://arxiv.org/abs/2212.13138} {Large language models encode clinical knowledge}.
\newblock \emph{Preprint}, arXiv:2212.13138.

\bibitem[{Singhal et~al.(2023)Singhal, Tu, Gottweis, Sayres, Wulczyn, Hou, Clark, Pfohl, Cole-Lewis, Neal, Schaekermann, Wang, Amin, Lachgar, Mansfield, Prakash, Green, Dominowska, y~Arcas, Tomasev, Liu, Wong, Semturs, Mahdavi, Barral, Webster, Corrado, Matias, Azizi, Karthikesalingam, and Natarajan}]{arxiv:2305.09617}
Karan Singhal, Tao Tu, Juraj Gottweis, Rory Sayres, Ellery Wulczyn, Le~Hou, Kevin Clark, Stephen Pfohl, Heather Cole-Lewis, Darlene Neal, Mike Schaekermann, Amy Wang, Mohamed Amin, Sami Lachgar, Philip Mansfield, Sushant Prakash, Bradley Green, Ewa Dominowska, Blaise~Aguera y~Arcas, and 12 others. 2023.
\newblock \href {https://arxiv.org/abs/2305.09617} {Towards expert-level medical question answering with large language models}.
\newblock \emph{Preprint}, arXiv:2305.09617.

\bibitem[{Sonkar et~al.(2023)Sonkar, Liu, Mallick, and Baraniuk}]{arxiv:2305.13272}
Shashank Sonkar, Naiming Liu, Debshila~Basu Mallick, and Richard~G. Baraniuk. 2023.
\newblock \href {https://arxiv.org/abs/2305.13272} {Class: A design framework for building intelligent tutoring systems based on learning science principles}.
\newblock \emph{Preprint}, arXiv:2305.13272.

\bibitem[{Srivastava et~al.(2023)Srivastava, Rastogi, Rao, Shoeb, Abid, Fisch, Brown, Santoro, Gupta, Garriga-Alonso, Kluska, Lewkowycz, Agarwal, Power, Ray, Warstadt, Kocurek, Safaya, Tazarv, Xiang, Parrish, Nie, Hussain, Askell, Dsouza, Slone, Rahane, Iyer, Andreassen, Madotto, Santilli, Stuhlmüller, Dai, La, Lampinen, Zou, Jiang, Chen, Vuong, Gupta, Gottardi, Norelli, Venkatesh, Gholamidavoodi, Tabassum, Menezes, Kirubarajan, Mullokandov, Sabharwal, Herrick, Efrat, Erdem, Karakaş, Roberts, Loe, Zoph, Bojanowski, Özyurt, Hedayatnia, Neyshabur, Inden, Stein, Ekmekci, Lin, Howald, Orinion, Diao, Dour, Stinson, Argueta, Ramírez, Singh, Rathkopf, Meng, Baral, Wu, Callison-Burch, Waites, Voigt, Manning, Potts, Ramirez, Rivera, Siro, Raffel, Ashcraft, Garbacea, Sileo, Garrette, Hendrycks, Kilman, Roth, Freeman, Khashabi, Levy, González, Perszyk, Hernandez, Chen, Ippolito, Gilboa, Dohan, Drakard, Jurgens, Datta, Ganguli, Emelin, Kleyko, Yuret, Chen, Tam, Hupkes, Misra, Buzan, Mollo, Yang, Lee, Schrader,
  Shutova, Cubuk, Segal, Hagerman, Barnes, Donoway, Pavlick, Rodola, Lam, Chu, Tang, Erdem, Chang, Chi, Dyer, Jerzak, Kim, Manyasi, Zheltonozhskii, Xia, Siar, Martínez-Plumed, Happé, Chollet, Rong, Mishra, Winata, de~Melo, Kruszewski, Parascandolo, Mariani, Wang, Jaimovitch-López, Betz, Gur-Ari, Galijasevic, Kim, Rashkin, Hajishirzi, Mehta, Bogar, Shevlin, Schütze, Yakura, Zhang, Wong, Ng, Noble, Jumelet, Geissinger, Kernion, Hilton, Lee, Fisac, Simon, Koppel, Zheng, Zou, Kocoń, Thompson, Wingfield, Kaplan, Radom, Sohl-Dickstein, Phang, Wei, Yosinski, Novikova, Bosscher, Marsh, Kim, Taal, Engel, Alabi, Xu, Song, Tang, Waweru, Burden, Miller, Balis, Batchelder, Berant, Frohberg, Rozen, Hernandez-Orallo, Boudeman, Guerr, Jones, Tenenbaum, Rule, Chua, Kanclerz, Livescu, Krauth, Gopalakrishnan, Ignatyeva, Markert, Dhole, Gimpel, Omondi, Mathewson, Chiafullo, Shkaruta, Shridhar, McDonell, Richardson, Reynolds, Gao, Zhang, Dugan, Qin, Contreras-Ochando, Morency, Moschella, Lam, Noble, Schmidt, He, Colón,
  Metz, Şenel, Bosma, Sap, ter Hoeve, Farooqi, Faruqui, Mazeika, Baturan, Marelli, Maru, Quintana, Tolkiehn, Giulianelli, Lewis, Potthast, Leavitt, Hagen, Schubert, Baitemirova, Arnaud, McElrath, Yee, Cohen, Gu, Ivanitskiy, Starritt, Strube, Swędrowski, Bevilacqua, Yasunaga, Kale, Cain, Xu, Suzgun, Walker, Tiwari, Bansal, Aminnaseri, Geva, Gheini, T, Peng, Chi, Lee, Krakover, Cameron, Roberts, Doiron, Martinez, Nangia, Deckers, Muennighoff, Keskar, Iyer, Constant, Fiedel, Wen, Zhang, Agha, Elbaghdadi, Levy, Evans, Casares, Doshi, Fung, Liang, Vicol, Alipoormolabashi, Liao, Liang, Chang, Eckersley, Htut, Hwang, Miłkowski, Patil, Pezeshkpour, Oli, Mei, Lyu, Chen, Banjade, Rudolph, Gabriel, Habacker, Risco, Millière, Garg, Barnes, Saurous, Arakawa, Raymaekers, Frank, Sikand, Novak, Sitelew, LeBras, Liu, Jacobs, Zhang, Salakhutdinov, Chi, Lee, Stovall, Teehan, Yang, Singh, Mohammad, Anand, Dillavou, Shleifer, Wiseman, Gruetter, Bowman, Schoenholz, Han, Kwatra, Rous, Ghazarian, Ghosh, Casey, Bischoff,
  Gehrmann, Schuster, Sadeghi, Hamdan, Zhou, Srivastava, Shi, Singh, Asaadi, Gu, Pachchigar, Toshniwal, Upadhyay, Shyamolima, Debnath, Shakeri, Thormeyer, Melzi, Reddy, Makini, Lee, Torene, Hatwar, Dehaene, Divic, Ermon, Biderman, Lin, Prasad, Piantadosi, Shieber, Misherghi, Kiritchenko, Mishra, Linzen, Schuster, Li, Yu, Ali, Hashimoto, Wu, Desbordes, Rothschild, Phan, Wang, Nkinyili, Schick, Kornev, Tunduny, Gerstenberg, Chang, Neeraj, Khot, Shultz, Shaham, Misra, Demberg, Nyamai, Raunak, Ramasesh, Prabhu, Padmakumar, Srikumar, Fedus, Saunders, Zhang, Vossen, Ren, Tong, Zhao, Wu, Shen, Yaghoobzadeh, Lakretz, Song, Bahri, Choi, Yang, Hao, Chen, Belinkov, Hou, Hou, Bai, Seid, Zhao, Wang, Wang, Wang, and Wu}]{arxiv:2206.04615}
Aarohi Srivastava, Abhinav Rastogi, Abhishek Rao, Abu Awal~Md Shoeb, Abubakar Abid, Adam Fisch, Adam~R. Brown, Adam Santoro, Aditya Gupta, Adrià Garriga-Alonso, Agnieszka Kluska, Aitor Lewkowycz, Akshat Agarwal, Alethea Power, Alex Ray, Alex Warstadt, Alexander~W. Kocurek, Ali Safaya, Ali Tazarv, and 432 others. 2023.
\newblock \href {https://arxiv.org/abs/2206.04615} {Beyond the imitation game: Quantifying and extrapolating the capabilities of language models}.
\newblock \emph{Preprint}, arXiv:2206.04615.

\bibitem[{{Stability AI Language Team}(2024)}]{StableLM-2-1.6B}
{Stability AI Language Team}. 2024.
\newblock \href {https://huggingface.co/stabilityai/stablelm-2-1.6b} {Stable lm 2 1.6b}.

\bibitem[{Su et~al.(2023{\natexlab{a}})Su, Lu, Pan, Murtadha, Wen, and Liu}]{arxiv:2104.09864}
Jianlin Su, Yu~Lu, Shengfeng Pan, Ahmed Murtadha, Bo~Wen, and Yunfeng Liu. 2023{\natexlab{a}}.
\newblock \href {https://arxiv.org/abs/2104.09864} {Roformer: Enhanced transformer with rotary position embedding}.
\newblock \emph{Preprint}, arXiv:2104.09864.

\bibitem[{Su et~al.(2023{\natexlab{b}})Su, Lan, and Cai}]{openalpaca}
Yixuan Su, Tian Lan, and Deng Cai. 2023{\natexlab{b}}.
\newblock \href {https://github.com/yxuansu/OpenAlpaca} {Openalpaca: A fully open-source instruction-following model based on openllama}.

\bibitem[{Sudalairaj et~al.(2024)Sudalairaj, Bhandwaldar, Pareja, Xu, Cox, and Srivastava}]{arxiv:2403.01081}
Shivchander Sudalairaj, Abhishek Bhandwaldar, Aldo Pareja, Kai Xu, David~D. Cox, and Akash Srivastava. 2024.
\newblock \href {https://arxiv.org/abs/2403.01081} {Lab: Large-scale alignment for chatbots}.
\newblock \emph{Preprint}, arXiv:2403.01081.

\bibitem[{Sun et~al.(2024)Sun, Liu, Bair, and Kolter}]{arxiv:2306.11695}
Mingjie Sun, Zhuang Liu, Anna Bair, and J.~Zico Kolter. 2024.
\newblock \href {https://arxiv.org/abs/2306.11695} {A simple and effective pruning approach for large language models}.
\newblock \emph{Preprint}, arXiv:2306.11695.

\bibitem[{Talmor et~al.(2019)Talmor, Herzig, Lourie, and Berant}]{arxiv:1811.00937}
Alon Talmor, Jonathan Herzig, Nicholas Lourie, and Jonathan Berant. 2019.
\newblock \href {https://arxiv.org/abs/1811.00937} {Commonsenseqa: A question answering challenge targeting commonsense knowledge}.
\newblock \emph{Preprint}, arXiv:1811.00937.

\bibitem[{Taori et~al.(2023)Taori, Gulrajani, Zhang, Dubois, Li, Guestrin, Liang, and Hashimoto}]{alpaca}
Rohan Taori, Ishaan Gulrajani, Tianyi Zhang, Yann Dubois, Xuechen Li, Carlos Guestrin, Percy Liang, and Tatsunori~B. Hashimoto. 2023.
\newblock \href {https://github.com/tatsu-lab/stanford\_alpaca} {Stanford alpaca: An instruction-following llama model}.

\bibitem[{Tay et~al.(2023)Tay, Dehghani, Tran, Garcia, Wei, Wang, Chung, Shakeri, Bahri, Schuster, Zheng, Zhou, Houlsby, and Metzler}]{arxiv:2205.05131}
Yi~Tay, Mostafa Dehghani, Vinh~Q. Tran, Xavier Garcia, Jason Wei, Xuezhi Wang, Hyung~Won Chung, Siamak Shakeri, Dara Bahri, Tal Schuster, Huaixiu~Steven Zheng, Denny Zhou, Neil Houlsby, and Donald Metzler. 2023.
\newblock \href {https://arxiv.org/abs/2205.05131} {Ul2: Unifying language learning paradigms}.
\newblock \emph{Preprint}, arXiv:2205.05131.

\bibitem[{Tay et~al.(2022)Tay, Wei, Chung, Tran, So, Shakeri, Garcia, Zheng, Rao, Chowdhery, Zhou, Metzler, Petrov, Houlsby, Le, and Dehghani}]{arxiv:2210.11399}
Yi~Tay, Jason Wei, Hyung~Won Chung, Vinh~Q. Tran, David~R. So, Siamak Shakeri, Xavier Garcia, Huaixiu~Steven Zheng, Jinfeng Rao, Aakanksha Chowdhery, Denny Zhou, Donald Metzler, Slav Petrov, Neil Houlsby, Quoc~V. Le, and Mostafa Dehghani. 2022.
\newblock \href {https://arxiv.org/abs/2210.11399} {Transcending scaling laws with 0.1\% extra compute}.
\newblock \emph{Preprint}, arXiv:2210.11399.

\bibitem[{"Teknium" et~al.(2024{\natexlab{a}})"Teknium", Goddard, "interstellarninja", "theemozilla", "karan4d", and "huemin\_art"}]{Hermes-2-Theta-Llama-3-8B}
"Teknium", Charles Goddard, "interstellarninja", "theemozilla", "karan4d", and "huemin\_art". 2024{\natexlab{a}}.
\newblock \href {https://huggingface.co/NousResearch/Hermes-2-Theta-Llama-3-8B} {Hermes-2-theta-llama-3-8b}.

\bibitem[{"Teknium" et~al.(2024{\natexlab{b}})"Teknium", "interstellarninja", "theemozilla", "karan4d", and "huemin\_art"}]{Hermes-2-Pro-Llama-3-8B}
"Teknium", "interstellarninja", "theemozilla", "karan4d", and "huemin\_art". 2024{\natexlab{b}}.
\newblock \href {https://huggingface.co/NousResearch/Hermes-2-Pro-Llama-3-8B} {Hermes-2-pro-llama-3-8b}.

\bibitem[{"Teknium" et~al.(2024{\natexlab{c}})"Teknium", "theemozilla", "karan4d", and "huemin\_art"}]{Nous-Hermes-2-Mistral-7B-DPO}
"Teknium", "theemozilla", "karan4d", and "huemin\_art". 2024{\natexlab{c}}.
\newblock \href {https://huggingface.co/NousResearch/Nous-Hermes-2-Mistral-7B-DPO} {Nous hermes 2 mistral 7b dpo}.

\bibitem[{Thrush et~al.(2025)Thrush, Potts, and Hashimoto}]{thrush2025improving}
Tristan Thrush, Christopher Potts, and Tatsunori Hashimoto. 2025.
\newblock \href {https://openreview.net/forum?id=huuKoVQnB0} {Improving pretraining data using perplexity correlations}.
\newblock In \emph{Proceedings of the International Conference on Learning Representations (ICLR)}.

\bibitem[{Tissera(2023)}]{SynthIA-7B-v1.3}
Migel Tissera. 2023.
\newblock \href {https://huggingface.co/migtissera/Synthia-13B} {Synthia-7b-v1.3: Synthetic intelligent agent}.

\bibitem[{{Together Computer}(2023)}]{together2023redpajama}
{Together Computer}. 2023.
\newblock \href {https://github.com/togethercomputer/RedPajama-Data} {Redpajama-data: An open source recipe to reproduce llama training dataset}.

\bibitem[{Touvron et~al.(2023{\natexlab{a}})Touvron, Lavril, Izacard, Martinet, Lachaux, Lacroix, Rozière, Goyal, Hambro, Azhar, Rodriguez, Joulin, Grave, and Lample}]{arxiv:2302.13971}
Hugo Touvron, Thibaut Lavril, Gautier Izacard, Xavier Martinet, Marie-Anne Lachaux, Timothée Lacroix, Baptiste Rozière, Naman Goyal, Eric Hambro, Faisal Azhar, Aurelien Rodriguez, Armand Joulin, Edouard Grave, and Guillaume Lample. 2023{\natexlab{a}}.
\newblock \href {https://arxiv.org/abs/2302.13971} {Llama: Open and efficient foundation language models}.
\newblock \emph{Preprint}, arXiv:2302.13971.

\bibitem[{Touvron et~al.(2023{\natexlab{b}})Touvron, Martin, Stone, Albert, Almahairi, Babaei, Bashlykov, Batra, Bhargava, Bhosale, Bikel, Blecher, Ferrer, Chen, Cucurull, Esiobu, Fernandes, Fu, Fu, Fuller, Gao, Goswami, Goyal, Hartshorn, Hosseini, Hou, Inan, Kardas, Kerkez, Khabsa, Kloumann, Korenev, Koura, Lachaux, Lavril, Lee, Liskovich, Lu, Mao, Martinet, Mihaylov, Mishra, Molybog, Nie, Poulton, Reizenstein, Rungta, Saladi, Schelten, Silva, Smith, Subramanian, Tan, Tang, Taylor, Williams, Kuan, Xu, Yan, Zarov, Zhang, Fan, Kambadur, Narang, Rodriguez, Stojnic, Edunov, and Scialom}]{arxiv:2307.09288}
Hugo Touvron, Louis Martin, Kevin Stone, Peter Albert, Amjad Almahairi, Yasmine Babaei, Nikolay Bashlykov, Soumya Batra, Prajjwal Bhargava, Shruti Bhosale, Dan Bikel, Lukas Blecher, Cristian~Canton Ferrer, Moya Chen, Guillem Cucurull, David Esiobu, Jude Fernandes, Jeremy Fu, Wenyin Fu, and 49 others. 2023{\natexlab{b}}.
\newblock \href {https://arxiv.org/abs/2307.09288} {Llama 2: Open foundation and fine-tuned chat models}.
\newblock \emph{Preprint}, arXiv:2307.09288.

\bibitem[{Tow(2023)}]{StableLMAlphaV2Models}
Jonathan Tow. 2023.
\newblock \href {https://huggingface.co/stabilityai/stablelm-base-alpha-7b-v2} {Stablelm alpha v2 models}.

\bibitem[{Tow et~al.(2023)Tow, Bellagente, Mahan, and Riquelme}]{StableLM-3B-4E1T}
Jonathan Tow, Marco Bellagente, Dakota Mahan, and Carlos Riquelme. 2023.
\newblock \href {https://huggingface.co/stabilityai/stablelm-3b-4e1t} {Stablelm 3b 4e1t}.

\bibitem[{Tunstall et~al.(2023)Tunstall, Beeching, Lambert, Rajani, Rasul, Belkada, Huang, von Werra, Fourrier, Habib, Sarrazin, Sanseviero, Rush, and Wolf}]{arxiv:2310.16944}
Lewis Tunstall, Edward Beeching, Nathan Lambert, Nazneen Rajani, Kashif Rasul, Younes Belkada, Shengyi Huang, Leandro von Werra, Clémentine Fourrier, Nathan Habib, Nathan Sarrazin, Omar Sanseviero, Alexander~M. Rush, and Thomas Wolf. 2023.
\newblock \href {https://arxiv.org/abs/2310.16944} {Zephyr: Direct distillation of lm alignment}.
\newblock \emph{Preprint}, arXiv:2310.16944.

\bibitem[{van~der Maaten and Hinton(2008)}]{vanderMaaten-2008-tsne}
Laurens van~der Maaten and Geoffrey Hinton. 2008.
\newblock \href {http://jmlr.org/papers/v9/vandermaaten08a.html} {Visualizing data using t-sne}.
\newblock \emph{Journal of Machine Learning Research}, 9(86):2579--2605.

\bibitem[{Vanroy(2024)}]{arxiv:2412.04092}
Bram Vanroy. 2024.
\newblock \href {https://arxiv.org/abs/2412.04092} {Geitje 7b ultra: A conversational model for dutch}.
\newblock \emph{Preprint}, arXiv:2412.04092.

\bibitem[{Varoquaux et~al.(2015)Varoquaux, Buitinck, Louppe, Grisel, Pedregosa, and Mueller}]{DBLP:journals/sigmobile/VaroquauxBLGPM15}
Ga{\"{e}}l Varoquaux, Lars Buitinck, Gilles Louppe, Olivier Grisel, Fabian Pedregosa, and Andreas Mueller. 2015.
\newblock \href {https://doi.org/10.1145/2786984.2786995} {Scikit-learn: Machine learning without learning the machinery}.
\newblock \emph{GetMobile Mob. Comput. Commun.}, 19(1):29--33.

\bibitem[{Virtanen et~al.(2020)Virtanen, Gommers, Oliphant, Haberland, Reddy, Cournapeau, Burovski, Peterson, Weckesser, Bright, {van der Walt}, Brett, Wilson, Millman, Mayorov, Nelson, Jones, Kern, Larson, Carey, Polat, Feng, Moore, {VanderPlas}, Laxalde, Perktold, Cimrman, Henriksen, Quintero, Harris, Archibald, Ribeiro, Pedregosa, {van Mulbregt}, and {SciPy 1.0 Contributors}}]{2020SciPy-NMeth}
Pauli Virtanen, Ralf Gommers, Travis~E. Oliphant, Matt Haberland, Tyler Reddy, David Cournapeau, Evgeni Burovski, Pearu Peterson, Warren Weckesser, Jonathan Bright, St{\'e}fan~J. {van der Walt}, Matthew Brett, Joshua Wilson, K.~Jarrod Millman, Nikolay Mayorov, Andrew R.~J. Nelson, Eric Jones, Robert Kern, Eric Larson, and 16 others. 2020.
\newblock \href {https://doi.org/10.1038/s41592-019-0686-2} {{{SciPy} 1.0: Fundamental Algorithms for Scientific Computing in Python}}.
\newblock \emph{Nature Methods}, 17:261--272.

\bibitem[{von Werra et~al.(2023)von Werra, Havrilla, reciprocated, Tow, cat state, Phung, Castricato, Matiana, Alan, Thakur, Bukhtiyarov, aaronrmm, Milo, Daniel, King, Shin, Kim, Wei, Romero, Pochinkov, Sanseviero, Adithyan, Siu, Simonini, Blagojevic, Song, Witten, alexandremuzio, and crumb}]{leandro-von-werra-2023-7790115}
Leandro von Werra, Alex Havrilla, Max reciprocated, Jonathan Tow, Aman cat state, Duy~V. Phung, Louis Castricato, Shahbuland Matiana, Alan, Ayush Thakur, Alexey Bukhtiyarov, aaronrmm, Fabrizio Milo, Daniel, Daniel King, Dong Shin, Ethan Kim, Justin Wei, Manuel Romero, and 10 others. 2023.
\newblock \href {https://doi.org/10.5281/zenodo.7790115} {{CarperAI/trlx: v0.6.0: LLaMa (Alpaca), Benchmark Util, T5 ILQL, Tests}}.

\bibitem[{Wang(2021)}]{mesh-transformer-jax}
Ben Wang. 2021.
\newblock \href {https://github.com/kingoflolz/mesh-transformer-jax} {{Mesh-Transformer-JAX: Model-Parallel Implementation of Transformer Language Model with JAX}}.

\bibitem[{Wang and Komatsuzaki(2021)}]{gpt-j}
Ben Wang and Aran Komatsuzaki. 2021.
\newblock \href {https://github.com/kingoflolz/mesh-transformer-jax} {{GPT-J-6B: A 6 Billion Parameter Autoregressive Language Model}}.

\bibitem[{Wang et~al.(2023{\natexlab{a}})Wang, Cheng, Yu, and Liu}]{openchat}
Guan Wang, Sijie Cheng, Qiying Yu, and Changling Liu. 2023{\natexlab{a}}.
\newblock \href {https://doi.org/10.5281/zenodo.8105775} {{OpenChat: Advancing Open-source Language Models with Imperfect Data}}.

\bibitem[{Wang et~al.(2024{\natexlab{a}})Wang, Cheng, Zhan, Li, Song, and Liu}]{arxiv:2309.11235}
Guan Wang, Sijie Cheng, Xianyuan Zhan, Xiangang Li, Sen Song, and Yang Liu. 2024{\natexlab{a}}.
\newblock \href {https://arxiv.org/abs/2309.11235} {Openchat: Advancing open-source language models with mixed-quality data}.
\newblock \emph{Preprint}, arXiv:2309.11235.

\bibitem[{Wang et~al.(2023{\natexlab{b}})Wang, Goodson, Lian, Pentland, Cook, Vong, and "Teknium"}]{OpenOrcaxOpenChatPreview2}
Guan Wang, Bleys Goodson, Wing Lian, Eugene Pentland, Austin Cook, Chanvichet Vong, and "Teknium". 2023{\natexlab{b}}.
\newblock \href {https://huggingface.co/Open-Orca/OpenOrcaxOpenChat-Preview2-13B} {Openorcaxopenchatpreview2: Llama2-13b model instruct-tuned on filtered openorcav1 gpt-4 dataset}.

\bibitem[{Wang et~al.(2024{\natexlab{b}})Wang, Zheng, Wang, Song, and Huang}]{shenzhi-wang-2024}
Shenzhi Wang, Yaowei Zheng, Guoyin Wang, Shiji Song, and Gao Huang. 2024{\natexlab{b}}.
\newblock \href {https://doi.org/10.57967/hf/2316} {Llama3-8b-chinese-chat (revision 6622a23)}.

\bibitem[{Wang et~al.(2023{\natexlab{c}})Wang, Ivison, Dasigi, Hessel, Khot, Chandu, Wadden, MacMillan, Smith, Beltagy, and Hajishirzi}]{arxiv:2306.04751}
Yizhong Wang, Hamish Ivison, Pradeep Dasigi, Jack Hessel, Tushar Khot, Khyathi~Raghavi Chandu, David Wadden, Kelsey MacMillan, Noah~A. Smith, Iz~Beltagy, and Hannaneh Hajishirzi. 2023{\natexlab{c}}.
\newblock \href {https://arxiv.org/abs/2306.04751} {How far can camels go? exploring the state of instruction tuning on open resources}.
\newblock \emph{Preprint}, arXiv:2306.04751.

\bibitem[{Wang et~al.(2024{\natexlab{c}})Wang, Chen, Dai, Xu, Li, and Wu}]{arxiv:2407.01906}
Zihan Wang, Deli Chen, Damai Dai, Runxin Xu, Zhuoshu Li, and Y.~Wu. 2024{\natexlab{c}}.
\newblock \href {https://arxiv.org/abs/2407.01906} {Let the expert stick to his last: Expert-specialized fine-tuning for sparse architectural large language models}.
\newblock \emph{Preprint}, arXiv:2407.01906.

\bibitem[{Warstadt et~al.(2020)Warstadt, Parrish, Liu, Mohananey, Peng, Wang, and Bowman}]{10.1162/tacl_a_00321}
Alex Warstadt, Alicia Parrish, Haokun Liu, Anhad Mohananey, Wei Peng, Sheng-Fu Wang, and Samuel~R. Bowman. 2020.
\newblock \href {https://doi.org/10.1162/tacl_a_00321} {Blimp: The benchmark of linguistic minimal pairs for english}.
\newblock \emph{Transactions of the Association for Computational Linguistics}, 8:377--392.

\bibitem[{Wortsman et~al.(2022)Wortsman, Ilharco, Gadre, Roelofs, Gontijo-Lopes, Morcos, Namkoong, Farhadi, Carmon, Kornblith, and Schmidt}]{arxiv:2203.05482}
Mitchell Wortsman, Gabriel Ilharco, Samir~Yitzhak Gadre, Rebecca Roelofs, Raphael Gontijo-Lopes, Ari~S. Morcos, Hongseok Namkoong, Ali Farhadi, Yair Carmon, Simon Kornblith, and Ludwig Schmidt. 2022.
\newblock \href {https://arxiv.org/abs/2203.05482} {Model soups: averaging weights of multiple fine-tuned models improves accuracy without increasing inference time}.
\newblock \emph{Preprint}, arXiv:2203.05482.

\bibitem[{Wu et~al.(2024)Wu, Zheng, He, and Yu}]{arxiv:2401.02731}
Haoyuan Wu, Haisheng Zheng, Zhuolun He, and Bei Yu. 2024.
\newblock \href {https://arxiv.org/abs/2401.02731} {Parameter-efficient sparsity crafting from dense to mixture-of-experts for instruction tuning on general tasks}.
\newblock \emph{Preprint}, arXiv:2401.02731.

\bibitem[{Xia et~al.(2024)Xia, Gao, Zeng, and Chen}]{arxiv:2310.06694}
Mengzhou Xia, Tianyu Gao, Zhiyuan Zeng, and Danqi Chen. 2024.
\newblock \href {https://arxiv.org/abs/2310.06694} {Sheared llama: Accelerating language model pre-training via structured pruning}.
\newblock \emph{Preprint}, arXiv:2310.06694.

\bibitem[{Xiao et~al.(2024)Xiao, Tian, Chen, Han, and Lewis}]{arxiv:2309.17453}
Guangxuan Xiao, Yuandong Tian, Beidi Chen, Song Han, and Mike Lewis. 2024.
\newblock \href {https://arxiv.org/abs/2309.17453} {Efficient streaming language models with attention sinks}.
\newblock \emph{Preprint}, arXiv:2309.17453.

\bibitem[{Xiao et~al.(2023)Xiao, Liu, Zhang, and Xing}]{arxiv:2311.13534}
Shitao Xiao, Zheng Liu, Peitian Zhang, and Xingrun Xing. 2023.
\newblock \href {https://arxiv.org/abs/2311.13534} {Lm-cocktail: Resilient tuning of language models via model merging}.
\newblock \emph{Preprint}, arXiv:2311.13534.

\bibitem[{Xin et~al.(2024{\natexlab{a}})Xin, Guo, Shao, Ren, Zhu, Liu, Ruan, Li, and Liang}]{arxiv:2405.14333}
Huajian Xin, Daya Guo, Zhihong Shao, Zhizhou Ren, Qihao Zhu, Bo~Liu, Chong Ruan, Wenda Li, and Xiaodan Liang. 2024{\natexlab{a}}.
\newblock \href {https://arxiv.org/abs/2405.14333} {Deepseek-prover: Advancing theorem proving in llms through large-scale synthetic data}.
\newblock \emph{Preprint}, arXiv:2405.14333.

\bibitem[{Xin et~al.(2024{\natexlab{b}})Xin, Ren, Song, Shao, Zhao, Wang, Liu, Zhang, Lu, Du, Gao, Zhu, Yang, Gou, Wu, Luo, and Ruan}]{arxiv:2408.08152}
Huajian Xin, Z.~Z. Ren, Junxiao Song, Zhihong Shao, Wanjia Zhao, Haocheng Wang, Bo~Liu, Liyue Zhang, Xuan Lu, Qiushi Du, Wenjun Gao, Qihao Zhu, Dejian Yang, Zhibin Gou, Z.~F. Wu, Fuli Luo, and Chong Ruan. 2024{\natexlab{b}}.
\newblock \href {https://arxiv.org/abs/2408.08152} {Deepseek-prover-v1.5: Harnessing proof assistant feedback for reinforcement learning and monte-carlo tree search}.
\newblock \emph{Preprint}, arXiv:2408.08152.

\bibitem[{Xu et~al.(2023{\natexlab{a}})Xu, Sun, Zheng, Geng, Zhao, Feng, Tao, and Jiang}]{arxiv:2304.12244}
Can Xu, Qingfeng Sun, Kai Zheng, Xiubo Geng, Pu~Zhao, Jiazhan Feng, Chongyang Tao, and Daxin Jiang. 2023{\natexlab{a}}.
\newblock \href {https://arxiv.org/abs/2304.12244} {Wizardlm: Empowering large language models to follow complex instructions}.
\newblock \emph{Preprint}, arXiv:2304.12244.

\bibitem[{Xu et~al.(2023{\natexlab{b}})Xu, Guo, Duan, and McAuley}]{arxiv:2304.01196}
Canwen Xu, Daya Guo, Nan Duan, and Julian McAuley. 2023{\natexlab{b}}.
\newblock \href {https://arxiv.org/abs/2304.01196} {Baize: An open-source chat model with parameter-efficient tuning on self-chat data}.
\newblock \emph{Preprint}, arXiv:2304.01196.

\bibitem[{Xu et~al.(2024{\natexlab{a}})Xu, Kim, Sharaf, and Awadalla}]{arxiv:2309.11674}
Haoran Xu, Young~Jin Kim, Amr Sharaf, and Hany~Hassan Awadalla. 2024{\natexlab{a}}.
\newblock \href {https://arxiv.org/abs/2309.11674} {A paradigm shift in machine translation: Boosting translation performance of large language models}.
\newblock \emph{Preprint}, arXiv:2309.11674.

\bibitem[{Xu et~al.(2024{\natexlab{b}})Xu, Sharaf, Chen, Tan, Shen, Durme, Murray, and Kim}]{arxiv:2401.08417}
Haoran Xu, Amr Sharaf, Yunmo Chen, Weiting Tan, Lingfeng Shen, Benjamin~Van Durme, Kenton Murray, and Young~Jin Kim. 2024{\natexlab{b}}.
\newblock \href {https://arxiv.org/abs/2401.08417} {Contrastive preference optimization: Pushing the boundaries of llm performance in machine translation}.
\newblock \emph{Preprint}, arXiv:2401.08417.

\bibitem[{{Xwin-LM Team}(2023)}]{xwin-lm}
{Xwin-LM Team}. 2023.
\newblock \href {https://github.com/Xwin-LM/Xwin-LM} {Xwin-lm}.

\bibitem[{Yadav et~al.(2023{\natexlab{a}})Yadav, Tam, Choshen, Raffel, and Bansal}]{arxiv:2306.01708}
Prateek Yadav, Derek Tam, Leshem Choshen, Colin Raffel, and Mohit Bansal. 2023{\natexlab{a}}.
\newblock \href {https://arxiv.org/abs/2306.01708} {Ties-merging: Resolving interference when merging models}.
\newblock \emph{Preprint}, arXiv:2306.01708.

\bibitem[{Yadav et~al.(2023{\natexlab{b}})Yadav, Tam, Choshen, Raffel, and Bansal}]{NEURIPS2023_1644c9af}
Prateek Yadav, Derek Tam, Leshem Choshen, Colin~A Raffel, and Mohit Bansal. 2023{\natexlab{b}}.
\newblock \href {https://proceedings.neurips.cc/paper_files/paper/2023/file/1644c9af28ab7916874f6fd6228a9bcf-Paper-Conference.pdf} {Ties-merging: Resolving interference when merging models}.
\newblock In \emph{Advances in Neural Information Processing Systems}, volume~36, pages 7093--7115. Curran Associates, Inc.

\bibitem[{Yamagiwa et~al.(2024)Yamagiwa, Takase, and Shimodaira}]{DBLP:conf/emnlp/YamagiwaTS24}
Hiroaki Yamagiwa, Yusuke Takase, and Hidetoshi Shimodaira. 2024.
\newblock \href {https://aclanthology.org/2024.findings-emnlp.28} {Axis tour: Word tour determines the order of axes in ica-transformed embeddings}.
\newblock In \emph{Findings of the Association for Computational Linguistics: {EMNLP} 2024, Miami, Florida, USA, November 12-16, 2024}, pages 477--506. Association for Computational Linguistics.

\bibitem[{Yang et~al.(2024{\natexlab{a}})Yang, Yang, Hui, Zheng, Yu, Zhou, Li, Li, Liu, Huang, Dong, Wei, Lin, Tang, Wang, Yang, Tu, Zhang, Ma, Yang, Xu, Zhou, Bai, He, Lin, Dang, Lu, Chen, Yang, Li, Xue, Ni, Zhang, Wang, Peng, Men, Gao, Lin, Wang, Bai, Tan, Zhu, Li, Liu, Ge, Deng, Zhou, Ren, Zhang, Wei, Ren, Liu, Fan, Yao, Zhang, Wan, Chu, Liu, Cui, Zhang, Guo, and Fan}]{qwen2}
An~Yang, Baosong Yang, Binyuan Hui, Bo~Zheng, Bowen Yu, Chang Zhou, Chengpeng Li, Chengyuan Li, Dayiheng Liu, Fei Huang, Guanting Dong, Haoran Wei, Huan Lin, Jialong Tang, Jialin Wang, Jian Yang, Jianhong Tu, Jianwei Zhang, Jianxin Ma, and 43 others. 2024{\natexlab{a}}.
\newblock \href {https://arxiv.org/abs/2407.10671} {Qwen2 technical report}.
\newblock \emph{Preprint}, arXiv:2407.10671.

\bibitem[{Yang et~al.(2024{\natexlab{b}})Yang, Zhang, Kuang, Xie, Huang, and Ananiadou}]{arxiv:2309.13567}
Kailai Yang, Tianlin Zhang, Ziyan Kuang, Qianqian Xie, Jimin Huang, and Sophia Ananiadou. 2024{\natexlab{b}}.
\newblock \href {https://arxiv.org/abs/2309.13567} {Mentallama: Interpretable mental health analysis on social media with large language models}.
\newblock \emph{Preprint}, arXiv:2309.13567.

\bibitem[{Yang et~al.(2022)Yang, Wang, Gan, Zhu, Zhang, Wu, Gao, Zhang, and Sakai}]{arxiv:2210.08590}
Ping Yang, Junjie Wang, Ruyi Gan, Xinyu Zhu, Lin Zhang, Ziwei Wu, Xinyu Gao, Jiaxing Zhang, and Tetsuya Sakai. 2022.
\newblock \href {https://arxiv.org/abs/2210.08590} {Zero-shot learners for natural language understanding via a unified multiple choice perspective}.
\newblock \emph{Preprint}, arXiv:2210.08590.

\bibitem[{Yax et~al.(2025)Yax, Oudeyer, and Palminteri}]{yax2025phylolm}
Nicolas Yax, Pierre-Yves Oudeyer, and Stefano Palminteri. 2025.
\newblock \href {https://openreview.net/forum?id=rTQNGQxm4K} {Phylo{LM}: Inferring the phylogeny of large language models and predicting their performances in benchmarks}.
\newblock In \emph{The Thirteenth International Conference on Learning Representations}.

\bibitem[{Yu et~al.(2024{\natexlab{a}})Yu, Yu, Yu, Huang, and Li}]{arxiv:2311.03099}
Le~Yu, Bowen Yu, Haiyang Yu, Fei Huang, and Yongbin Li. 2024{\natexlab{a}}.
\newblock \href {https://arxiv.org/abs/2311.03099} {Language models are super mario: Absorbing abilities from homologous models as a free lunch}.
\newblock \emph{Preprint}, arXiv:2311.03099.

\bibitem[{Yu et~al.(2024{\natexlab{b}})Yu, Jiang, Shi, Yu, Liu, Zhang, Kwok, Li, Weller, and Liu}]{arxiv:2309.12284}
Longhui Yu, Weisen Jiang, Han Shi, Jincheng Yu, Zhengying Liu, Yu~Zhang, James~T. Kwok, Zhenguo Li, Adrian Weller, and Weiyang Liu. 2024{\natexlab{b}}.
\newblock \href {https://arxiv.org/abs/2309.12284} {Metamath: Bootstrap your own mathematical questions for large language models}.
\newblock \emph{Preprint}, arXiv:2309.12284.

\bibitem[{Yue et~al.(2024{\natexlab{a}})Yue, Ni, Zhang, Zheng, Liu, Zhang, Stevens, Jiang, Ren, Sun, Wei, Yu, Yuan, Sun, Yin, Zheng, Yang, Liu, Huang, Sun, Su, and Chen}]{arxiv:2311.16502}
Xiang Yue, Yuansheng Ni, Kai Zhang, Tianyu Zheng, Ruoqi Liu, Ge~Zhang, Samuel Stevens, Dongfu Jiang, Weiming Ren, Yuxuan Sun, Cong Wei, Botao Yu, Ruibin Yuan, Renliang Sun, Ming Yin, Boyuan Zheng, Zhenzhu Yang, Yibo Liu, Wenhao Huang, and 3 others. 2024{\natexlab{a}}.
\newblock \href {https://arxiv.org/abs/2311.16502} {Mmmu: A massive multi-discipline multimodal understanding and reasoning benchmark for expert agi}.
\newblock \emph{Preprint}, arXiv:2311.16502.

\bibitem[{Yue et~al.(2024{\natexlab{b}})Yue, Zheng, Zhang, and Chen}]{arxiv:2405.03548}
Xiang Yue, Tuney Zheng, Ge~Zhang, and Wenhu Chen. 2024{\natexlab{b}}.
\newblock \href {https://arxiv.org/abs/2405.03548} {Mammoth2: Scaling instructions from the web}.
\newblock \emph{Preprint}, arXiv:2405.03548.

\bibitem[{{YuLan-Team}(2023)}]{YuLan-Chat}
{YuLan-Team}. 2023.
\newblock \href {https://github.com/RUC-GSAI/YuLan-Chat} {Yulan-chat: An open-source bilingual chatbot}.

\bibitem[{Zellers et~al.(2019)Zellers, Holtzman, Bisk, Farhadi, and Choi}]{arxiv:1905.07830}
Rowan Zellers, Ari Holtzman, Yonatan Bisk, Ali Farhadi, and Yejin Choi. 2019.
\newblock \href {https://arxiv.org/abs/1905.07830} {Hellaswag: Can a machine really finish your sentence?}
\newblock \emph{Preprint}, arXiv:1905.07830.

\bibitem[{Zhang and Sennrich(2019)}]{arxiv:1910.07467}
Biao Zhang and Rico Sennrich. 2019.
\newblock \href {https://arxiv.org/abs/1910.07467} {Root mean square layer normalization}.
\newblock \emph{Preprint}, arXiv:1910.07467.

\bibitem[{Zhang et~al.(2024{\natexlab{a}})Zhang, Song, Ye, and Gao}]{arxiv:2311.07052}
Chen Zhang, Dawei Song, Zheyu Ye, and Yan Gao. 2024{\natexlab{a}}.
\newblock \href {https://arxiv.org/abs/2311.07052} {Towards the law of capacity gap in distilling language models}.
\newblock \emph{Preprint}, arXiv:2311.07052.

\bibitem[{Zhang et~al.(2024{\natexlab{b}})Zhang, Du, Chen, Liang, Luo, Zheng, Zhu, Cheng, Xu, Guo, Zhang, Qu, Wang, Yuan, Li, Wang, Liu, Tsai, Zhang, Lin, Huang, and Fu}]{arxiv:2401.11944}
Ge~Zhang, Xinrun Du, Bei Chen, Yiming Liang, Tongxu Luo, Tianyu Zheng, Kang Zhu, Yuyang Cheng, Chunpu Xu, Shuyue Guo, Haoran Zhang, Xingwei Qu, Junjie Wang, Ruibin Yuan, Yizhi Li, Zekun Wang, Yudong Liu, Yu-Hsuan Tsai, Fengji Zhang, and 3 others. 2024{\natexlab{b}}.
\newblock \href {https://arxiv.org/abs/2401.11944} {Cmmmu: A chinese massive multi-discipline multimodal understanding benchmark}.
\newblock \emph{Preprint}, arXiv:2401.11944.

\bibitem[{Zhang et~al.(2022{\natexlab{a}})Zhang, Gan, Wang, Zhang, Zhang, Yang, Gao, Wu, Dong, He, Zhuo, Yang, Huang, Li, Wu, Lu, Zhu, Chen, Han, Pan, Wang, Wang, Wu, Zeng, and Chen}]{fengshenbang}
Jiaxing Zhang, Ruyi Gan, Junjie Wang, Yuxiang Zhang, Lin Zhang, Ping Yang, Xinyu Gao, Ziwei Wu, Xiaoqun Dong, Junqing He, Jianheng Zhuo, Qi~Yang, Yongfeng Huang, Xiayu Li, Yanghan Wu, Junyu Lu, Xinyu Zhu, Weifeng Chen, Ting Han, and 6 others. 2022{\natexlab{a}}.
\newblock Fengshenbang 1.0: Being the foundation of chinese cognitive intelligence.
\newblock \emph{CoRR}, abs/2209.02970.

\bibitem[{Zhang et~al.(2024{\natexlab{c}})Zhang, Ding, Qi, Zeng, Li, Zhu, Chen, and Zhou}]{UltraMedical}
Kaiyan Zhang, Ning Ding, Biqing Qi, Sihang Zeng, Haoxin Li, Xuekai Zhu, Zhang-Ren Chen, and Bowen Zhou. 2024{\natexlab{c}}.
\newblock \href {https://github.com/TsinghuaC3I/UltraMedical} {Ultramedical: Building specialized generalists in biomedicine.}

\bibitem[{Zhang et~al.(2022{\natexlab{b}})Zhang, Roller, Goyal, Artetxe, Chen, Chen, Dewan, Diab, Li, Lin, Mihaylov, Ott, Shleifer, Shuster, Simig, Koura, Sridhar, Wang, and Zettlemoyer}]{arxiv:2205.01068}
Susan Zhang, Stephen Roller, Naman Goyal, Mikel Artetxe, Moya Chen, Shuohui Chen, Christopher Dewan, Mona Diab, Xian Li, Xi~Victoria Lin, Todor Mihaylov, Myle Ott, Sam Shleifer, Kurt Shuster, Daniel Simig, Punit~Singh Koura, Anjali Sridhar, Tianlu Wang, and Luke Zettlemoyer. 2022{\natexlab{b}}.
\newblock \href {https://arxiv.org/abs/2205.01068} {Opt: Open pre-trained transformer language models}.
\newblock \emph{Preprint}, arXiv:2205.01068.

\bibitem[{Zhang et~al.(2023)Zhang, Aljunied, Gao, Chia, and Bing}]{arxiv:2306.05179}
Wenxuan Zhang, Sharifah~Mahani Aljunied, Chang Gao, Yew~Ken Chia, and Lidong Bing. 2023.
\newblock \href {https://arxiv.org/abs/2306.05179} {M3exam: A multilingual, multimodal, multilevel benchmark for examining large language models}.
\newblock \emph{Preprint}, arXiv:2306.05179.

\bibitem[{Zhao et~al.(2018)Zhao, Wang, Yatskar, Ordonez, and Chang}]{arxiv:1804.06876}
Jieyu Zhao, Tianlu Wang, Mark Yatskar, Vicente Ordonez, and Kai-Wei Chang. 2018.
\newblock \href {https://arxiv.org/abs/1804.06876} {Gender bias in coreference resolution: Evaluation and debiasing methods}.
\newblock \emph{Preprint}, arXiv:1804.06876.

\bibitem[{Zhao et~al.(2023{\natexlab{a}})Zhao, Kaga, and Sawada}]{rinna-youri-7b}
Tianyu Zhao, Akio Kaga, and Kei Sawada. 2023{\natexlab{a}}.
\newblock \href {https://huggingface.co/rinna/youri-7b} {rinna/youri-7b}.

\bibitem[{Zhao and Sawada(2023)}]{rinna-japanese-gpt-neox-3.6b}
Tianyu Zhao and Kei Sawada. 2023.
\newblock \href {https://huggingface.co/rinna/japanese-gpt-neox-3.6b} {rinna/japanese-gpt-neox-3.6b}.

\bibitem[{Zhao et~al.(2023{\natexlab{b}})Zhao, Wakatsuki, Kaga, Mitsuda, and Sawada}]{rinna-bilingual-gpt-neox-4b}
Tianyu Zhao, Toshiaki Wakatsuki, Akio Kaga, Koh Mitsuda, and Kei Sawada. 2023{\natexlab{b}}.
\newblock \href {https://huggingface.co/rinna/bilingual-gpt-neox-4b} {rinna/bilingual-gpt-neox-4b}.

\bibitem[{Zheng et~al.(2023)Zheng, Chiang, Sheng, Zhuang, Wu, Zhuang, Lin, Li, Li, Xing, Zhang, Gonzalez, and Stoica}]{arxiv:2306.05685}
Lianmin Zheng, Wei-Lin Chiang, Ying Sheng, Siyuan Zhuang, Zhanghao Wu, Yonghao Zhuang, Zi~Lin, Zhuohan Li, Dacheng Li, Eric~P. Xing, Hao Zhang, Joseph~E. Gonzalez, and Ion Stoica. 2023.
\newblock \href {https://arxiv.org/abs/2306.05685} {Judging llm-as-a-judge with mt-bench and chatbot arena}.
\newblock \emph{Preprint}, arXiv:2306.05685.

\bibitem[{Zhong et~al.(2023)Zhong, Cui, Guo, Liang, Lu, Wang, Saied, Chen, and Duan}]{arxiv:2304.06364}
Wanjun Zhong, Ruixiang Cui, Yiduo Guo, Yaobo Liang, Shuai Lu, Yanlin Wang, Amin Saied, Weizhu Chen, and Nan Duan. 2023.
\newblock \href {https://arxiv.org/abs/2304.06364} {Agieval: A human-centric benchmark for evaluating foundation models}.
\newblock \emph{Preprint}, arXiv:2304.06364.

\bibitem[{Zhou et~al.(2025)Zhou, Chen, Cahyawijaya, Duan, and Cai}]{zhou-etal-2025-linguistic}
Xinyu Zhou, Delong Chen, Samuel Cahyawijaya, Xufeng Duan, and Zhenguang Cai. 2025.
\newblock \href {https://aclanthology.org/2025.coling-main.459/} {Linguistic minimal pairs elicit linguistic similarity in large language models}.
\newblock In \emph{Proceedings of the 31st International Conference on Computational Linguistics}, pages 6866--6888, Abu Dhabi, UAE. Association for Computational Linguistics.

\bibitem[{Zhu et~al.(2023{\natexlab{a}})Zhu, Frick, Wu, Zhu, Ganesan, Chiang, Zhang, and Jiao}]{starling2023}
Banghua Zhu, Evan Frick, Tianhao Wu, Hanlin Zhu, Karthik Ganesan, Wei-Lin Chiang, Jian Zhang, and Jiantao Jiao. 2023{\natexlab{a}}.
\newblock Starling-7b: Improving llm helpfulness \& harmlessness with rlaif.

\bibitem[{Zhu et~al.(2023{\natexlab{b}})Zhu, Sharma, Frujeri, Dong, Zhu, Jordan, and Jiao}]{arxiv:2306.02231}
Banghua Zhu, Hiteshi Sharma, Felipe~Vieira Frujeri, Shi Dong, Chenguang Zhu, Michael~I. Jordan, and Jiantao Jiao. 2023{\natexlab{b}}.
\newblock \href {https://arxiv.org/abs/2306.02231} {Fine-tuning language models with advantage-induced policy alignment}.
\newblock \emph{Preprint}, arXiv:2306.02231.

\bibitem[{Zhu et~al.(2025)Zhu, Ahmed, Kuditipudi, and Liang}]{zhu2025independence}
Sally Zhu, Ahmed~M Ahmed, Rohith Kuditipudi, and Percy Liang. 2025.
\newblock \href {https://openreview.net/forum?id=leJWwts8P9} {Independence tests for language models}.

\bibitem[{Zhuang et~al.(2025)Zhuang, Wu, Wen, Li, Jiao, and Ramchandran}]{zhuang2025embedllm}
Richard Zhuang, Tianhao Wu, Zhaojin Wen, Andrew Li, Jiantao Jiao, and Kannan Ramchandran. 2025.
\newblock \href {https://openreview.net/forum?id=Fs9EabmQrJ} {Embed{LLM}: Learning compact representations of large language models}.
\newblock In \emph{The Thirteenth International Conference on Learning Representations}.

\bibitem[{Ziegler et~al.(2020)Ziegler, Stiennon, Wu, Brown, Radford, Amodei, Christiano, and Irving}]{arxiv:1909.08593}
Daniel~M. Ziegler, Nisan Stiennon, Jeffrey Wu, Tom~B. Brown, Alec Radford, Dario Amodei, Paul Christiano, and Geoffrey Irving. 2020.
\newblock \href {https://arxiv.org/abs/1909.08593} {Fine-tuning language models from human preferences}.
\newblock \emph{Preprint}, arXiv:1909.08593.

\end{thebibliography}
